\documentclass[letterpaper, 10 pt, journal]{IEEEtran}  

\IEEEoverridecommandlockouts                              

\usepackage{amsmath,amsfonts,amssymb}
\usepackage{mathtools}
\usepackage{commath}
\usepackage{color}
\usepackage{threeparttable}
\usepackage{multirow}
\usepackage{makecell}
\usepackage{algorithm}
\usepackage{algorithmic}
\usepackage{url}
\usepackage{graphicx}
\usepackage{subfigure}
\usepackage{rotating}
\usepackage{makecell}
\usepackage{adjustbox}
\begin{document}


\title{
High Definition Map Mapping and Update: A General Overview and Future Directions
}

\author{ Benny Wijaya$^{1}$, Kun Jiang$^{1}$, Mengmeng Yang$^{1}$, Tuopu Wen$^{1}$, Yunlong Wang$^{1}$, Xuewei Tang$^{1}$, Zheng Fu$^{1}$, Gracelynn Soesanto$^{2}$, Taohua Zhou$^{1}$, Jinyu Miao$^{1}$, Peijin Jia$^{1}$, and Diange Yang$^{1}$  
\thanks{$^{1}$Benny Wijaya, Kun Jiang, Mengmeng Yang, Tuopu Wen, Yunlong Wang, Xuewei Tang, Zheng Fu, Gracelynn Soesanto, Taohua Zhou, Jinyu Miao, Peijin Jia, and Diange Yang are with 
State Key Laboratory of Automotive Safety and Energy, and Center for Intelligent Connected Vehicles and Transportation, 
the School of Vehicle and Mobility,
Tsinghua University, Beijing, 100084, China. 
}
\thanks{$^{2}$ Gracelynn Soesanto is with Department of Industrial Engineering, Tsinghua University, Beijing 100084, China
}
\thanks{Corresponding author: Diange Yang, email: ydg@mail.tsinghua.edu.cn, Mengmeng Yang, email: yangmm\_qh@tsinghua.edu.cn
}%
}



\maketitle
\thispagestyle{empty}
\pagestyle{empty}

\begin{abstract}
   Along with the rapid growth of autonomous vehicles (AVs), more and more demands are required for environment perception technology. Among others, HD mapping has become one of the more prominent roles in helping the vehicle realize essential tasks such as localization and path planning. While increasing research effort have been directed toward HD Map development. However, a comprehensive overview of the overall HD map mapping and update framework is still lacking. This article introduces the development and current state of the algorithm involved in creating HD map mapping and its maintenance. As part of this study, the primary data preprocessing approach of processing raw data to information ready to feed for mapping and update purposes, semantic segmentation, and localization are also briefly reviewed. Moreover, the map taxonomy, ontology, and quality assessment are extensively discussed, the map data's general representation method is presented, and the mapping algorithm ranging from SLAM to transformers learning-based approaches are also discussed. The development of the HD map update algorithm, from change detection to the update methods, is also presented. Finally, the authors discuss possible future developments and the remaining challenges in HD map mapping and update technology. This paper simultaneously serves as a position paper and tutorial for those new to HD map mapping and update domains.
\end{abstract}


\section{INTRODUCTION}

    \IEEEPARstart{I}ntelligent vehicles use various sensors to achieve different levels of autonomy (L1-L5, e.g., see \cite{SAEInternational2018}), such as cameras, Global Navigation Satellite System (GNSS), Radio Detection and Ranging (Radar), Light Detection and Ranging (LIDAR). However, these sensors have a limited perception range, and they are very vulnerable to corner-case situations as well as weather changes. To overcome the limitations, the pre-built digital 3D map can be utilized to improve the perception and robustness of the localization for intelligent vehicles. Many intelligent vehicles rely on very precise 3D maps \cite{Levinson2011}, which are also called high-definition (HD) maps.
    \begin{figure*}[h]
        \centering
        \includegraphics[width=0.92 \linewidth]{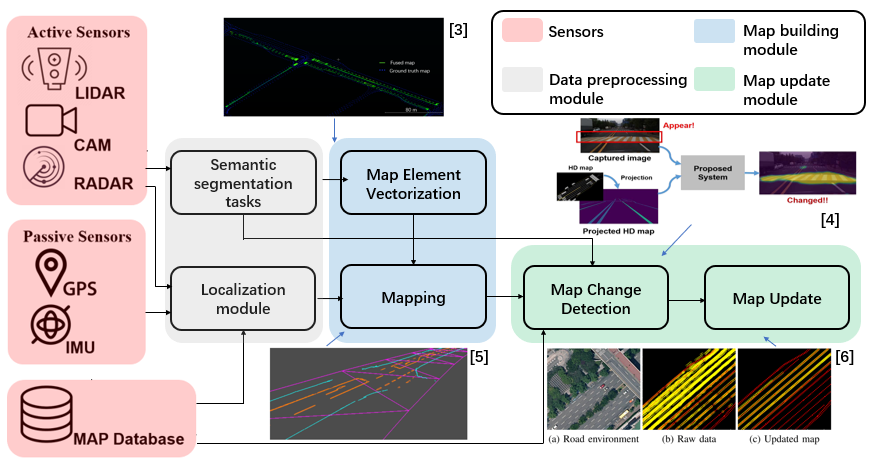}
        \caption{General pipeline of the processes involved in HD map mapping and update sequence. Figure is created and modified based on the depictions in \cite{Wijaya2022}\cite{Heo2020}\cite{Herb2019}\cite{Kim2021}}
        \label{fig:pipeline}
    \end{figure*} 
    An HD map is precise with rich lane-level information for autonomous driving purposes and has revolutionized standard maps in multiple paradigms. HD maps are basically the ground truth models representation at a scale of nearly 1:1 and they are made by machines for machines \cite{Meng2022}. It can provide robust information about the static environment in a more extensive range than traditional onboard sensors of more than 200 m or even around corners. The features in the map can be fused with the recognition results from camera/LIDAR to realize high-accuracy localization of the vehicle \cite{Schreiber2013}. Compared with the car navigation map, an HD map significantly improves the localization accuracy to a few centimeters level\cite{Wen2020}. 
    HD maps can be regarded as an additional source of information in the application of ADAS since they can heavily impact confidence in other sensors, improve computational efficiency, and increase convenience in terms of data accessibility, which is very crucial in increasing autonomy while still maintaining the safety standard of intelligent vehicles \cite{Felix2021}.

    Furthermore, HD map also has richer and more detailed content, such as lane boundaries, lane centerlines, road markings on the ground, traffic signs, traffic poles, and traffic light position. It also has several key attributes in HD maps in lane boundaries, such as type, color, and width. This realization reflects a more realistic detail of the surrounding real-world situation, which is useful in implementing a higher level of autonomy of intelligent vehicles on the road. However, these advantages do not come easy because maintaining a precise HD map would require a huge effort since subtle changes of the road features change every day at all times in random places. For example, the lane boundary or lane centerlines can subtly fade away, influencing the map-matching algorithm for the localization. This type of error can yield a bad localization result and may even lead to accidents. Thus, it is desired to have an accurate and frequent update of HD map. At present, the production of HD maps requires professional data collection, that is, professional surveying and mapping using the Mobile Mapping System (MMS) \cite{Chen2010}\cite{Joshi2015}. These mapping fleets are very expensive to set up, limiting the number and map coverage area.

    Moreover, this limitation also influences the map update frequency, which is very important in keeping the HD map representative of the actual world situation. Since each mapping vehicle's workload is so much, it is long and prone to errors in the process of making and updating HD maps. In addition, when we also consider the rapid development of the urban area with thousands of km of the road being built every day, relying fully on these mapping vehicle fleets becomes insufficient to satisfy the HD map's needs. Based on the above points, it is concluded that the use of the MMS fleet for map update cannot meet the actual needs of map update because it lacks real-time dynamic update capabilities.

    In the past few years, there has been rapid development of artificial intelligence technology and intelligence in mass-produced vehicles, such as ADAS functions, which include lane detection in lane-keeping functions. These trends in the automotive industry can help detect road element information in real-time, providing a wealth of data for mapping and updates. As a result, it is now possible to map and update the HD map based on the mass-produced vehicle data called the crowdsourced method. At present, more and more researchers focus on crowdsource mapping and update of HD maps to reduce the cost and increase the update frequencies. The research approaches in this area mainly vary in the sensors believed to be equipped for intelligent vehicles' future. Some conservative approaches this by only using cheap monocular camera and consumer-grade GNSS/IMU to update the map. While others are confident that the cost of LIDAR sensors will reduce significantly in the future, making it affordable for mass-produced vehicles. When the vehicles are driving, they will be collecting data, and the collected data of every car can be aggregated and then used to update HD maps. This is the concept of crowdsourcing updates of HD maps. Once the map update is completed in the cloud, the update package can be passed back to the vehicles. Then real-time HD map updates and services can be realized. The major drawback of crowdsourced data is its high uncertainty. The challenge is the appropriate update of HD maps by reflecting environmental changes while overcoming various uncertainties from low-quality observations.

    There is a survey on high-definition map for automated driving, which has been published by \cite{Liu2020} where they introduced the comprehensive review of navigation history which ultimately leads to HD map development, focusing on HD map structure, functionalities, and standardization. Furthermore, they also provide an analysis of HD map-based vehicle localization. In 2023, \cite{Bao2023} wrote a review on HD map creation that covers the extraction methods for HD map elements such as road networks and poles. However, they fail to cover the new trend in online HD map construction as started by Tsinghua MARS Lab in the CVPR autonomous driving challenge 2023, and they are limited to only the map element reconstruction task.  Recently, \cite{Tang2023a} organizes the survey of creation process of HD map update through visual sensors. This survey manage to provide a specific view of one of the lifecycle of HD map but not the whole picture as a whole. Different from them, in this review, we focused on the full lifecycle phases of the modules required in the making of HD map and its maintenance, which at the same time also give the readers a sufficient understanding of the process of mapping and update starting from the data preprocessing module, map building module, and map update module. The process mainly contain mainly three steps, data collection and its processes accordingly, map creation or mapping processes, and map maintenance or map update.
    The main contributions of our work can be summarized as four points: 
    \begin{enumerate}
    \item  An organized survey of the processes included in the four-step lifecycles of HD maps (data acquisition, data processing, mapping, and update).
    \item The latest developments in the main algorithms, specifically semantic segmentation and localization methods used in the data preprocessing, are presented.
    \item Latest trends on HD map mapping (SLAM and crowdsourced) and HD map update such as change detection and crowdsourced update method are fully investigated.
    \item A list of the remaining challenge and future directions which can be useful for the development of HD map in autonomous driving.
    \end{enumerate}

    The remainder of the paper is organized as follows. Section 2 briefly introduces the HD map taxonomy, ontology, map representation and the requirements of HD map. Section 3 discusses the data preprocessing module involved in the acquisition step. Section 4 discusses the general mapping approaches related to SLAM and crowdsourced methods in autonomous driving. Section 5 provides an overview of HD map updates which emphasizes on change detection research and crowdsourced update. Section 6 explores the challenges and research direction that can be explored further to advance the technology in possible future development. Section 7 draws conclusions on current research works.

\section{HD Map Introduction}

\subsection{HD Map Taxonomy and Ontology}
The core idea of HD maps is originated from the necessity to localize vehicles as accurately as possible in order to ensure safety in autonomy mode. The early generation of digital maps is not able to satisfy this requirement since it only operates at the lane-level accuracy \cite{Zhang2016}. Although vehicle positioning technology has made significant progress, it quickly reaches the limit of what is possible to achieve in terms of accuracy without the aid of an accurate reference in the form of a map. The clear definition of high-definition map function-wise can also be reduced to detailed digital maps information that provides highly accurate data to support the application of level 3 autonomous vehicles and above.

HD maps typically include several layers, each representing a different aspect of the environment. Several standards have been derived from dealing with the complexity of the driving environment, such as: OpenDrive \cite{asam2021}, NDS\cite{Hubertus2019}, ADASIS\cite{TomTom2020}, Local Dynamic Map (LDM)\cite{ETSI2011}, and Tsinghua \cite{Jiang2019}.
OpenDrive consists of two main map contents: coordinate system and road network \cite{asam2021}. The coordinate system of the road geometry is unified into the ground plane, and the map elements are represented in the distance to the road centerline axis. In the road network, the information stored is reference lines, lane and line properties, 3D geometry information, elevation profile, lane boundary information, and other information such as parking lots and railway tracks. This information can also be utilized for path planning and map monitoring purposes \cite{Diaz-Diaz2022}.
\begin{figure}[h]
    \centering
    \includegraphics[width=1 \linewidth]{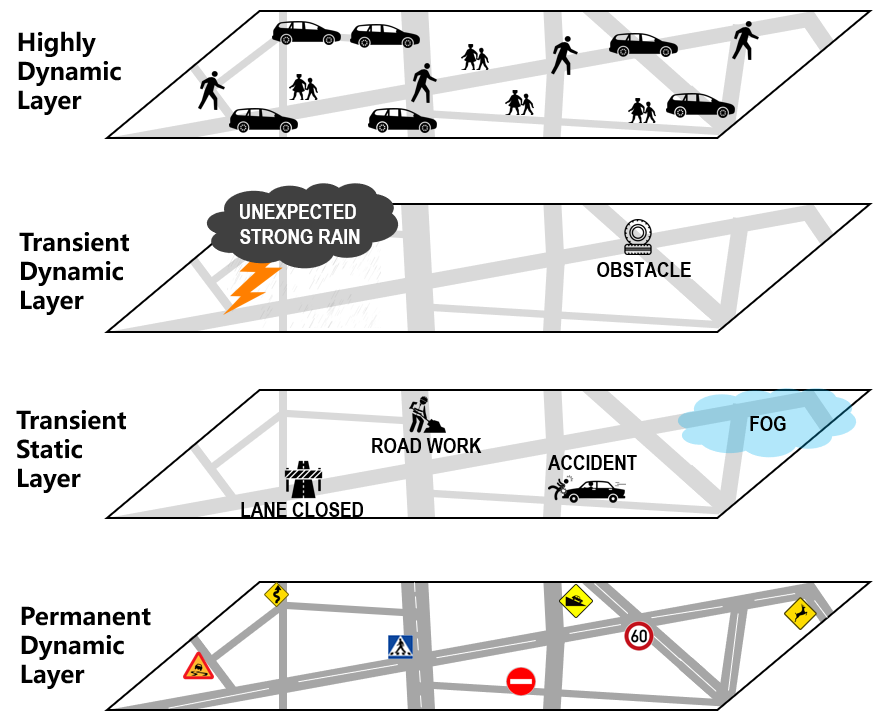}
    \caption{The illustration of HD map layers according to the definition derived from LDM. The figure is redrawn and modified based on depictions in \cite{AECC2020}}
    \label{fig:layers}
\end{figure} 

NDS can be divided several layers of information. Some basic information include lanes, localization landmarks, obstacles, and routing \cite{Hubertus2019}. Similar to the previous standard, the lane information is represented by some basic properties such as lane geometries, lane boundaries, lane groups, and lane relations. Splines and elevation profiles represent Lane geometries, and lane groups represent the lane-group network. Localization landmarks are represented by signs, poles, walls, traffic lights, and many more. They are stored in a vectorized manner, such as splines, polyline, and height profiles. The volatile data, such as routings, are represented by traffic condition, speed limits, and road signs.

ADASIS is designed for ADAS application and emphasizes the data transmission between server and the vehicles.The map information provided is similar to other map standards, including road information, map elements, and road topology \cite{TomTom2020}. Derived from the Local Dynamic Map (LDM)\cite{ETSI2011}, which is standardized in Europe, Automotive Edge Computing Consortium (AECC) \cite{AECC2020} model HD map into several dynamic layers of information which can be described as:

\begin{enumerate}
    \item \textbf{Highly Dynamic Layer:} This layer includes information that changes in a matter of seconds or less, such as position and state information of vehicles, trucks, buses, motorbikes, bicycles, and pedestrians.
    \item \textbf{Transient Dynamic Layer:} This layer includes information that might experience change in minutes, such as fallen tree trunks, illegally parked vehicles, sudden changes in local weather, tornado, and trash.
    \item \textbf{Transient Static Layer:} This layer includes information that might experience a change in several hours, such as the position and state information of the road work, traffic accidents, lane closures, and vehicles broken down.
    \item \textbf{Permanent Static Layer:} This layer includes information that changes at daily intervals or even longer, such as lanes, traffic signals, traffic rules, and the 3D geometry structure of the road topology. This layer is also often called a static map.
\end{enumerate}

\begin{figure}[h]
    \centering
    \includegraphics[width=1 \linewidth]{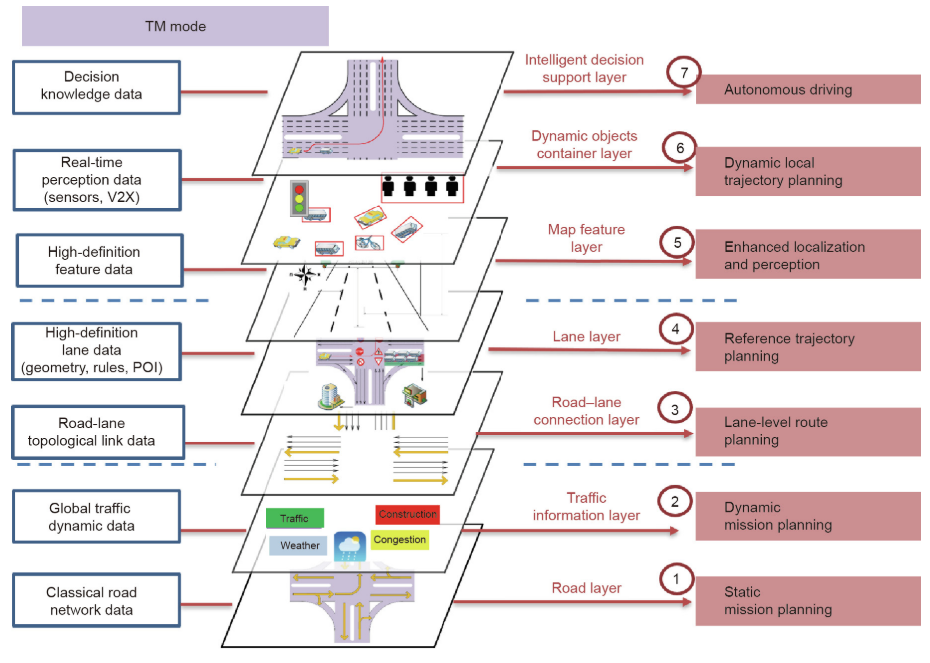}
    \caption{The illustration of the seven layer adaptive map architecture for autonomous driving, V2X: vehicle-to-everything.  The figure is obtained from \cite{Jiang2019} work with permission from the author}
    \label{fig:tsinghua}
\end{figure} 

Tsinghua standard focused on the application of autonomous driving by providing information related to perception, positioning, and decision-making \cite{Jiang2019}. Tsinghua map standards can be divided into seven layers: road layer, traffic information layer, road-lane connection layer, lane layer, map feature layer, dynamic objects container layer, and intelligent decision support layer. The objective of the map data is to achieve optimal routing directly on large lane-level road networks. The effectiveness of the map is then shown in \cite{jiang2022}, where the intelligent support layer is used as an information container for cooperative perception. 

The ontological approach for defining a shared conceptualization as a formal yet explicit specification can be applied to HD maps by describing it as several aspects of the environment in terms of semantic, temporal, and spatial data. 
The semantics part of the HD map can be referred to as the naming convention of the HD map as it provides all the information necessary for autonomous diving, including location, class, and types. Deriving from the ontological approach, \cite{qiu2020} proposes a knowledge architecture layer that differentiates low-level and high-level ontologies based on various map data to model the road environment. Therefore, the comparative quality aspects for each environment terminology can also be set quantitatively and qualitatively.

\subsection{HD Map Requirement and Evaluation}
In general, people in the industry will refer to the accuracy requirement for a map to be called HD map as 10-20 cm \cite{Liu2020}. However, to the author's knowledge, the international community as a whole has never before drafted the specific requirements for such a map in an official document. The general consensus remains that the accuracy of HD maps needs to be maintained. \cite{Javanmardi2018} specifies the important factors that need to be satisfied for accurate map localization using HD map, such as map element location accuracy, feature dilution of precision (FDOP) that signifies the distribution of features in the map space, layout similarity, and representation quality. Among these, layout similarity correlates the most, followed by representation quality, location accuracy, and FDOP. \cite{Brimicombe2020} creates the guideline to measure the quality of HD map, which includes the classification and the metrics used to determine the quality, including consistency, accuracy, and completeness, as shown in Table. \ref{tab:quality}.

\begin{table}[h]
\caption{Quality dimensions of map environment aspects following the guideline of \cite{Brimicombe2020}}
\label{tab:quality}
\centering
\def\arraystretch{1.3}
\setlength\tabcolsep{4pt} 
\begin{tabular}{clcccccccccccccc}
\cline{1-3}
& \multicolumn{1}{c}{\textbf{Quality Name}}& $\text{Quality Metrics}$ \\ \hline
\multirow{5}{*}{\thead{\textbf{Semantic}\\ \textbf{Data}}}
& Naming & Name accuracy   \\
& Classification types & Number of class\\
& Classification accuracy & $\%$  \\
& Semantic consistency &yes/no   \\
& Semantic completeness &$\%$    \\ \hline

\multirow{2}{*}{\thead{\textbf{Temporal}\\ \textbf{Data}}} 
& Temporal accuracy & $\%$  \\
& Temporal frequency & $s$ (and related sub-units)  \\
& Temporal consistency & yes/no \\
\hline
\multirow{3}{*}{\thead{\textbf{Spatial}\\ \textbf{Data}}} 
& Spatial accuracy& $m$ (and related sub-units) \\
& Spatial coverage & $m^2$ (and related sub-units) \\
& Spatial precision& $m$ (and related sub-units) \\
& Spatial resolution& $point/m^3$\\
\hline
\end{tabular}
\end{table}

Temporal refers to the time in which the HD map is created or updated. The metrics related to this aspect include accuracy, frequency, and consistency. Finally, the most essential part of the HD map is the spatial data, including the position or location metrics such as accuracy, coverage, precision, and resolution. Readers who are interested in the explanation of each of the quality dimensions can refer to \cite{Logo2021}, which describes the meaning of each quality dimension clearly. Recently, \cite{Krehlik2023} broke down the minimum required accuracy for HD maps for both static and dynamic models according to the vehicle geometry, and in the baseline model for the static case, the required accuracy is 32 cm. In 2022, \cite{Li2022} published a paper that started the trend in HD map online mapping development as it became the pioneer in evaluating HD map generation results, cementing the path towards comparison between learning methods. They propose both semantic and instance metrics to evaluate the performance of the HD map mapping algorithm. \cite{Zhang2023b} proposes a rasterization evaluation metric to detect deviations of vectorized-based HD maps. They tailor the rasterization model based on various geometric shapes that are effectively applicable to a wide range of map elements.

\subsection{Data Acquisition}

\begin{figure}[h]
    \centering
    \includegraphics[width=0.9 \linewidth]{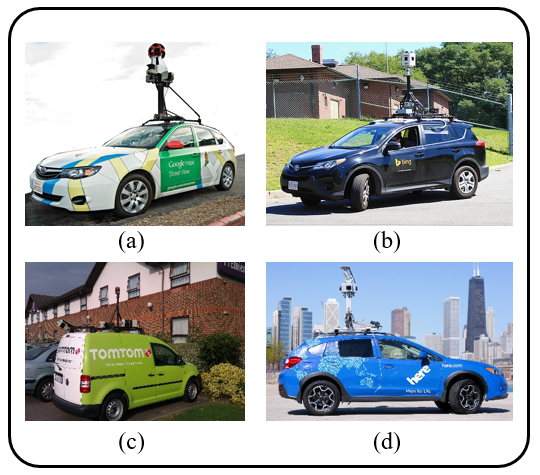}
    \caption{The mobile mapping vehicle of giant mapping companies: (a) Google. (b) Bing. (c) TomTom. (d) Here. }
    \label{fig:mapping_veh}
\end{figure}

The accuracy of High-Definition (HD) maps is attributed to the use of sophisticated equipment in their data acquisition process. As depicted in Fig. \ref{fig:mapping_veh}, the state-of-the-art approach involves deploying these mobile mapping vehicles for data collection. These vehicles are usually equipped with high-end mapping sensors, which include GNSS-RTK, IMU, LiDAR, 360 cameras, and long-range and millimeter radar. 

The majority of mapping companies, including HERE, which plans to produce highly automated driving (HAD) maps from mapping vehicles, use this strategy to approach the HD map mapping problem. Other companies, such as Lyft Lvl 5, have collected 1000 hours of data from 20 fleets of autonomous vehicles, capturing 26.000 km of suburban route scenario \cite{Houston2020}. This massive amount of data is usually computed offline on the server and stored in the cloud, allowing cloud services to store and share the mapping information. Although this collection method is the best, it poses a practical problem with the resources required to supply such a massive fleet of expensive mapping vehicles, which poses an industrial challenge in using this collection method. 
In order to address this, researchers are looking into the potential of crowdsourced data acquisition, in which a "common" intelligent vehicle with fundamental sensors gathers the data. Reference \cite{Massow2016} is notable for being the first to execute crowdsourced mapping of HD maps through vehicle GPS trajectories. Now, the crowdsourced approach is one of the hot topics in this research direction afterward \cite{
Kim2018,Liang2018,Liebner2019,Pannen2019,Liu2019,Pannen2020,Chao2020,Li2020,Qi2021,Kimlidar2021,Kimlidar2021,Zhanabatyrova2023,Xue}. \cite{Gao2017,Gao2017a} proposes a trust-based recruitment framework for crowdsourced vehicles. They were among the first to consider this problem in this domain. They also explore the learning-based recruitment system in the next iteration \cite{Gao2020}. Based on the crowdsourced vehicle's trajectory, they distribute rewards. Cao et al.,\cite{Cao2019} explore the selection criteria of the crowdsourced vehicle, which is defined as workers. They model the problem into a classic multi-armed bandit (MAB) process, aiming to achieve the highest quality of mapping results. Worker's attributes such as trajectory, crowdsourcing budget (cost), marginal utility (total of map elements observed), and platform utility (number of hours available) are proposed. 
Besides relying entirely on the vehicles to collect the mapping data, some researchers also use roadside sensors such as camera and LiDAR \cite{Liu2023} to perform data collection for HD map creation and update.
To sum up, the challenge in data collection for HD mapping in autonomous driving lies in balancing coverage, update frequency, and data reliability. Centralized data collection offers extensive coverage but often struggles with frequent updates. Crowdsourced data acquisition, on the other hand, provides rapid and frequent map updates, a critical factor for dynamic driving environments. However, this method's reliability is a major concern, primarily due to the varying accuracy of on-board vehicle sensors. The inconsistent data acquisition times and presentation across different datasets further complicate the task of determining accurate and timely updates. This intricate balance between data reliability, update frequency, and coverage is essential for the effective use of HD maps in autonomous driving systems and creates a new paradigm of problem that considers the recruitment process that is able to balance the disadvantages of this approach.

\section{Data Preprocessing Module}
\begin{figure*}[h]
    \centering
    \includegraphics[width=1 \linewidth]{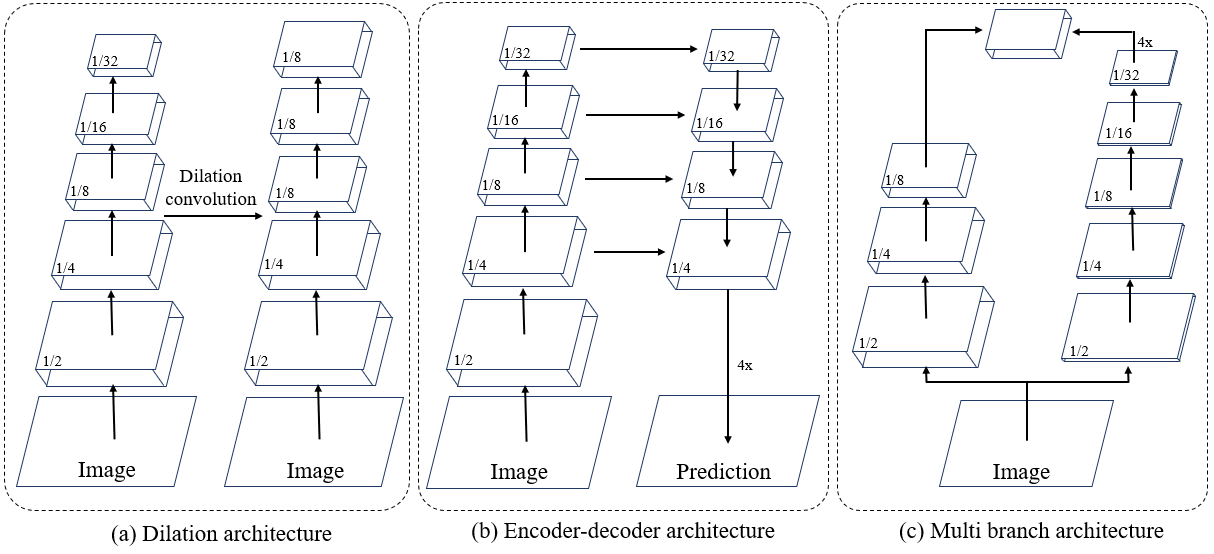}
    \caption{The architecture of image segmentation networks: (a) Dilation architecture. (b) Encoder-decoder architecture. (c) Multi branch architecture. Readers interested in the comprehensive descriptions of each architecture are advised to refer to the paper individually. Figure is redrawn and modified based on depictions in \cite{Yu2021}}
    \label{fig:architecture_1}
\end{figure*}

This section presents the data preprocessing algorithm for making and maintaining HD maps, specifically semantic segmentation and localization tasks. Given vehicles' raw data, this step is required to transform the data into information required and yet light in computation. Therefore, it is a necessity to have algorithms that can be robust and perform in real-time applications. The result of this process is then processed in the mapping module and update module subsequently as shown in Fig. \ref{fig:pipeline}. This section will briefly introduce the development, trend, and state-of-the-art method of each task.

\subsection{Semantic segmentation}
It is always desired to have a robust and accurate module for detecting objects in 2D and 3D point cloud space for mapping tasks in intelligent vehicles. Several sensors are responsible for these tasks. Among them are LIDAR, Camera, Radar. Below, we will briefly explain the development of the algorithm on each of these sensors and the multi-sensor fusion that supports the advancement in high definition mapping and update in general. In this section, the multi-sensor fusion is also briefly described in the LIDAR and Radar section, where LIDAR-vision and Radar-vision are essential in developing semantic segmentation in these segments.

\subsubsection{Camera}
Object detection tasks have been popularized by the development of convolutional neural networks (CNN) as a significant breakthrough in machine learning in general \cite{Krizhevsky2012}. This technology then quickly adapted to the application for scene understanding or perception in intelligent vehicle \cite{Geiger2012} \cite{Cordts2016}\cite{Zhou2019}. Recently, semantic segmentation has offered an essential role in the perception area of the intelligent vehicle because of the complete information it offers. It can also support the development of local mapping in the intelligent vehicle by detecting static features such as lane boundaries, road markings, traffic signs, and moving objects such as vehicles. These targets can be recognized with a relatively high IOU (Intersection over Union). IOU is a commonly used measure for determining how accurate a proposed image segmentation is when compared to a ground-truth segmentation. The idea is to assign semantic labels to each pixel, and with the development of deep learning, neural networks can achieve a very good performance in semantic segmentation tasks. \cite{Long2015} proposed the fully convolutional neural network (FCNN) for the first time, which realized the end-to-end image segmentation. 

\begin{table}[h]
\caption{Segmentation IoU (\%) on the CamVid \cite{Brostow2009} test where the model is trained on ImageNet \cite{JiaDeng2009} and Cityscapes \cite{Cordts2016} datasets.}
\label{tab:loc}
\centering
\def\arraystretch{1.3}
\setlength\tabcolsep{4pt} 
\begin{tabular}{clcccccccccccccc}
\cline{1-5}
& \multicolumn{1}{c}{\textbf{Method}}& $\text{Architecture}$& $\text{mIoU}$ & $\text{FPS}$  \\ \hline
\multirow{4}{*}{\thead{\textbf{Large}\\ \textbf{model}}}
& SegNet~\cite{Badrinarayanan2017} &Encoder-decoder & 60.1 & 4.6  \\
& Deeplab~\cite{Chen2018}&Dilation & 61.6 & 4.9 \\
& Dilation8~\cite{Yu2016}&Dilation & 65.3 & 4.4  \\
& PSPNet~\cite{Zhao2017}&Dilation & 69.1 & 5.4  \\ \hline

\multirow{9}{*}{\thead{\textbf{Lightweight}\\ \textbf{model} }} 
& ENet~\cite{Paszke2016} & Dilation &  51.3 & 61.2  \\
& DFANet A~\cite{Li2019}& Encoder-decoder & 64.7 & 120 \\
& DFANet B~\cite{Li2019}& Encoder-decoder & 59.3 & 160 \\
& ICNet~\cite{Zhao2018}& Multi branch &  67.1 & 27.8 \\
& BiSeNetV1~\cite{Yu2018}& Multi branch &  65.6 & 175  \\
& BiSeNetV2~\cite{Yu2021}& Multi branch & 76.7 & 124.5  \\
& DDRNet-23~\cite{Pan2022}& Multi branch & 78.6 & 182.4  \\
& PIDNet~\cite{Xu2022}& Multi branch & 80.1 & 153.7  \\

\hline
\end{tabular}
\end{table}

\begin{table*}[t]
\caption{Segmentation IoU (\%) on the KITTISemantic test \cite{Behley2019} datasets.}
\label{tab:loc}
\centering
\def\arraystretch{1.4}
\setlength\tabcolsep{4pt} 
\begin{tabular}{clcccccccccccccc}
\cline{1-9}
& \multicolumn{1}{c}{\textbf{Method}} & $\text{Road}$ & $\text{Parking}$ & $\text{Sidewalk}$ & $\text{Other-ground}$ & $\text{Fence}$ & $\text{Pole}$ & $\text{Traffic-sign}$  \\ \hline
\multirow{4}{*}{\thead{\textbf{2D semantic } \\ \textbf{segmentation}}}
& SqueezeSegV3~\cite{Xu2020} & 91.7& 63.4 & 74.8 & 26.4 & 59.4 & 49.6 & 58.9  \\
& RangeNet++~\cite{Milioto2019} & \textbf{91.8} & \textbf{65.0} & 75.2 & 27.8 & 58.6 & 47.9 & 55.9 \\
& PolarNet~\cite{Zhang2020} & 90.8 & 61.7 & 74.4 & 21.7 & 61.3 & 51.8 & 57.5 \\
& SalsaNext~\cite{Cortinhal2020} & 91.7 & 63.7 & \textbf{75.8} & \textbf{29.1} & \textbf{64.2} & \textbf{54.3} & \textbf{62.1} \\ \hline

\multirow{3}{*}{\thead{\textbf{3D semantic } \\ \textbf{segmentation}}} 
& RandLA-Net~\cite{Hu2020} & 90.7 & 60.3 & \textbf{73.7} & 20.4 & 56.3 & 49.2 & 47.7 \\
& KPConv~\cite{Thomas2019} & 88.8 & 61.3 & 72.7 & \textbf{31.6} & \textbf{64.2} & 56.4 & 47.4 \\
& MinkNet~\cite{Choy2019} &  \textbf{91.1} & \textbf{63.8} & 69.7 & 29.3 & 57.1 & \textbf{57.3} & \textbf{60.1} \\ \hline

\multirow{5}{*}{\thead{\textbf{Hybrid} \\ \textbf{segmentation}}} 
& FusionNet~\cite{Zhanglidar2020} & \textbf{91.8} & \textbf{68.8} & \textbf{77.1} & 30.8 & 69.4 & 60.4 & 66.5 \\
& 3D-MiniNet~\cite{Alonso2020} & 91.6 & 64.2 & 74.5 & 25.4 & 60.8 & 48.0 & 56.6 \\
& SPVNAS~\cite{Tang2020} & 90.2 & 67.6  & 75.4 & 21.8 & 66.9 & 64.3 & 67.3 \\
& Real-time network~\cite{Xie2022} & 91.2 & 34.1  & 74.8 & 0 & 29.2 & 40.4 & 29.6 \\
& (AF)\textsuperscript{2}-S3Net~\cite{Cheng2021} &  91.3 & \textbf{68.8} & 72.5 & \textbf{53.5} & 63.2 & 61.5 & \textbf{71.0}\\ 
& 2DPASS~\cite{Yan2022} &  89.7 & 67.4 & 74.7 & 40.0 & \textbf{72.9} & \textbf{65.0} & 70.4\\ 
\hline
\end{tabular}
\vspace{1ex}

{\centering *Note: All of these evaluations are done using KITTISemantic \cite{Behley2019} datasets. \par}
\end{table*}

There are three types of architecture in the backbone framework of the image segmentation network, dilation network, encoder-decoder network, and multi-branch network as shown in Fig. \ref{fig:architecture_1}. Dilation network omits the downsampling operations and keep the upsampling operation which maintains the feature representation in high-resolution \cite{Chen2015}\cite{Chen2018b} \cite{Zhao2017} \cite{Zhao2018b}\cite{Yu2020}. Encoder-decoder network provides the recovery of the high-resolution feature representation in the decoder part after the compression process in the encoder part \cite{Lin2017} \cite{Peng2017} \cite{Yu2018}. These architectures are designed specifically to provide the best semantic segmentation solution with no regard for computational cost and inference speed. When the researcher tries to apply these networks in real-life applications, these become a problem. Therefore, more and more people have started to modify this architecture to simplify the calculation process to accelerate the computational time. Starting from restricting the input data  \cite{Zhao2018}\cite{Mazzini2018}\cite{Romera2018}) and pruning the channels in the early stage of the calculation to directly improve the inference speed \cite{Badrinarayanan2017} \cite{Paszke2016} \cite{Chollet2017}). DFANet \cite{Li2019} recycles the features to simplify and improve the feature representation process. Although these methods can satisfy the real-time requirement for image segmentation tasks, the reduction in size and channels create a large deterioration in the accuracy. Recently, researchers have adopted a new approach by using two/multi-branch architecture. ICNet \cite{Zhao2018} was the first to propose a multi-branch architecture with three branches of different depths to process various resolution inputs to achieve real-time semantic segmentation. Fast-SCNN \cite{Poudel2019} follows suit with two branch of architecture philosophy. It learns to down-sample the input and processes it afterward. BiSeNet \cite{Yu2018} \cite{Yu2021}  proposed Bilateral Segmentation Network where it has two branches, a detailed branch, and a semantic branch as shown in Fig. \ref{fig:architecture_1}. Detail branches are designed to capture spatial details with wide channels and shallow layers. On the other hand, the semantic branch has smaller channels with deeper layers to capture the semantic context. 
\cite{Pan2022} is the first to propose a deep high-resolution representation into real-time semantic segmentation rich in context information. It consists of residual and bottleneck blocks that provide a speed-to-accuracy trade-off when the depth and width are scaled. \cite{Xu2022} created a bridge between PID controllers designed for controllers to the image segmentation task, which inspired the additional branch. They can achieve state-of-the-art results in real-time semantic segmentation tasks by utilizing boundary prediction to ensure precise annotation around the semantic context.
Readers who are interested in the most recent survey about deep learning algorithms in this subject can refer to \cite{Minaee2021}.

\subsubsection{LIDAR}

Image semantic segmentation is very critical in the map update process. However, the sensors' limitations to record data in poor lighting conditions, lack of depth information, and limited scan areas make it difficult to rely entirely on vision-based sensors as the only source of perception information. In contrast, Light Detection and Ranging (LIDAR) sensors can provide a reliable source of depth information regardless of the lighting conditions with a high-frequency flow of data which gives itself an edge by comparison. 
However, \cite{Tsushima2020} also elaborates the limitation from LIDAR sensor caused by low point density and coarseness at the time of measurement due to insufficient light irradiation. 

There are a lot of research have been done on this space with the introduction of the public dataset such as KITTISemantic \cite{Behley2019}, RELLIS-3D \cite{Jiang2021}, and TheSemanticKITTI \cite{Garbade2021} LIDAR data often comes in an unstructured format and varying in sparsity according to the object distance relative to the sensors. These characteristics pose the main challenge of fully utilizing the sensor data as the primary sensor to provide semantic segmentation results. However, more and more research has been done to address these problems. The earliest works in this space rely on the projection of the detected object from the images to the 3D point cloud space to achieve 3D object detection, and this method is also referred to as 2D semantic segmentation for LIDAR sensor \cite{Wu2018}. This method is based on an encoder-decoder network based on a fully convolutional neural network (FCNN) and recurrent neural network (RNN) layer. Then, they further refine the method in \cite{Wu2019} and \cite{Xu2020} to improve the loss function and batch normalization model. RangeNet++ \cite{Milioto2019} derived from DarkNet backbone of YOLOv3\cite{Redmon2018} to provide an efficient way of predicting segmentation results by fast K-nearest neighborhood (KNN) algorithm on the point cloud. PolarNet \cite{Zhang2020} introduces a new approach of using a polar bird eye view (BEV). This polar grid provides data-driven features using PointNet \cite{Qi2017} rather than manually designed features. Then, SalsaNet \cite{Aksoy2020} uses ResNet blocks as the encoder part and upsamples the features in the decoder part. They also use the BEV approach in their method, and it is extended in SalsaNext \cite{Cortinhal2020} by proposing a new, improved encoder-decoder to achieve state-of-the-art results in 2D semantic segmentation*. 

Then, it develops into a direct feed of point clouds into the 3D convolutional network where voxel representation is used to perform 3D convolutions \cite{Maturana2015}\cite{Chang2015} \cite{Wang2020}. PointNet \cite{Qi2017} and PointNet++ \cite{Qi2017a} introduces sampling of different point cloud scales to extract features. This method is particularly slow when processing a large point cloud data. RandLA-Net \cite{Hu2020} downsamples features randomly to accelerate the calculation process and at the same time introduce a local feature aggregation module to increase the receptive field for each 3D point. KPConv \cite{Thomas2019} introduced a new way to process the point without any pre-processing step directly. Then, MinkNet \cite{Choy2019} introduces a novel 4D convolution with an open-source code to differentiate sparse tensors automatically. This method can achieve state-of-the-art results compared to other 3D semantic segmentation methods*.

Finally, the hybrid methods where voxel-based, image-lidar projection, and/or point-wise based operations are utilized to process the point cloud. This method is not standard in the past because of memory limitations. However, with the development of a memory-efficient algorithm, this method can provide a meaningful result. FusionNet \cite{Zhanglidar2020} uses voxel-based mini-PointNet, which directly projects features from neighborhood voxel to the target voxel, which results in an efficient calculation to process extensive scale point cloud data. Then 3D-MiniNet \cite{Alonso2020} uses a projection method based on a learning algorithm to extract features from 3D data then feeds it to 2D FCNN to predict semantic segmentation results. SPVNAS \cite{Tang2020} derived from MinkNet library \cite{Choy2019} can propose a hybrid 4D sparse convolution and point-wise based operations to achieve a tremendous semantic segmentation result. Finally, (AF)\textsuperscript{2}-S3Net \cite{Cheng2021} which also built upon MinkNet \cite{Choy2019} model can be converted into an end-to-end encoder-decoder with the addition of attention layers to achieve state-of-the-art results when compared to other hybrid methods*. In the latest development, \cite{Xie2022} can achieve a comparable result in some areas such as road and sidewalk detection with fewer parameters which speeds up the processing time by 2.17 times. 
\cite{Yan2022} ables to boost the performance of LiDAR point cloud semantic segmentation by leveraging an auxiliary modal fusion which is rich in semantic and structural information and knowledge distillation.
Readers who are interested in the most recent survey about LIDAR semantic segmentation dataset and methods refer to \cite{Gao2021}.

\subsubsection{Radar}
Recent researches have progressively utilized various methods to improve object detection, as well as classification based on MMW radar data \cite{Lombacher2016}\cite{Schumann2018}\cite{Prophet2019}. Since the sparsity of the points provided from radar data, it is a very challenging problem to realize object detection and classification. Scientists approach this problem by accumulating radar data from multiple frames as radar-based grid maps. This accumulation provides rich points that improve the detection result to a certain degree. This collection of data is then fed to the segmentation networks \cite{Prophet2019}. Similar to image processing, a convolutional neural network is also used in radar segmentation problem \cite{Lombacher2016}\cite{Lombacher2017}. Then, a radar grid map is used to classify static map elements and recognize the orientation of each element represented in a grid format. 

\begin{figure}[h]
    \centering
    \includegraphics[width=1 \linewidth]{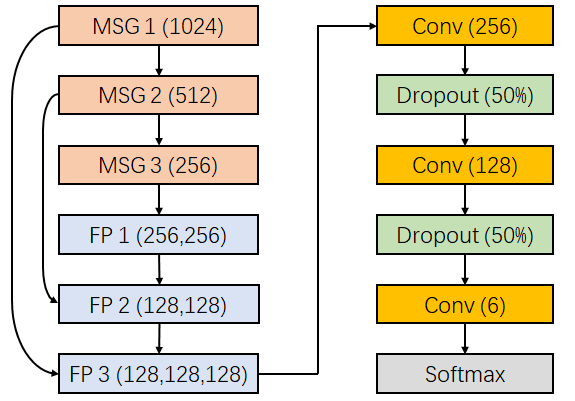}
    \caption{Radar semantic segmentation by PointNett++ \cite{Qi2017a}. Figure is redrawn and modified based on depictions in \cite{Zhou2020}.}
    \label{fig:segmentation_1}
\end{figure} 

There is another method that directly relies on the deep learning algorithm to process radar data. This approach is similar to LIDAR segmentation and usually uses neural network such as PointNet++ \cite{Qi2017a}. The network first modified to suits the density and sampling rate of radar data \cite{Schumann2018}\cite{Braun2021}. The data processing framework of radar semantic segmentation is showed in Fig. \ref{fig:segmentation_1}. Besides, CNN, RNN network LTSM (Long-Short-Term Memory) is also used to classify static and dynamic traffic elements \cite{Pohlen2017}\cite{Scheiner2018}. 
\cite{Ouaknine2021} proposes a lightweight architecture by utilizing multi-view radar to detect and localize moving objects and their method is able to determine their velocity.
Readers who are interested in the most recent survey about deep learning method in millimeter-wave radar semantic segmentation can refer to \cite{Zhou2020}.

\subsubsection{Discussion}
To summarize, the overview of the advancement of the semantic segmentation technique has shown an encouraging trend towards obtaining accurate semantic information from various sensor equipment. Image domain as the forefront of semantic segmentation has always provided the best accuracy in real-time performance. However, the reliability factor of the camera in terms of outside factors, such as lighting and weather, cripple its robustness. LIDAR and Radar approaches offer a very attractive solution to lighting problems, as these sensors do not affect the change of night and day. However, these methods rely heavily on images on the training of its algorithm, as the state-of-the-art method for these sensors requires a 2D image domain for training input. Given the present trend of LIDAR and Radar approaches, the bottleneck will soon be on the image semantic segmentation algorithm. However, to reach this stage, more and more research is required. Nevertheless, semantic segmentation research in the image domain is also encouraged to ensure faster computational capabilities with more accurate segmentation results. Readers interested in the most recent survey about map element extraction method can refer to \cite{Bao2022}.

\subsection{Localization}

\begin{table*}[t]
\caption{List of Visual Localization Methods}
\label{tab:loc}
\centering
\def\arraystretch{1.4}
\setlength\tabcolsep{4pt} 
\begin{tabular}{cccccccc}
\cline{1-8}
\textbf{Method} & $\text{Year}$& $\text{Mono}$ & $\text{Stereo}$ & $\text{IMU}$ & $\text{Accuracy}$ & $\text{VO/VIO/SLAM}$ & $\text{Open Source}$  \\ \hline

Mono-SLAM\cite{Davison2007} &2007 &\checkmark & &  &Adequate  &SLAM  & \cite{Kim_mono2015}    \\
LSD-SLAM\cite{Engel2013a}\cite{Engel2015}\cite{Engel2018} &2018 &\checkmark &\checkmark & & Good &SLAM &\cite{Engel2018}\\
SVO \cite{Forster2014}\cite{Forster2016} &2016 &\checkmark  &\checkmark  &  & Excellent &VO/VIO  &\cite{Forster2014}  \\ 
MSCKF\cite{Mourikis2007}\cite{Zhu2017} &2017 &\checkmark  &  &\checkmark &Adequate &VO/VIO &\cite{Zhu2017}  \\ 
DSM\cite{Zubizarreta2019} &2019 &\checkmark  &  &  & Excellent  &SLAM  &\cite{Zubizarreta2019}  \\ 
OKVIS\cite{Leutenegger2015} &2015 &\checkmark  &\checkmark  &\checkmark  &Good  &VO/VIO  &\cite{Leutenegger2015}  \\ 
VINS-Fusion\cite{Qin2018a}\cite{Qin2018} &2018 &\checkmark  &\checkmark  &\checkmark  & Good  &VO/VIO  &\cite{Qin2018}  \\ 
OpenVINS\cite{Geneva2020} &2020 &\checkmark  &\checkmark  &\checkmark   &Excellent &VO/VIO  &\cite{Geneva2020} \\ 
ORB-SLAM\cite{Mur-Artal2015} \cite{Mur-Artal2017} \cite{Mur-Artal2017a} \cite{Campos2021} &2021 &\checkmark  &\checkmark  &\checkmark  &Outstanding   &SLAM &\cite{Campos2021}  \\ 
SOFT2\cite{Cvisic2015} \cite{Cvisic2018} \cite{Cvisic2022} \cite{Cvisic2023}  &2023 &  &\checkmark  &\checkmark  &Outstanding   &VO/VIOO &\cite{Cvisic2023}  \\ \hline

\end{tabular}
\end{table*}
Localization is one of the major subsystems of autonomous vehicles. Currently, the primary sensors used for localization on vehicles are GNSS, IMU, cameras. LIDAR and HD Map. In principle, vehicle localization held the key to the accuracy of the map created by the intelligent vehicle. Thus, making this process is an integral part of the architecture. Below, we will briefly explain the development of the algorithm on these sensors that support the advancement in high definition map building and update in general. This section divides the camera and LIDAR sections into odometry and map matching methods. Readers can refer to \cite{Lu2021} for the more complete and most recent real-time performance localization techniques survey.




\subsubsection{GNSS and GNSS-IMU}
GNSS is a prevalent method to provide absolute localization solutions for a vehicle. However, its reliability can only be maintained in open areas as it is often affected by NLOS, signal block, or multipath problem \cite{Qin2017} \cite{Brebler2016}. The current trend is to combine GPS data with other measurements from other resources, including IMU, visual odometry, LIDAR odometry, and HD map. The current standard approach try to improve the accuracy and reliability by correction method, including filtering \cite{Park2017}, fusion \cite{Schreiber2016}, and map matching \cite{Amini2014}. \cite{Park2017} proposes an abnormal signal discerning framework to improve the robustness for GNSS-based localization. \cite{Adjrad2017} is able to improve the localization accuracy by removing periodic signals and aiding the height information obtained from a digital map. \cite{Kumar2014} can enhance the accuracy by analyzing the NLOS signal delays. In the recent development of GNSS, the real-time kinematic (RTK) technology can even reach centimeter-level of accuracy \cite{Jackson2018}. This technology requires the antennas to be calibrated precisely to receive the signals transferred from GNSS satellites then the correction will be performed from the data received from base stations where its location is known. The problem with this method is the high cost of the sensor, making it unsuitable for general application. Therefore, another approach to making the GNSS localization solution viable is adding an IMU sensor. This sensor can provide information such as acceleration pitch rate and has strong robustness towards interference\cite{Wang2016}. It complements GNSS wonderfully, as it can guarantee continuous localization when GNSS data is interrupted\cite{Azree2018}. \cite{Belhajem2018} proposed a machine-learning algorithm to compensate for the deviations of IMU data during GNSS failure. This method can achieve meter-level positioning accuracy. \cite{Ndjeng2009} proposed a dead reckoning (DR) based multiple interactive models (IMM), which improves the accuracy and integrity of vehicle localization when GNSS data experiencing signal block.\cite{Gim2018} proposed a pattern recognition of pitch rate generated from IMU sensor to calculate vehicle localization. This method matches the pattern of vibration and vehicle motion with the pre-built map for position estimation. This localization can achieve meter-level accuracy. Along with the development of other sensors such as LIDAR, camera, and radar. GNSS sensor becomes the initialization sensor where its absolute localization is used and refined according to the sensor used, explained in the next section.

\subsubsection{Visual Odometry, Visual Inertial Odometry, and Visual SLAM}

Visual SLAM approaches such as filtering and batch optimization are standard methods to solve the visual odometry problem. By using subsequent image frames, the probability distribution over all states can be updated, and finally, the filtering method can estimate the motion estimation of the camera. MonoSLAM \cite{Davison2007} is one of the first methods which can achieve the real-time requirement for motion estimation. This method exploits the sparse feature-based map and extended Kalman filter (EKF) framework. \cite{Civera2008} extended the previous work, and later they added RANSAC algorithm \cite{Fischler1981} to remove feature outliers \cite{Civera2010}. Other method adopting batch optimization deal with the problem by iteratively looking for maximum posteriori estimation (MAP), and it is normally solved by bundle adjustment (BA) algorithm. In general, this method requires an abundance of computational power. \cite{Strasdat2010} is able to reduce the complexity and at the same time outperform the filtering approach's accuracy and efficiency. In the next iteration, \cite{Strasdat2011} proposes a sliding window with a BA approach and also adds loop closure which improves the localization accuracy. Then, \cite{Mur-Artal2015} \cite{Mur-Artal2017} proposed a new method called ORBSLAM and ORBSLAM2, which is the first SLAM algorithm that can work in real-time and provide quick and accurate localization. They also extended the algorithm by adding an inertial sensor to ensure zero drift localization \cite{Mur-Artal2017a}.
\cite{Cvisic2015} \cite{Cvisic2018} proposed SOFT, a novel algorithm to perform robust and fast visual odometry through feature detection and tracking. In the next iteration called SOFT2, \cite{Cvisic2022} \cite{Cvisic2023} introduce a multiple hyphothesis perspective correction (MHPC) that performs perspective correction.
Another method called the direct approach is also proposed by \cite{Newcombe2011} which relies on dense visual information. \cite{Engel2013a} utilizes semi-dense information and further reduces the requirements into sparse points, which improves the calculation dramatically from needing GPU to CPU-only for real-time application \cite{Engel2018}. 

Similar to Visual SLAM, extending the visual odometry using an inertial sensor, the filtering-based method such as kalman-filter is also implemented in \cite{Cvisic2015} to perform outlier rejection and a sliding window approach is also used to improve the relative motion accuracy in \cite{Mourikis2007}. Batch optimization is commonly used in VIO implementation, such as \cite{Leutenegger2015}\cite{Forster2016}\cite{Leutenegger2015} \cite{Cvisic2023} supplement inertial sensor data between keyframes as the constraints and solve it using graph optimization problem. \cite{Forster2016} use the inertial constraints in the early initialization stage. The integration of visual motion estimation remains open for improvement, especially for mapping purposes. The localization accuracy should achieve a similar level in accuracy to an HD map.

\subsubsection{LIDAR Odometry, LIDAR Inertial Odometry, and LIDAR SLAM}
The LIDAR odometry can be estimated from the successive point cloud scans over time, like visual odometry. There are three common methods to estimate LIDAR odometry, 3D registration based (dense), 3D feature based (sparse), and 3D deep learning method. The first method relies on all the point cloud data, which means the calculation burden is very high, and the real-time capability is hard to be achieved \cite{Tam2013}. The classic method to perform point association is iterative closest point (ICP)\cite{Zhang1994}, including point-line ICP \cite{Censi2008}, point-plane ICP \cite{Low2004}, and generalized ICP \cite{Segal2009}. \cite{Mendes2016} proposes loop closure algorithm together with ICP and pose graph process to reduce the drift caused by consecutive registration. \cite{Kuramachi2015} uses IMU data to compensate for bad initial guesses to achieve an accurate localization. \cite{Kovalenko2019} proposed downsampling via normal covariance filter (NCF) method and outlier rejection via geometric correspondence rejector to achieve an accurate odometry result. \cite{Dellenbach2021} use an elastic trajectory, which allows pose continuity intrascan and discontinuity in between scans to make sure the robustness in high-frequency movements.  \cite{Palieri2021} proposed a health monitoring approach where it feeds on IMU, wheel inertial odometry (WIO), kinematics inertial odometry (KIO), and VIO to select the priority reference localization to supplement LIDAR odometry result. \cite{Chen2022} proposed a direct alignment of dense point clouds which is down-sampled to achieve computational tractability.
In 2014, LOAM \cite{Zhang2014} began to popularize the 3D feature-based method as it reached the top of the KITTI LIDAR odometry benchmark and stayed at the top for seven years. Features based method relies on the handcrafted features such as planes\cite{Ye2019}, lines\cite{Shan2020a} \cite{Pan2021}, edges\cite{Shan2020b}, and ground points\cite{Shan2018}. These works utilize the knowledge gained from VO techniques and translate it into the 3D domain. TVL-SLAM \cite{Chou2021} proposes an independent channel for visual and LIDAR frontend and optimizes the measurement results in the backend optimizations. \cite{Zheng2021} proposes a method to efficiently register point cloud by using non-ground spherical images and birds-eye-view maps to exclude ground points. 
\cite{Dellenbach2022} proposed a novel approach called continuous-time ICP (CT-ICP) with loop closure steps that can work in real-time mode. This method improves the precision and robustness in high-frequency motions by allowing elastic distortion in between scan registration.

Some researchers approached the problem by using a deep learning algorithm to solve the LIDAR odometry problem. \cite{Nicolai2016} is the first method to use the deep learning approach by transferring the 3D point cloud into the image domain and feeding the data into the network. The network will output the displacement and changes in orientation between two input frames. \cite{Wang2017} uses panoramic depth images as the representation of LIDAR data. LORAX \cite{Elbaz2017} introduces super-points which is a subset of points positioned inside of a sphere-like local surface. LocNet \cite{Yin2018} handcrafted a rotational invariant representation (RIR) which is generated from the ring distribution of the point clouds. \cite{Cho2019} proposed a spherical coordinate system to project the LIDAR frames to 2D representation. Deep CLR \cite{Horn2020} proposed a new architecture that applies flow embedding to generate features that describe the motions of each feature point. \cite{Adis2021} extends the previous network by adding a plane point extraction. This increases the computational time by reducing the point cloud size by up to 40\% - 50\%. Readers who are interested in the most recent survey about odometry, including vision-based systems can refer to \cite{Mohamed2019}.

\subsubsection{Radar}
In contrast to LIDAR and vision-based localization, radar-based localization can quickly obtain real-time performance because of the low computing load of sparse points, thus making it memory efficient \cite{Scheiner2018}. In general, radar-based localization has a lower localization accuracy when compared to LIDAR or vision-based localization because semantic data provided by radar are not easily detected, and the points are relatively sparse.  \cite{Schuster2016} presented Cluster-SLAM where it utilizes particle filter and clustering method to perform SLAM localization. \cite{Scheiner2018} proposes to construct a radar map and use it to match with the radar image to obtain vehicle localization. This pipeline is similar to SLAM approaches in general. \cite{Yoneda2018} propose a probabilistic model for omnidirectional radar data and perform the test in the snow. Their approach can achieve an excellent accuracy result of 0.25 m. \cite{Holder2019} proposed a pose graph method and loop closure algorithm to solve the localization task. 

The current state-of-the-art in this space is given to \cite{Adolfsson2021} where it applies the filtering technique that selects the strongest k value per azimuth and filters the radar data to compute a set of oriented surface points for accurate scan matching. \cite{Burnett2021} proposed a combination of probabilistic trajectory estimation and keypoint features generated from deep learning networks. This approach achieves the state-of-the-art method \cite{Adolfsson2021} without needing to manually handcrafting features.

\subsubsection{Map matching based localization}

\begin{figure*}[h]
    \centering
    \includegraphics[width=1.0 \linewidth]{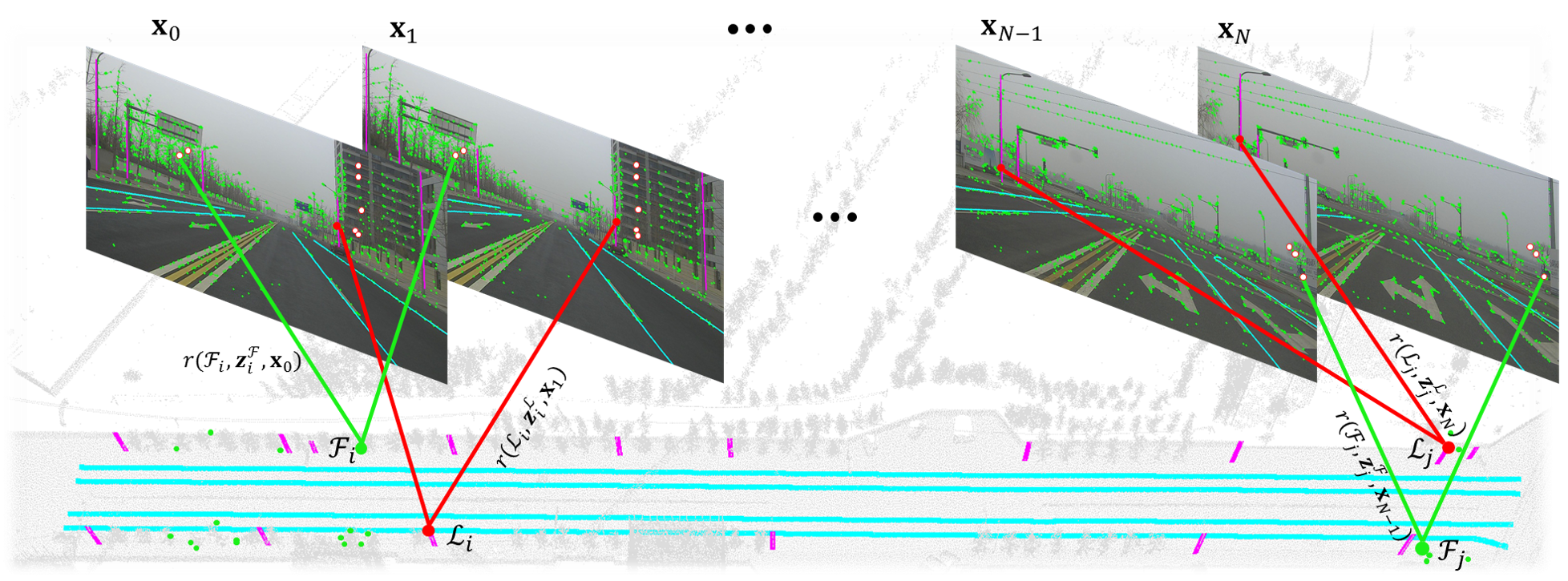}
    \caption{The illustration of tightly-coupled map matching method via joint sliding window optimization to estimate vehicle pose using feature points and vector HD map landmarks. Figure is obtained from \cite{Wen2020} work with permission from the authors.}
    \label{fig:map_matching}
\end{figure*} 

Map matching localization algorithm is one of the main purposes of providing an accurate HD map. It is tightly linked with the accuracy and reliability of HD maps for localization purposes. Map matching has been the subject of ongoing research, and it can be divided into two categories, online and offline modes. In online mode, the map matching procedure is performed on the go. Consequently, this approach puts emphasis on the calculation speed to achieve real-time performance.

On the other hand, offline modes emphasize accuracy as the map matching process is done after the trajectory is completed and time is no longer a constraint. This method can be viable for applications such as map building because of the importance of accurate localization compared to real-time performance. However, the online map-matching algorithm can fit this pipeline for change detection purposes, which is usually done online. Considering this brief introduction to map matching methods, readers who are interested in the most complete and most cited survey about this subject can refer to \cite{Quddus2007}\cite{Kubicka2018}. 

\begin{table}[h]
\caption{Performance List of Geometric Map Matching Method based on \cite{White2000}}
\label{tab:performance}
\centering
\def\arraystretch{1.4}
\setlength\tabcolsep{4pt} 
\begin{tabular}{cc}
\cline{1-2}
\textbf{Method} & $\text{Matching Accuracy}$ \\ \hline

Point-to-curve  &53\%-67\% \\
Point-to-curve + heading &66\%-85\% \\
Point-to-curve + route &66\%-85\%  \\
Curve-to-curve & 61\%-72\% \\ \hline
\end{tabular}
\end{table}

Geometric algorithms are the most common and early approach for map matching. There are three types of classification according to \cite{Bernstein1996}: point-to-point point-to-curve, curve-to-curve. In the end, the point-to-curve method leads to the development of the online map matching method, and curve-to curve leads to offline map matching method advancement. \cite{White2000} introduces four basic geometric methods: point-to-curve, point-to-curve with consideration of heading, point-to-curve with consideration of route topology, and curve-to-curve. The results are summarized in Table. \ref{tab:performance}.

\begin{table*}[t]
\caption{List of Map Matching Localization Methods}
\label{tab:loc}
\centering
\def\arraystretch{1.4}
\setlength\tabcolsep{4pt} 
\resizebox{\linewidth}{!}{
\begin{tabular}{ccccccccccccc}
\cline{1-13}
\textbf{Method} & $\text{Year}$& $\text{Methodological}$ & $\text{Map Types}$  & $\text{Online}$& $\text{Offline}$ & $\text{GNSS}$ & $\text{IMU}$ & $\text{Monocular}$ & $\text{Stereo}$ & $\text{LIDAR}$  & $\text{Other}$& $\text{Accuracy (m)}$  \\ \hline

\cite{Bernstein1996} &1996 &Geometric &Navigation map & &\checkmark &\checkmark & & & & & & n/a  \\
\cite{White2000} &2000 &Geometric &Navigation map & &\checkmark &\checkmark  & & & & & & n/a\\
\cite{Chen2011} &2011 &Geometric &Navigation map & &\checkmark &\checkmark  & & & & & & n/a\\
\cite{Schreiber2013} &2013 &Feature matching &Vector map &\checkmark & &\checkmark  &\checkmark & &\checkmark & & & n/a\\
\cite{Schindler2013} &2013 &Feature matching &Vector map & &\checkmark &\checkmark &\checkmark &\checkmark & &\checkmark & &1.00\\
\cite{Wolcott2014} &2014 &Feature matching &Dense point cloud map & &\checkmark &\checkmark &\checkmark &\checkmark & &\checkmark & &0.25\\
\cite{Quddus2015} &2015 &Geometric &Navigation map & &\checkmark &\checkmark  & & & & & & n/a\\
\cite{Caselitz2016} &2016 &Feature matching &Dense point cloud map & &\checkmark &\checkmark  &  &\checkmark & & & & 0.30\\
\cite{Werber2016} &2016 &Landmark matching &Navigation map & &\checkmark &\checkmark & & & & &MMW radar & n/a\\

\cite{Suhr2017} &2017 &Feature matching &Vector map &\checkmark & &\checkmark &\checkmark &\checkmark & & & Wheel encoder & 0.73\\
\cite{Li2017} &2017 &Particle filter &Navigation map & &\checkmark &\checkmark  & & & & & & 4.70\\
\cite{Li2018} &2018 &Multi hypotheses &Navigation map & &\checkmark &\checkmark  & & & & & &1.75\\
\cite{Ghallabi2018} &2018 &Feature matching & Vector map &\checkmark & &\checkmark &\checkmark & & &\checkmark & & 0.04\\
\cite{Stenborg2018} &2018 &Feature matching &Dense point cloud map & &\checkmark & &\checkmark & & & & &0.60\\
\cite{Xiao2018a} &2018 &Feature matching &Vector map & &\checkmark &\checkmark  & &\checkmark & & & & 0.35\\
\cite{Ma2019} &2019 &Feature matching &Lightweight vector map &\checkmark & &\checkmark  &\checkmark &\checkmark & &\checkmark & &0.51\\
\cite{MinKang2020} &2020 &Feature matching &Vector map &\checkmark & &\checkmark  & &\checkmark & & & & 0.48\\
\cite{Xiao2020} &2020 &Feature matching &Lightweight vector map &\checkmark & &\checkmark &\checkmark &\checkmark & & & & 0.24\\
\cite{Wen2020} &2020 &Feature matching &Lightweight vector map &\checkmark & &\checkmark &\checkmark &\checkmark & & & & 0.15\\

\cite{Wang2021} &2021 &Feature matching &Lightweight vector map & &\checkmark  &\checkmark & &\checkmark & & &Wheel encoder & 0.13\\

\cite{Jiang2021a} &2021 &Probabilistic model &Lightweight vector map & &\checkmark  &\checkmark & & & & & Microphone & 0.30\\

\cite{Zhang2022aa} &2022 &Neural network &Lightweight vector map &\checkmark  & &\checkmark &\checkmark &\checkmark & & &Wheel encoder & 0.13\\

\cite{Zhang2022} &2022 &End-end learning &Lightweight vector map &\checkmark &  &\checkmark &\checkmark & & & & Multi-view camera  &0.09\\
\cite{Kim2023} &2023 &Geometric &Lightweight vector map &\checkmark &  &\checkmark &\checkmark &\checkmark & & &  &3.69\\
\hline

\end{tabular}
}
\end{table*}

In the literature, distance metric to compare the points and curve earned much attention \cite{Chen2011}\cite{Quddus2015}. \cite{Li2017}\cite{Li2018} presented a lane map-matching based on filtering and multiple hypotheses algorithm with lane-level accuracy. However, these methods fall out of favor because of the increasing requirement in localization accuracy for specific tasks and the increasing number of map alternatives (lanelet maps \cite{Bender2014}, dense maps, and vectorized HD maps) that can provide an instrumental detail in terms of accuracy and completeness. In recent years, several researchers have investigated the idea of using sensors such as cameras, LIDAR, radar, and HD maps to have a very precise localization result \cite{Wolcott2014}\cite{Ma2019}\cite{Xiao2020}. \cite{Schreiber2013} introduces lane markings and curb matching to provide precise and robust online localization. This approach uses the Kalman Filter technique to refine the map matching localization result. Although the result obtained from this experiment of 0.07 m in accuracy is very promising, the accuracy referred to here is the mean residuals of map projection on the stereo image, not the actual localization accuracy. \cite{Ghallabi2018} presented an approach composed of many lane markings represented as polylines. \cite{Werber2016} proposes to use MMW radar to perform landmark matching. They analyze the landmark matching performance for quick recognition of point groups. In \cite{Suhr2017}, they propose to include all the lane markings detected in the map matching algorithm and also impose a particle filter to improve the localization accuracy further. \cite{MinKang2020} proposed a map matching method based on ICP based rigid map and achieved an error of 0.475 m. \cite{Wen2020} proposed a tightly coupled method to jointly optimize the vehicle pose through feature points generated from consecutive image frames as shown in Fig. \ref{fig:map_matching}. \cite{Feng2020} is the first to propose a deep learning algorithm to solve the map matching localization problem. They train the network to fit the localization given raw GNSS data only. 
Wang \cite{Wang2021} proposes a novel association method with sliding window factor graph optimization in urban roads. \cite{Zhang2022aa} reconstructs the local semantic map and then matches it to the vectorized map through the neural network in highway situations. Then, BEV-Locator \cite{Zhang2022}shows a large improvement in accuracy when their experiment reports mean absolute errors of 0.052m and 0.135 in lateral and longitudinal error. They are the first to create an end-to-end network, from extraction to localization results. The network predicts the optimal pose from a data-driven learning framework. They encode visual features by transforming surrounding images into Birds-Eye-View (BEV) space while the map data is encoded to form map queries. Their result remains the state-of-the-art method for performing map matching localization to the time this manuscript is written. Kim \cite{Kim2023} proposes a road shape classification method alongside robust map matching localization. This method constraints lane lines into three categories: straight line, circular arc, and clothoid curve. Although, laterally this method can achieve centimeter-level accuracy for real vehicle experimental results, however longitudinally, this approach is very inferior compared to the rest of the methods presented here. \cite{Jiang2021a} is the first to propose using sound localization and matching the vehicle trajectories with the HD map. This method can achieve 30 cm in localization accuracy.

\subsubsection{Discussion}
In summary, the analysis of localization approaches has shown the promising trend of obtaining accurate and yet low-cost AV localization, beginning from the reliance of GNSS sensor to provide the localization solution as a whole to fusion approach of using the camera, IMU, radar, and LIDAR. GNSS remains the core technique in these approaches as it can generate a quick location initialization for a more sophisticated technique. Besides GNSS, IMU and camera also become popular as they can compensate the GNSS problem of NLOS, multipath, and signal block with a small addition in cost. Furthermore, the rise of vision-based localization can be seen in tandem with the advancement of object detection and semantic segmentation. It allows the camera to be paired with an HD map to provide a very good localization accuracy with a very economical solution. The issue with camera localization is mainly the reliability of outside conditions such as weather, time (night), and sparse landmarks. Moreover, the lack of depth information in monocular setup also increases the challenge to provide a very accurate localization in general. The introduction of LIDAR based approach, which offers a very accurate depth estimation and is not sensitive towards time, becomes a promising localization solution. However, implementing a LIDAR sensor in a vehicle will tremendously increase the overall monetary and computational cost. This method still needs to address the challenges in scenarios with sparse vertical cues similar to the vision-based approach. Other ways of implementing radar, wheel encoder, and microphone are also provided in this section. These approaches serve as a viable proof of concept that can be explored; however, their localization accuracy and robustness remain an issue compared to the state-of-the-art method. Future research is still required to focus on improving accuracy, real-time application capabilities, and providing low-cost localization solutions.

\section{HD Map Building}
In order to generate the HD map used in intelligent vehicles, professional mapping vehicles equipped with state-of-the-art sensors for performing the above-mentioned tasks are mainly used through three processes \cite{KichunJo2014}. First, the mapping vehicle travels along target routes in order to acquire the mapping data (data preprocessing). Next, the features acquired from the mapping vehicle are accumulated based on the vehicle's trajectory on the map according to the types of the features (map building). Finally, the features in the map are refined and confirmed (map update). 
This section will first introduce the taxonomy and ontology of the HD map, which explain the works done in this aspect, then shift the focus on the mapping process of HD map, including the vectorization process of the map element detected in the previous section.

\subsection{Map Element Vectorization}

Several map representations have been proposed in these years. For more information regarding this topic, readers are suggested to refer to \cite{EbrahimiSoorchaei2022}.
In this subsection, we will emphasize the vector representation, which is the most common permanent static layer. There are three basic steps in performing the vectorization process of semantic data: The first step is noise filtering. Since semantic segmentation is not completely accurate, removing the noise generated from this process is necessary. Then, the points can be processed in the SLAM algorithm to determine the spatial position in a 3D coordinate. Moreover, the KD-Tree construction can be used to filter out outliers \cite{Greenspan2003}. 

\begin{figure}[h]
    \centering
    \includegraphics[width=0.9 \linewidth]{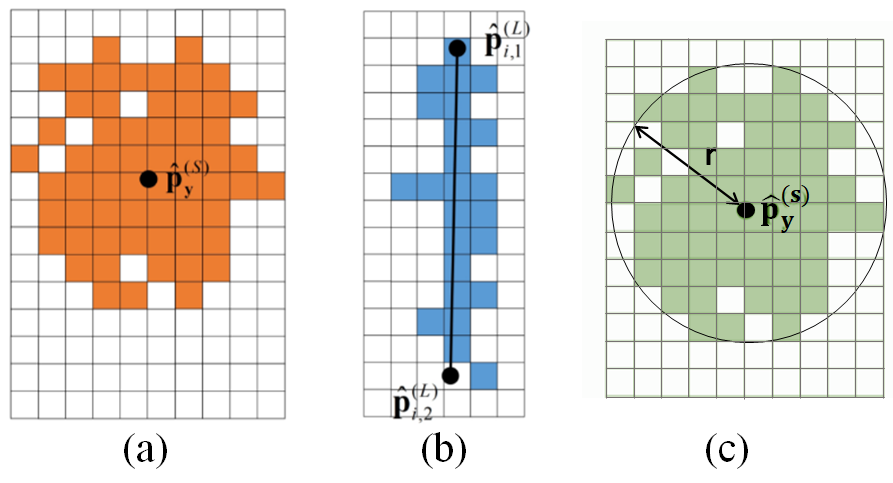}
    \caption{The illustration of vectorization of map feature element: (a) Point target. (b) Line target. (c) Plane target. Figure is redrawn and modified based on depictions in \cite{Xiao2018}}
    \label{fig:vectorized}
\end{figure} 

The next step is clustering, and there are several clustering approaches, such as euclidian cluster extraction \cite{Rusu2010} and RANSAC \cite{Fischler1981}. RANSAC algorithm can be used to perform these two tasks simultaneously for every object. There are three basic types of map feature element vectorization results as shown in Fig. \ref{fig:vectorized}. These types are matched according to the shape characteristic of the map element. In the traffic scene, the point target is the object whose geometric shape can be abstracted as a point in the image, such as traffic lights, traffic signs, road markings, and so on. A line target is an object whose geometry can be abstracted as a line segment, such as lamp post, lane line, etc. Because of the irregular shape of the lane line in the area of curve and intersection, it is not easy to model it with specific points and lines in the image coordinate system. In vector feature extraction, it is described as a broken line feature, and all image points are used as geometric description \cite{Xiao2020}. Plane objects are usually used to describe the traffic sign. Because of the varying shape of these map elements, the description can also vary from circular-plane shape to bounding box plane shape. Extending from the vectorization of the map element, data modeling of the HD map is also essential. In \cite{Poggenhans2018}, Lanelet2 HD map framework is proposed to meet the high demands of HD maps in terms of completeness, accuracy, verifiability, and extensibility. \cite{Kang2020} proposed a research-friendly HD map data model by extending the popular node-edge model that researchers widely use. Their model considers both on-road and off-road data to be extensible for future use cases and information available on various three-dimensional objects in the outdoor space. Their work extended the vectorized format by adding details to the data, such as intersection, edge, type, photo, latitude, and longitude. VAD \cite{Jiang2023} explores the fully vectorized representation
of the driving scene to incorporate this information for the path-planning task. Guo \cite{Guo2023} introduced a new framework for predicting trajectories in corner-case scenarios by leveraging scenario engineering technology. This method not only offers a bridge between simulation systems and real-life scenarios but also improves the reliability of trajectory prediction for autonomous vehicles in challenging scenarios. The motivation for these developments is partly due to several use cases that stand to gain from HD mapping data.

\subsection{Map Building Modules}
In general, map-building modules connect the semantic information obtained from images to 3D spatial points. This process involves 3D reconstruction and fusion algorithms for a larger area. The 3D reconstruction method involves geometric, graph-based, SLAM, and learning approaches. For interested readers who want to delve deep into the HD map generation technique may refer to the tutorial \cite{Jeong2022} for generating HD maps for automated driving in urban environments. The details for each of these methods are explained in the subsection below:

\subsubsection{Geometric approaches}
The problem of reconstructing a scene from two images was first proposed by \cite{longuet-higgins1987computer}. Then, \cite{schoenberger2016sfm, schoenberger2016mvs} popularize the multi-view geometric solutions, where multiple images are combined to reconstruct the key points of the images into a 3D representation. The COLMAP has benefited the community at large and has become the baseline for many researchers in this direction. In essence, the mapping accuracy of this approach correlates directly with the tracking algorithm's accuracy and the vehicle's pose accuracy.
In \cite{Dabeer2017}, an end-to-end approach toward traffic signs and traffic lane mapping is proposed. This approach utilizes consumer-grade sensors, such as front-facing monocular cameras and consumer-grade GPS. They use a triangulation algorithm and offline bundle adjustment to reconstruct the sign and lanes. They are able to achieve a relative error of 15 cm after 25 journeys. \cite{Golovnin2020} uses the simple projection matrix from the Samsung Galaxy S7 smartphone camera to generate HD maps and is able to achieve 5\% measurement error at a distance of 15 m from the camera.
In \cite{Paz2020}, semantic data obtained from a vehicle is directly transformed into its 3D point cloud counterparts. However, the resulting map is dense and unsuitable for real-time applications.
\cite{Chawla2020} is one of the first to propose the use of multi-view geometry to estimate the traffic sign positioning. Their strength lies in the learning-based self-calibration, depth, and motion estimation, which can facilitate the reconstruction of the traffic sign pose. The traffic sign maps achieve 1,26 m in absolute position accuracy, which is still far from the required accuracy of HD maps. \cite{Das2020} proposed an end-to-end mapping solution using vehicle odometry, GNSS, and stereo image data. They can estimate and get rid of temporal localization bias. They also show that a map made with a loop-closure detection algorithm can cut the biggest error in offset by 56.53\% and boost accuracy by 24.39\%. \cite{Wen2022} created a new data association algorithm with a depth-ranking strategy for tracking map elements that were found. They used motion flow and geometric consistency as the main similarity metrics to make sure the algorithm was correct. The mapping time of this method, from segmentation to reconstruction of roadside map elements, requires less than 400 ms, which is the fastest. \cite{Zhang2023a} is able to reconstruct short road segments by transforming the dense raw points into curvature-continuous clothoid-based paths with sparse parameters while maintaining accuracy under a given deviation limit. However, this method has not yet been successful in online mapping adaptation, and it remains a work for the future. \cite{Zhang2023} coupled the geometric representation with the learning approaches and is able to achieve state-of-the-art results. It is important to note that this paper is the first to harness the critical geometric properties of the attention-based learning framework.

\subsubsection{SLAM Mapping}

The objective of the mapping method is the long-term deployment of a map that can be shared among vehicles as a service \cite{Mooney2014}. In the previous section, SLAM approaches were mentioned, from vision-based to LIDAR-based. Those approaches regard the SLAM approach as necessary to obtain localization, making the mapping process secondary. However, this section will focus on the SLAM approaches, where the map is the goal of the algorithm's utilization. \cite{Thrun2006} is the first to use the GraphSLAM method to process the mapping data into large-scale mapping in an urban area. Their method recursively estimates the map element location based on the RLS algorithm. \cite{Grisetti2010} wrote the tutorial on using GraphSLAM, which further catapults the method's popularity. \cite{Soheilian2010} presents a 2D graph correspondence to perform an automatic approach to map road markings, crosswalks, and dashed lines in dense urban areas. They impose geometric specification constraints in the recognition and reconstruction processes. This approach uses stereo cameras installed on a mapping vehicle. \cite{Joshi2015} proposes a general approach of using coarse prior maps to generate accurate HD maps with the GraphSLAM technique and particle filtering algorithm. Their approach can generate lane lines with a mean accuracy of less than 10 cm in urban and highway scenarios. \cite{Ilci2020} uses professional-grade sensors (GNSS, IMU, LIDAR, and metric cameras) to perform the mapping. They can achieve cm-level absolute accuracy within the 50-meter range of the vehicle. 
\begin{figure*}[h]
    \centering
    \includegraphics[width=1 \linewidth]{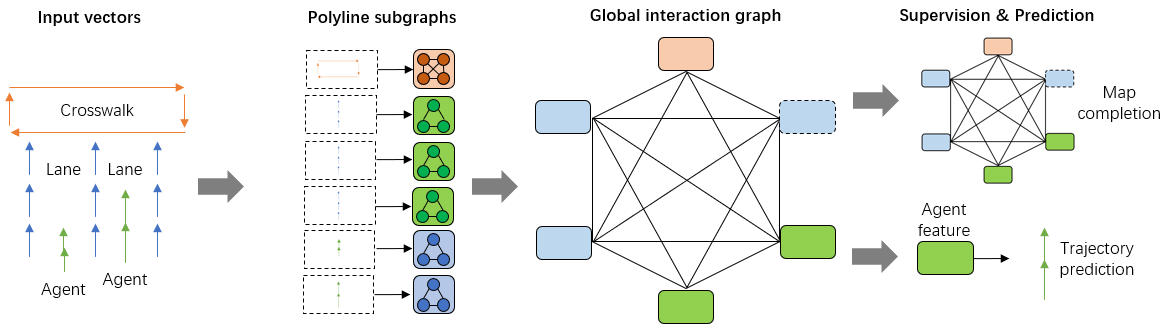}
    \caption{The VectorNet architecture, where the vectorized map information is passed through to a local graph network to obtain polyline-level features. They compute future trajectory losses from the node features corresponding to the moving agents and feature losses when the map features are masked out. Figure is redrawn and modified based on depictions in \cite{Gao2020a}}
    \label{fig:vector_rep}
\end{figure*} 

HDMapGen \cite{Mi2021} performs a coarse-to-fine HD map mapping using the hierarchical graph method to draw the road topologies and geometry. Their method is one of the fastest at the time, requiring only 0.2 seconds of computational time for a 200m x 200m area on the Argoverse dataset. Overall, the main challenge of graph-based SLAM methods is computational complexity, which directly translates to computational time. Although the creation of HD maps is usually done offline, the time required for mapping might not be of high importance. However, the trend towards crowdsourced mapping, which emphasizes collaboration in mapping to solve scalability and update frequency issues, is gaining traction. Consequently, more and more research is focused on developing online mapping methods.


Besides the use of SLAM in graph-based mapping, the traditional SLAM algorithm can also be used to create online mapping, as shown in the vision-based and lidar-based localization subsections. \cite{Fenwick2002} is one of the early works that provide a map in an online fashion. \cite{Gil2010} was the first to propose a SLAM method in which the map is shared among all the robots. The approach is tested in simulation, and the result shows the lack of computational power to meet the real-time demand for mapping and vehicle localization. Then, the next iteration uses the cloud to take advantage of larger computational power. \cite{Riazuelo2014} proposes a C2TAM where the mapping workload is shared among a cloud server and the vehicle. The DAvinCi \cite{Arumugam2010} presents a new architecture where all of the computations are being put forward to the cloud system. In this iteration, the author utilizes the FastSLAM \cite{Montemerlo-2003-8695} algorithm to be adapted to fit this approach. However, the delays and latencies induced by these computations on the real-time application are not considered. \cite{Mohanarajah2015} uses RGB-D odometry \cite{Civera2010} to perform SLAM mapping in the cloud system. In this system, they compress the keyframe data to reduce the bandwidth. However, even after the compression has been done, the bandwidth is still too large, making the author split the computations to make it viable. Cloud-based mapping is relatively new and has excellent potential to create accurate mapping. Interested readers can refer to \cite{Kehoe2015} for a general overview of this current practice and \cite{Cadena2016} for the overall review of the SLAM as a subject. It is important to note that the mapping result obtained from SLAM approaches can be dense in theory; however, it lacks semantic information, which is one of the necessary aspects of HD maps.

\subsubsection{Online Learning approaches}
Early learning approaches treated the map learning tasks as a segmentation problem in BEV space. HDMapNet \cite{Li2022} was the first to predict semantic information of maps directly from camera images and LiDAR point clouds. This work was notably the first to estimate vectorized local semantic maps through an end-to-end approach. Their novel methodology set a precedent, inspiring subsequent research that emphasizes the direct vectorization of local semantic maps within the network framework. Next came Lift-Splat-Shoot \cite{Philion2020}, which further provides a solid foundation for the development of online learning approaches in the BEV space. It is important to note that the rasterized mapping result from these maps did not include the idea of instances, so they cannot be used for tasks that need vectorized maps, like motion planning or motion forecasting \cite{Gao2020a}. SuperFusion \cite{Dong2022} proposes a fusion of both LiDAR and camera data to generate vectorized HD maps for long-range distances up to 90 m. However, the mapping results will significantly deteriorate when one of the data is missing. 

More recent approaches utilize the transformer idea inspired by DETR\cite{carion2020end} and are able to achieve promising results. End-to-end vectorized local HD map generation is now a hot topic, and more and more works have been published to address this problem.
The MAPTR project \cite{Liao2022} is one of the first to use transformer architecture. They are focused on structured modeling for making HD maps and their approach is called MapTR. They use classification, point-to-point, and edge direction loss to model the loss function and improve the encoder-decoder transformer architecture and hierarchical bipartite matching. In the next iteration, which is called MapTRv2 \cite{Liao2023}, they are able to improve the accuracy by decoupling self-attention tailored for the hierarchical query mechanism that significantly reduces memory consumption. They also introduce a one-to-many set prediction branch to accelerate convergence. Then, they utilize auxiliary dense supervision on both BEV and normal views, which increases the performance. They also extend the structure by adding centerlines for downstreaming motion planning tasks.
VectorMapNet \cite{Liu2023a} was a groundbreaking framework for autonomously generating HD maps. This approach uniquely combines end-to-end vectorized map learning, leveraging the power of polylines to represent complex road geometries accurately and efficiently. The method's creative use of polyline representation and set detection models is a big step forward in the field of self-driving cars. It works better than traditional rasterized map predictions at accurately capturing detailed road geometries.
BeMapNet \cite{Qiao2023a} employs piecewise Bezier curves to parameterize and map complex road elements efficiently. The approach offers significant improvements over existing state-of-the-art methods in terms of map accuracy and versatility, showcasing its potential to enhance the capabilities of autonomous driving systems.
InstaGraM \cite{Shin2023} models the polylines of map elements with bidirectional edges to predict edges optimally. They are able to extend a graph-based approach to graph neural networks. Their method is able to eliminate the need to perform heuristic post-processing for a large computational cost to realize real-time polyline performance.  
PivotNet \cite{Ding2023} proposes a novel Point-to-Line Mask module to enhance point-line relationships in network modelling and introduces a Pivot Dynamic Matching module for handling the topology in dynamic point sequences.
ScalableMap \cite{Yu2023} proposes a concept of hierarchical sparse map representation (HSMR) to perform map abstraction in a sparse manner and yet remains accurate. They integrate this representation with cascaded decoding layers from DETR \cite{carion2020end} and exploit the scalability of vectorized map elements to further restrain the structured information of map elements. They manage to effectively capture long-distance information, a feat similar to the one proposed by \cite{Dong2022}, however, this approach only requires a camera only instead of a fusion of camera and lidar. 
StreamMapNet \cite{Yuan2023} addresses the limitations of previous methods by extending the perception range and effectively leveraging temporal information through streaming strategy. The method makes the quality and consistency of vectorized local HD maps much better over time. 
\cite{Qiao2023} presents a novel architecture called MachMap. They propose a map compaction scheme, reducing vectorized points by 93\% without performance degradation, and a strong query-based paradigm for map element representation. This approach significantly enhances compactness and efficiency in HD map construction, outperforming other methods and becoming the winner in the first online mapping competition \cite{TsinghuaMARS2023OnlineHDMap}. MapNeXt \cite{li2024} is an upgraded version of the MapTR architecture for real-time, high-definition map construction. This work focuses on enhancing the model's training dynamics and scaling approaches, leading to substantial improvements in map accuracy and inference speed. ADMap \cite{Hu2024} also extends the framework used in MapTR architecture. They add interactive attention to instances at the decoder layer. This helps the network better understand the links between point levels by using the extracted instance embeddings. Their method can address the problem of jitter and distortion in vector point sequences. It is crucial to recognize that most of these methods, which rely solely on camera sensor data, may experience performance degradation under conditions that limit scene visibility. Moreover, the potential for changes in camera calibration presents a challenge, as these methods often assume fixed constraints. This underscores the necessity for an end-to-end approach capable of learning and adjusting to calibration changes in real-time. In general, these mapping algorithms can create an accurate local HD map that consists of simple map topologies. Further research regarding handling complex map topology scenarios, such as complex intersections, in a real-time manner still remains an open challenge.

\subsubsection{Map fusion approaches}
\begin{figure*}[h]
    \centering
    \includegraphics[width=1 \linewidth]{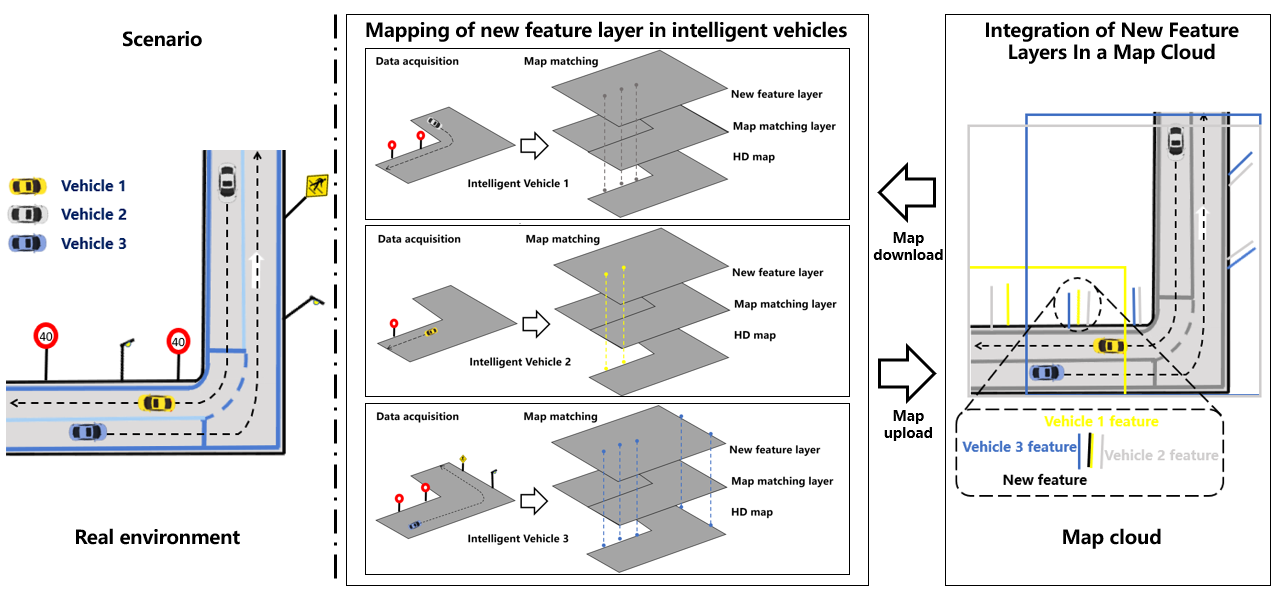}
    \caption{The mapping process of the new feature layer with a map cloud system. Figure is redrawn and modified based on depictions in \cite{Kim2018}}
    \label{fig:crowdsourced_mapping}
\end{figure*}

Fusion approaches are the extension of the local mapping algorithm and are very important in the HD map creation process. In \cite{Tao2010}, local maps are built individually, and they are fused in a centralized manner on the local server. This is one of the first studies on map fusion approaches.\cite{Massow2016} use geometric constraints of lane line width to create a fusion model of lane lines and achieve a 0.5 m average in lateral accuracy. LineNet \cite{Liang2018} proposed a clustering algorithm to fuse the road lines. They successfully merged several maps created from inaccurate GPS data and large-interval image capture on the overlapping area. They approach the problem by first detecting the lane with the CNN algorithm and reconstructing the ground surface in three different road conditions (straight, turning, and crossroad). Then, the lane lines were merged using a clustering algorithm based on the hierarchical distance between lines. This method was applied again to zoomed images several times to obtain a stable line position. This manipulation process will continue until the minimum line distance has been obtained. \cite{Herb2019} propose a two-step semantic mapping: a multi-session fusion that combines multiple maps using feature-based alignment and reconstruction of semantic edge maps. The front-end method relies on a visual odometry approach to extract lane boundaries. These lane boundaries from multiple journeys later matched pairwise using a rough estimate of global position from each single session keyframe. Then, the RANSAC algorithm \cite{Fischler1981} is performed to remove the outliers before the map reconstruction is performed. Inside the semantic map reconstruction process, the edge points have to be confirmed of their existence in multiple image keyframes.
Later, the confirmed edge was refined by connecting nearby unique edges or adding new edges on an empty surface. It is important to note that these works focus mainly on the HD map element on the road surface, such as lane lines and lane marking. There are only a few works that perform crowdsourced mapping of the vertical element of the map, such as traffic poles, traffic lights, and traffic signs. \cite{Kim2018} proposed a fusion mapping architecture of feature layers to add new map elements to the database shown in Fig. \ref{fig:crowdsourced_mapping}. 
\cite{Stoven-Dubois2019} introduces a collaborative approach to localizing map landmarks (traffic signs). They can improve localization accuracy from 8 m to 1.5m on average. In the next iteration, \cite{Stoven-Dubois2020} proposes a promising solution of HD map update also by graph-based approach. The vehicles are assumed to have GPS and monocular cameras installed. They also rely on the overlapping landmarks of the map to realize the update. Their result shows that the cross-correlation approach achieved 0.08 – 0.2 m accuracy after 1000 vehicles passing by. \cite{Wijaya2022} propose a point-to-point approach to combine multiple local maps using pixel-wise confidence generated from the integral matrix. This approach shows that it can effectively remove unwanted map features, such as road surfaces. \cite{Song2023} proposes an enhanced semantic alignment algorithm alongside a semantic aggregation algorithm, both evaluated in practical scenarios. They come up with a two-step semantic alignment algorithm based on semantic GICP (Generalised Iterative Closest Point) that is meant to combine local maps better. This method notably improves the accuracy of semantic alignment. Furthermore, they develop a semantic aggregation algorithm that utilizes lane-line logic constraints. This algorithm effectively lessens the effect of abnormal data on the instantiation of semantic elements, showing that it can be used to improve the accuracy of maps. Map fusion method will become more and more relevant in the HD map generation techniques given the trend towards crowdsourcing and the increase in research in online local mapping based on learning algorithms.

\subsubsection{Discussion}
In summary, map-building modules have evolved significantly, incorporating various techniques to connect semantic information from images to 3D spatial points. The methods range from geometric approaches, which use multi-view geometry and triangulation algorithms for reconstructing 3D scenes, to SLAM mapping, focusing on the long-term deployment of maps for shared vehicle services. Geometric approaches, as seen in the works of Longuet-Higgins \cite{longuet-higgins1987computer} and others, have advanced from basic scene reconstruction to complex traffic signs and lane mapping using consumer-grade sensors. These have evolved into more sophisticated methods that combines learning-based and geometric techniques for enhanced accuracy and speed. SLAM mapping, on the other hand, emphasizes the mapping process, using methods like GraphSLAM for large-scale urban mapping and incorporating advanced techniques for higher accuracy and real-time applications. These approaches, while efficient, often lack the semantic detail crucial for HD maps, leading to a shift towards online learning and map fusion approaches. Online learning approaches, such as MapTRv2, demonstrate real-time, high-definition map construction capabilities, catering to dynamic driving environments. Map fusion approaches extend local mapping resulting from online learning approaches, focusing on combining multiple maps for a comprehensive representation.
The trend in map-building is clearly moving towards a fusion of geometric, SLAM, and learning-based approaches for more accurate, dynamic, and semantically rich HD maps. In addition to this trend, recent works have shown a shift toward online and collaborative methods, pointing to a time when real-time, precise, and extensive HD mapping will be commonplace in autonomous vehicles.

\section{HD Map Update Module}
To perform update, the first thing to do is to match the perception result of the vehicle with the map database. This matching process can be performed on either the image or 3D world coordinate domains. The result of this matching process is the detection of change in the map database or confirmation of the validity of the map database. \cite{Quddus2006} proposed an integrity variable from 0 to 100 to model the integrity risk associated with the map matching process. According to the author, this risk represents the acceptable trade-off between the probabilities of missed detections and false alarms. Given the importance of the map-matching performance, \cite{Kubicka2015} proposed the first map-matching dataset to compare the state-of-the-art matching approach. However, this dataset can also be used as an offline training dataset to model learning algorithms to perform the map matching process. The downside of this dataset is that it only provides the tracks trajectories instead of the complete map element information usually provided from the HD map database. In general, the researcher will perform an update only after the map database change has been determined from the map-matching process. The update process varies from a direct update to an incremental update. 

\subsection{Change Detection}
Change detection is an integral part of an HD map update. It is necessary to find the "false data" on the map database before the update process. This step has recently gained traction in the research direction of HD map update \cite{Heo2020}\cite{Zhang2021}\cite{Jo2018}. There are two types of change detection: single-session change detection and multi-session change detection. Each of these types will be elaborated further in the subsection below.

\begin{figure*}[h]
    \centering
    \includegraphics[width=0.9 \linewidth]{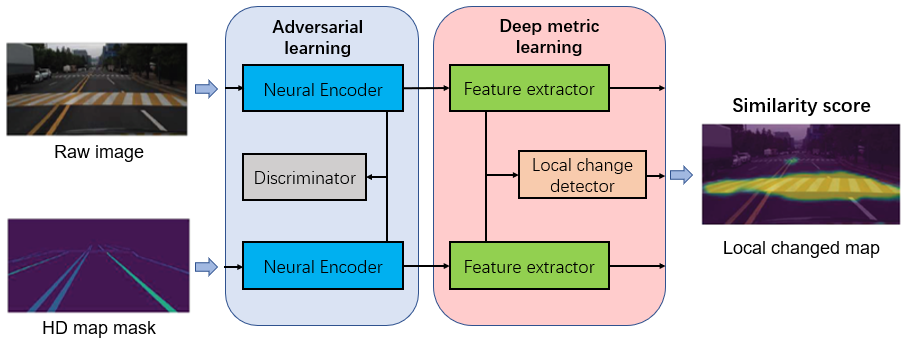}
    \caption{The illustration of deep learning framework which consist of adversarial learning, metric learning, and local change detection to estimate the change in the image directly. Figure is redrawn and modified based on depictions in \cite{Heo2020}}
    \label{fig:deep_metric}
\end{figure*} 

\subsubsection{Single Session Change Detection}

Single-session change detection is a change detection that is directly done in the vehicle. In which, the resulting confidence of then change is sent to the server for the update, or it can directly suggest an update to the server \cite{Heo2020}\cite{Zhang2021}, or other autonomous vehicles \cite{Jo2018}. \cite{Jo2018}introduces a simultaneous localization and map change update (SLAMCU) approach based on Dempster-Shafer evidence theory to evaluate the HD map feature existence. They also use a Rao-Blackwellized particle filter (RBPF) to determine the vehicle location and calculate the new map state, and this information will be shared with the other autonomous vehicles directly instead of the server. 

In \cite{Heo2020}, they use direct deep metric learning, which requires the HD map to be projected directly into the image space as shown in Fig \ref{fig:deep_metric}. In this space, the segmented image is compared with the projected map to determine the change by using adversarial learning to extract features from the neural encoder. Finally, the resulting similarity score will be processed to calculate the change between the map and the condition of the surroundings. \cite{Lambert2021} proposed a trust but verify (TbV) dataset to train a learning-based formulation to detect changes on a vector HD map. They found that the bird's eye view projection approach is inferior to direct projection to the image plane in terms of change detection. These approaches are implemented using consumer-based sensor equipment on the intelligent vehicle.
On the other hand, \cite{Zhang2021} utilizes mid to high-end sensors, including an industry camera, a high-end Global Navigation Satellite System (GNSS)/Inertial Measurement Unit (IMU), and an onboard computing platform, a real-time HD map change detection method for crowdsourcing update. They use the RANSAC algorithm \cite{Fischler1981} to match the detected features with the map with a matching degree coefficient derived from the overlapped area of the map element. Furthermore, the statistical result gives the change detection, and an update is suggested directly. With this, they have demonstrated the feasibility and significant value of crowdsourcing updates for HD maps. 
\cite{Zhanabatyrova2023} uses dashcam videos to perform change detection of traffic signs. They build the updated solution on top of existing methods such as SfM \cite{schoenberger2016sfm}, semantic segmentation, and object detection and are able to achieve 85\% accuracy in detecting changes. \cite{Cong2023} proposes a change-aware online 3D mapping framework (CAOM) in urban area. This system integrates data from Multi-Beam LiDAR (MBL) and Push-Broom LiDAR (PBL), achieving 95\% accuracy in change detection. The downside of this method is the sensor required is LiDAR which is not a common sensor equipped on mass-produced vehicles.

Although there are significant improvements in single-session change detection research, it is not sufficient to provide complete insight into the map trajectory passed by the vehicle. Although these methods are feasible and robust since the nature of the single session change detection is limited by the field of view of the sensor at the time of data acquisition, semi-static change problem, which refers to objects such as boxes in a logistic environment or parked cars on the side of the road. Their movement cannot be detected directly from measurements as their position infrequently changes at a larger time than the measurement time scale. Thus, the single-session vehicle cannot exclude semi-static objects directly from measurements to ensure the map element blocked by these semi-static objects confirms its existence. This problem will result in false change detection caused by the observation bias of the intelligent vehicle, making it difficult for the primary approach to detect changes in the map.

\subsubsection{Multi Session Change Detection}
\begin{figure}[h]
    \centering
    \includegraphics[width=0.9 \linewidth]{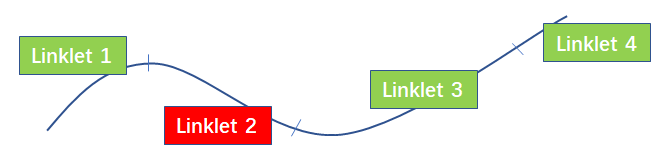}
    \caption{Visualization of road linklets and the standard definition map topology with links and vertices to determine the precise location of the update required. Figure is redrawn and modified based on depictions in \cite{Pannen2020}}
    \label{fig:linklet}
\end{figure}

Multi-session change detection is proposed to solve the observation bias problem caused by the sensor's field of view limitation. Many researchers have come up with multi-session change detection in order to update the HD map \cite{Pannen2019} \cite{Pannen2020}\cite{Li2020}\cite{Kimlidar2021}. The map change is usually decided by a confidence value. A different method would model this value differently according to the application. In \cite{Pannen2019}, they use mean particle weight, belief weight, mean inner lane geometry innovation, and mean lane geometry weight to indicate the quality of the map features localization solution. The intention behind these metrics is that if the map does not represent the real world, the quality of the localization solution will degrade and thus be reflected by lower weight values. The inner lane geometry metric introduced above has strong requirements for localization relative to the map to avoid wrong assignments between observations and lane marking entities in the map. In the next iterations, \cite{Pannen2020} proposed to use linklet criteria to divide the map topology into links and vertices to determine the precise location of change shown in Fig. \ref{fig:linklet}. They addressed the HD map update by utilizing floating car data (FCD) from vehicles already on the road and offer a practical approach to ensure HD map data is always up to date. Their method divides road networks into several small road links called linklets to separate the part of the map that needs updating. They introduce a regressor algorithm to determine the probability of change given the training dataset corresponding to the ground truth information. The algorithm will automatically initiate the map update sequence based on this probability of change. Crowdsourced images, which is obtained through multi-session vehicle trajectories, can also be used to perform lane marking change detection \cite{Li2020}. They model the confidence using the Bayesian model, updated through Gaussian belief function distribution by considering the noises from camera motion. They also perform a goodness-of-fit test in the map-matching process. However, this method is limited to the change of the lane marking only.
In \cite{Kimlidar2021}, they perform change detection from crowdsourced vehicles equipped with LIDAR sensors. They propose a probabilistic and evidential approach to update the existence field of the point cloud points and matching them with the map to label them as existing, new, or deleted. They can improve the result of single-session change detection by 14.46\% in the F1 score, calculated using precision and recall criteria compared with the multi-session approach. Wijaya et al.\cite{Wijaya2023} perform multi-session change detection along Yizhuang District in Beijing, China. They propose occupancy objects for map elements to determine the confidence of detection via the Bayesian recursion framework. This method is able to achieve 90\% accuracy in linklet areas that are experiencing map changes.

\subsubsection{Discussion}
Change detection is a relatively new domain in the research of HD maps. Although this process is required in the update mechanism, most researchers in the past focused on providing the best approach to updating in terms of its accuracy, not finding the update location. More and more researchers realized that finding changes on the map remains a challenge because they are often subtle. The single-session approach of using a learning algorithm seems to be the trend, as it offers direct evidence of change. However, the issue with single-session is always the observation bias, where a single vehicle cannot provide the whole observation required to determine the change in a map section. Multi-session, however, can provide this solution by accommodating several points of view and confidence modeling. This approach has the potential to find changes on the map wholly and effectively. Nevertheless, by using more vehicle data, the reliability of the data can become an issue as well. Further research in this domain is required to find the best confidence model to ensure data reliability and improve change detection accuracy.

\subsection{HD Map Update Method}
HD map update is the process of patching the HD map database according to the actual state of the map element on the road. There are two popular methods: direct updates and incremental updates. The mapping companies usually adopt the direct update route, which relies on mapping vehicles to perform the update. On the other hand, the incremental update approach often intersects more with the crowdsourced vehicles, which relies on the collaboration of crowdsourced data to generate the updated map element. Each of these methods will be elaborated further in the subsection below.

\subsubsection{Direct Update}
Direct update is still a common approach for HD map maker companies to ensure the quality of the map. TomTom \cite{TomTomInternationalBV2019} leverages a combination of automated data fusion algorithms and a fleet of MMS equipped with state-of-the-art mapping sensors. They can process large volumes of data to update the map database. This process is usually slow since it requires mapping vehicles to travel to the update area to perform the update. The crowdsourcing trend also affects the HD map update process, in \cite{Kim2021}, they proposed an approach that could utilize only reliable information for the map update by using the uncertainty information of the crowdsourced data. With the novel lane observation learner method, which uses shell structures, various crowdsourced lanes were assigned with HD maps and clustered effectively. Their method also allows the update to be adjusted according to the aggressiveness variable, which decides the degree of change in the update process.  \cite{Kimlidar2021} proposes an update method by merging the point cloud points layer obtained from the crowdsourced vehicle with the HD map database. They use evidential theory to add and remove the points from the HD map database. This approach assumes that every crowdsourced vehicle is equipped with a sophisticated LIDAR sensor, which is large in data acquisition size. Therefore, only the update suggestion is transmitted to the server instead of the raw measurements. This approach puts a heavy calculation burden on the vehicle on the road, which might not be feasible for a mass-produced vehicle. \cite{Cho2023} proposed an update strategy to achieve frequent and accurate updates of HD map. They propose two main algorithms designed for lane lines: the convex hull method to perform landmark searching and recursive Bayesian estimation to track changes. They are able to achieve a 31\% accuracy to suggest an update and automate the whole process to remove human intervention and ensure timely updates with reduced costs.

\subsubsection{Incremental Update}
Incremental update refers to the dynamic update process in terms of change, where the updates take place incrementally. Usually, the process takes a short amount of time to perform small corrections of the HD map elements by adding, removing, or displacing them. \cite{Liu2019} proposed an incremental update framework to associate the update frequency with its accuracy improvement. They use the Kalman filter to fuse the map element's position and semantic confidence information. Their experimental result shows that the accuracy requirement of HD map in simulation can be achieved after several update iterations.\cite{Liebner2019} model the HD map update by graph-based approach to combine the information between vehicle sessions. These variables are later transferred into factors in the factor graph calculation. The resulting map will ensure that the lane markings are connected smoothly with those in the old HD map. Finally, they also apply another SLAM optimization to compute the final geometries. \cite{Chao2020} proposes an iterative approach to refine the map through confidence modeling. This research defines confidence through the map credibility and its influence to provide trajectory-matching results. They optimize the map element status (exist and not exist) to maximize the overall quality results. Furthermore, they add an index-based trajectory filter to improve its overall efficiency. Aside from the update algorithm itself, \cite{Qi2021} considers the relationship between the calculation power of the crowdsourced vehicle and its sensing range as a constraint when performing the crowdsourced update. They propose a heuristic approach to jointly minimize the communication burden between crowdsourced vehicles and reduce transmission data by 37\% compared to the nearest node baseline mechanism. In CrowdRep \cite{Wijaya2022a}, a reputation system is proposed as the basis of the blockchain framework to determine consensus regarding the status of the map data. This is the first application of blockchain technology in an HD map update application. This method is used to rate the quality of map data from crowdsourced vehicles to ensure trustworthy updates. With the increase in the trend of crowdsourcing, the safety of the map data and the ability to rate the quality of the map data is going to be important information in conducting HD map updates in the future.

\subsubsection{Learning-based Update}
With the new trend in the learning-based model on mapping, \cite{Xiong2023} introduces Neural Map Prior (NMP), a neural representation of global maps for enhancing local map inference and automatic global map updates in autonomous driving. It leverages cross-attention to dynamically relate current and prior features, significantly improving map prediction performance, even under challenging conditions like adverse weather. This method represents the first learning-based system for building a global map prior, demonstrating compatibility with various map segmentation and detection architectures and yielding considerable improvements in performance on the NuScenes dataset. In \cite{Sayed2023} propose PolyMerge where it utilized the transformers to directly merge polylines of road elements using the local vector map generated. They project the reconstructed network graph of similar polylines to each other. This implementation is the first to use the transformers method, which is very popular now in the map-building task in the HD map update task.

\subsubsection{Discussion}
There are three primary methodologies in HD map updates: direct, incremental, and learning-based updates. Direct updates, commonly employed by companies like TomTom, rely heavily on fleets of Mobile Mapping Systems (MMS) equipped with sophisticated sensors, processing large data volumes for map database updates. This method, though slow, has seen innovations that enhance efficiency and accuracy. Incremental updates, in contrast, focus on dynamic, smaller-scale adjustments. Liu's\cite{Liu2019} framework utilizing the Kalman filter and Liebner's \cite{Liebner2019} graph-based approach exemplifies this method's capability to refine maps iteratively. Meanwhile, learning-based updates are emerging as a transformative trend. Xiong's Neural Map Prior (NMP)\cite{Xiong2023} leverages neural representations for global map updates, while  Polymerge's \cite{Sayed2023} use of transformers marks a novel approach in polyline merging for road elements. Aside from the quality of the update, the transmission of update data is often left untouched. Recently, \cite{Chen2023} explores an inner approximation method to reduce the complexity of the problem to find a suboptimal solution. Their method can achieve low latency updates for up to 8 vehicle applications. 
Collectively, these advancements underscore a significant evolution in HD map update strategies, integrating sophisticated algorithms and learning models to enhance accuracy and efficiency of HD map to improve the safety of autonomous driving tasks.

\section{Challenges and Future Directions}

\subsection{Challenges}
Here, we summarise the challenges arising while compiling this survey's research papers. Given the research trends towards a low-cost solution for HD map mapping and updates, these challenges have been thoroughly compiled from each section above.

\subsubsection{Real-time application vs accuracy}
The trade-off between calculation time and accuracy has been of paramount importance in the development of semantic segmentation and localization modules. This classic challenge persists due to the increasing demand for precise detection and localization, which are crucial for creating and updating an accurate HD map. Moreover, the robustness of the segmentation algorithm may be compromised as different countries have various types of map elements, thereby complicating the training process. A dependence on large datasets presents an additional challenge, especially when only a small dataset is available, adversely affecting algorithm accuracy. Localization accuracy can also vary significantly across different scenarios. Additionally, the adoption of crowdsourced methods heightens the need for real-time applications in mass-produced vehicles. Another significant factor is the latency in updates, which can greatly impact the scheduling of HD map updates. Therefore, achieving a balance among all these factors is crucial for advancing research on HD map updates.

\subsubsection{Identifying mapping and update state-of-the-art model}
The state-of-the-art model is determined by a competition, in which the relevant work are compared at each performance parameter at a time. Ensuring such fairness is a challenging task in every research field. Here, we discuss the challenges specific to the HD map mapping and update tasks discussed in this review article.

\begin{figure}[h]
    \centering
    \includegraphics[width=0.8 \linewidth]{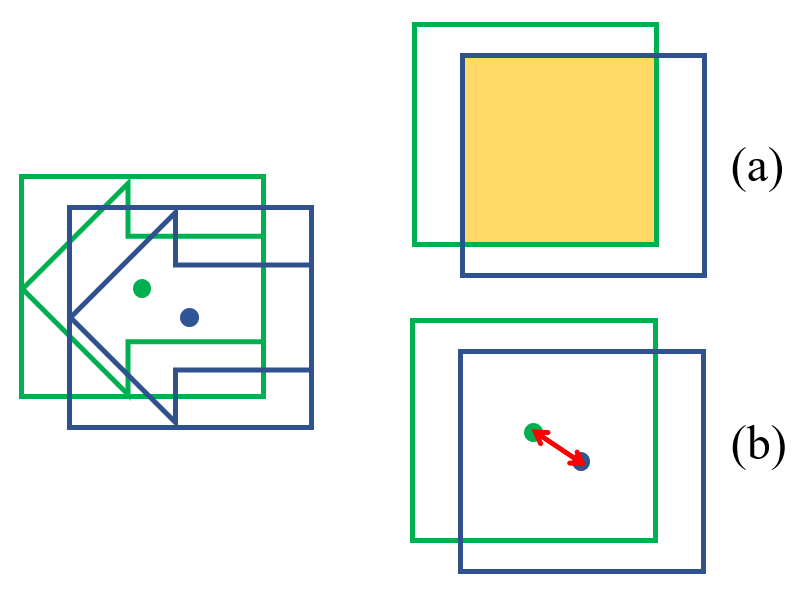}
    \caption{The quantification of map element accuracy: (a) Overlap percentage. (b) Distance between middle bounding box. Figure is redrawn and modified based on depictions in \cite{Zhang2021}}
    \label{fig:different_quantification}
\end{figure}

\textbf{Different quantification of mapping accuracy} 
When two different researchers use the same method to perform the mapping and update algorithm, their quantification of mapping accuracy might differ. For example, if the mapping accuracy is defined based on the matching degree, one paper might present the percentage of overlap between detected map element and map element from the database, while others defined it based on the precision of each map element meters. These two definitions as shown in Fig. \ref{fig:different_quantification} can create confusion when the researcher finds the state-of-the-art HD map mapping or update the model.
\textbf{Different metrics to measure performance}
Different metrics for performance creates challenges for comparing different research result. The presentation of the superior model using one metric does not necessarily translate linearly to the other metric. The most common metrics for mapping tasks (for example, accuracy) are usually misleading in a sense that the information provided is not followed with complementary information such as completeness of the map and accuracy of the map element classification. The lack of this information makes the mapping and update results rather dubious. Therefore, it is necessary to completely define a performance measurement system to ensure a suitable representation of the mapping and update result. This research direction may seem simple, but it can become the key to ensuring concrete advancement in this domain.

\textbf{Different quality of input data}\\
Differing the quality of input data can lead to a different result outcome. The quality referred to here does not only imply the sensor equipment quality, but it also implies the data acquisition condition, which can impact the outcome of the mapping and update result. Data acquisition, in this case, refers to the weather, traffic condition, and driving scenarios. These aspects imply a difference in the complexity of performing mapping and update tasks. Let us consider two researchers interested in performing mapping in two different areas, congested and free-flow traffic conditions. The observations towards map elements in free-flow traffic conditions will be more abundant than the congested ones. These differences alone can create a massive difference in the mapping result, and even though these two researchers are performing the mapping task using the same algorithm, the result can differ in reality.

\subsubsection{HD map security}
With the recent trend of performing HD map mapping and updating using crowdsourced data, the security aspect of the mapping itself can be a problem. When an entity wants to create an attack on the system to trigger a false update, the system has to detect these kinds of threats. The attack can be arranged by using a fleet of vehicles sending false data on a particular area. The false update triggered by this attack can lead to false information received by autonomous vehicles who are using the mapping service. Given the importance of HD maps, this issue can affect the vehicles' safety in the surroundings of the affected autonomous vehicle.

\subsubsection{Standardised datasets and competitions}
The fast-growing development in deep learning algorithms is an example of how good standardized quality data can accelerate a work domain, and the most worthy example is Computer Vision (CV). Some examples of specific CV tasks such as image classification and image segmentation. In some constraints situations, these methods can achieve near-human performance levels. The two notable impetuses to CV research were achieved after the release of ImageNet dataset \cite{JiaDeng2009} and Pascal Visual Object recognition dataset \cite{M.2010}. Different possibilities can arise for the domain of HD map mapping and update with access to standardized datasets and competition. Firstly, the performance improvement with the increasing amount of good-quality data. Secondly, the establishment of the state-of-the-art model given the standardization of evaluation criteria and competition rules for mapping and update. Thirdly, it can ease the implementation of a new model, in which a new algorithm can be validated and tried on a new set of datasets to speed up the development process in general. Some exclusive efforts to provide HD map datasets through LIDAR point cloud map are provided by KAIST university \cite{Jeong2018}, Argoverse \cite{Chang_2019_CVPR, Argoverse2}, NuScenes \cite{Caesar2020}, and Baidu \cite{DA4AD_2020_ECCV}. It is also important to note that the first HD map online mapping competition was conducted \cite{TsinghuaMARS2023OnlineHDMap}, and it quickly improved the advancement of the research direction.

\subsection{Future Directions}
This subsection highlights possible directions for future research. These directions are proposed in answer to the challenges which were presented in the previous subsection. Here, we highlight the importance of efficient optimization algorithms, and blockchain algorithm to increase HD map security.

\subsubsection{Efficient optimization algorithm}
This research direction is the continuous effort of the whole research community on this subject. This direction is more critical than ever since it is the driving force towards the real-life application of mapping and updating, especially in realizing it in a crowdsourced manner. As shown in \cite{Xue2022aa}, effective computing edge is very important in realizing fully distributed resource allocation. Efficient optimization can lower the sensors requirement of using state-of-the-art sensors, which is usually expensive. It can accelerate the adoption of the technology for road users. Besides efficient computation, efficient data is also required for the fast deployment of algorithms in various scenarios. The need for a small training dataset will accelerate the testing and implementation in the real world, which is excellent for the mass adoption of crowdsourced HD map mapping and update in the future.

\subsubsection{Blockchain algorithm}
The recent popularity of blockchain algorithms in the cryptocurrency domain of safekeeping the transfer ledger information between users brings the possibility of adapting the same idea on the crowdsourced map and updating the application. Since blockchain algorithms perform verification of the ledger by each member, the same can be applied in the map update domain, where the verification process is the change detection of the map element. The proposal of a new block translates to the map update stage. If this method can be realized, the safety of the HD map can be secured. The latest research in this domain also considers the contribution of the mapping data as a reputation value and use them to generate consensus to finally update the HD map data \cite{Wijaya2022a}. Here is a list of the latest research regarding attacks and defenses of crowdsourced mobile mapping services \cite{WangT2016}\cite{Wang2018}.

\section{Conclusion}
A concise general overview of the current state of the main algorithms involved in HD map mapping and update using an on-board sensing system was provided, along with the potentials and limitations of each detection and localization technique. From the performance point of view, it was shown that the LIDAR technique offers the greatest strength in building an accurate map, as it is often used in MMS fleets. However, this sensor's biggest problems are the high processing requirement and its high cost. Therefore, further optimization of LIDAR technology or an alternative vision-based approach could offer the path of commercial adoption from the mapping companies. Nevertheless, in other areas such as change detection and map update, a vision-based system proved feasible and effective, but further validation and resources are still required to ensure the validity of the state-of-the-art method in this space.
There is a common trends on the development path from data acquisition process, mapping, and update is low-cost and efficient solutions are desired. Along with the increase in performance of computing power in vehicle, the distributed system where some calculation can be delegated on the vehicles becomes popular in the effort to support the adoption of highly anticipated crowdsourced solution. However, with more and more adoption of low-cost crowdsourced techniques in both mapping and update can also potentially become a problem in terms of security, and hopefully, more and more researchers will start to focus on this research direction. 

\section*{Acknowledgment}
This work was supported in part by National Natural Science Foundation of China (U22A20104, 52102464), Beijing Natural Science Foundation (L231008), and Young Elite Scientist Sponsorship Program By BAST (BYESS2022153).



\bibliographystyle{unsrt}
\bibliography{ref}

\begin{thebibliography}{100}

\bibitem{SAEInternational2018}
{SAE International}.
\newblock {Taxonomy and Definitions for Terms Related to Driving Automation
  Systems for On-Road Motor Vehicles}.
\newblock {\em SAE International}, 4970(724):1--5, 2018.

\bibitem{Levinson2011}
Jesse Levinson, Jake Askeland, Jan Becker, Jennifer Dolson, David Held, Soeren
  Kammel, J.~Zico Kolter, Dirk Langer, Oliver Pink, Vaughan Pratt, Michael
  Sokolsky, Ganymed Stanek, David Stavens, Alex Teichman, Moritz Werling, and
  Sebastian Thrun.
\newblock {Towards fully autonomous driving: Systems and algorithms}.
\newblock {\em IEEE Intelligent Vehicles Symposium, Proceedings}, (March
  2015):163--168, 2011.

\bibitem{Wijaya2022}
Benny Wijaya, Kun Jiang, Mengmeng Yang, Tuopu Wen, Xuewei Tang, and Diange
  Yang.
\newblock {Crowdsourced Road Semantics Mapping Based on Pixel-Wise Confidence
  Level}.
\newblock {\em Automotive Innovation}, (0123456789), 2022.

\bibitem{Heo2020}
Minhyeok Heo, Jiwon Kim, and Sujung Kim.
\newblock {HD Map Change Detection with Cross-Domain Deep Metric Learning}.
\newblock {\em Iros}, pages 10218--10224, 2020.

\bibitem{Herb2019}
Markus Herb, Tobias Weiherer, Nassir Navab, and Federico Tombari.
\newblock {Crowd-sourced Semantic Edge Mapping for Autonomous Vehicles}.
\newblock {\em IEEE International Conference on Intelligent Robots and
  Systems}, pages 7047--7053, 2019.

\bibitem{Kim2021}
Kitae Kim, Soohyun Cho, and Woojin Chung.
\newblock {HD Map Update for Autonomous Driving with Crowdsourced Data}.
\newblock {\em IEEE Robotics and Automation Letters}, 6(2):1--7, 2021.

\bibitem{Meng2022}
Liqiu Meng.
\newblock {\em The ``Here and Now'' of HD Mapping for Connected Autonomous
  Driving}, pages 329--340.
\newblock Springer Nature Singapore, Singapore, 2022.

\bibitem{Schreiber2013}
Markus Schreiber, Carsten Kn{\"{o}}ppel, and Uwe Franke.
\newblock {LaneLoc: Lane marking based localization using highly accurate
  maps}.
\newblock {\em IEEE Intelligent Vehicles Symposium, Proceedings},
  (Iv):449--454, 2013.

\bibitem{Wen2020}
Tuopu Wen, Zhongyang Xiao, Benny Wijaya, Kun Jiang, Mengmeng Yang, and Diange
  Yang.
\newblock {High Precision Vehicle Localization based on Tightly-coupled Visual
  Odometry and Vector HD Map}.
\newblock {\em IEEE Intelligent Vehicles Symposium, Proceedings}, pages
  672--679, 2020.

\bibitem{Felix2021}
Felix Leach.
\newblock {\em Unsettled Issues on HD Mapping Technology for Autonomous Driving
  and ADAS}, pages 1--24.
\newblock 2021.

\bibitem{Chen2010}
Anning Chen, Arvind Ramanandan, and Jay~A. Farrell.
\newblock {High-precision lane-level road map building for vehicle navigation}.
\newblock {\em Record - IEEE PLANS, Position Location and Navigation
  Symposium}, pages 1035--1042, 2010.

\bibitem{Joshi2015}
Avdhut Joshi and Michael~R. James.
\newblock {Generation of accurate lane-level maps from coarse prior maps and
  lidar}.
\newblock {\em IEEE Intelligent Transportation Systems Magazine}, 7(1):19--29,
  2015.

\bibitem{Liu2020}
Rong Liu, Jinling Wang, and Bingqi Zhang.
\newblock {High Definition Map for Automated Driving: Overview and Analysis}.
\newblock {\em Journal of Navigation}, 73(2):324--341, 2020.

\bibitem{Bao2023}
Zhibin Bao, Sabir Hossain, Haoxiang Lang, and Xianke Lin.
\newblock {A review of high-definition map creation methods for autonomous
  driving}.
\newblock {\em Engineering Applications of Artificial Intelligence},
  122(March):106125, 2023.

\bibitem{Tang2023a}
Xuewei Tang, Kun Jiang, Mengmeng Yang, Zhaoyang Liu, Peijin Jia, Benny Wijaya,
  and Tuopu Wen.
\newblock {High-Definition Maps Construction Based on Visual Sensor : A
  Comprehensive Survey}.
\newblock {\em IEEE Transactions on Intelligent Vehicles}, PP:1--23, 2023.

\bibitem{Zhang2016}
Tao Zhang, Stefano Arrigoni, Marco Garozzo, Dian ge~Yang, and Federico Cheli.
\newblock {A lane-level road network model with global continuity}.
\newblock {\em Transportation Research Part C: Emerging Technologies},
  71:32--50, 2016.

\bibitem{asam2021}
{ASAM OpenDRIVE 1.7.0 Release}.
\newblock Technical report, 2021.

\bibitem{Hubertus2019}
Philip Hubertus, Martin Schleicher, Fabian Klebert, Georg Horn, and Markus
  Junker.
\newblock {NDS: The Benefits of a Common Map Data Standard for Autonomous
  Driving}.
\newblock page~11, 2019.

\bibitem{TomTom2020}
TomTom.
\newblock {Extending the vision of automated vehicles with HD Maps and ADASIS}.
\newblock {\em Https://Www.Tomtom.Com/Products/Hd-Map/}, page~9, 2020.

\bibitem{ETSI2011}
{ETSI Technical Committee Intelligent Transport System (ITS)}.
\newblock {Intelligent Transport Systems (ITS); Vehicular Communications; Basic
  Set of Applications; Local Dynamic Map (LDM); Rationale for and guidance on
  standardization}.
\newblock Technical report, 2011.

\bibitem{Jiang2019}
Kun Jiang, Diange Yang, Chaoran Liu, Tao Zhang, and Zhongyang Xiao.
\newblock {A Flexible Multi-Layer Map Model Designed for Lane-Level Route
  Planning in Autonomous Vehicles}.
\newblock {\em Engineering}, 5(2):305--318, 2019.

\bibitem{Diaz-Diaz2022}
Alejandro Diaz-Diaz, Manuel Ocana, Angel Llamazares, Carlos Gomez-Huelamo,
  Pedro Revenga, and Luis~M. Bergasa.
\newblock {HD maps: Exploiting OpenDRIVE potential for Path Planning and Map
  Monitoring}.
\newblock {\em IEEE Intelligent Vehicles Symposium, Proceedings},
  2022-June(Iv):1211--1217, 2022.

\bibitem{AECC2020}
AECC.
\newblock {Operational Behavior of a High Definition Map Application White
  Paper}.
\newblock 2020.

\bibitem{jiang2022}
Kun Jiang, Yining Shi, Benny Wijaya, Mengmeng Yang, Tuopu Wen, Zhongyang Xiao,
  and Diange Yang.
\newblock {Map Container: A Map-based Framework for Cooperative Perception}.
\newblock pages 1--10, 2022.

\bibitem{qiu2020}
Haonan Qiu, Adel Ayara, and Birte Glimm.
\newblock {A Knowledge Architecture Layer for Map Data in Autonomous Vehicles}.
\newblock {\em 2020 IEEE 23rd International Conference on Intelligent
  Transportation Systems, ITSC 2020}, 2020.

\bibitem{Javanmardi2018}
Ehsan Javanmardi, Mahdi Javanmardi, Yanlei Gu, and Shunsuke Kamijo.
\newblock {Factors to evaluate capability of map for vehicle localization}.
\newblock {\em IEEE Access}, 6:49850--49867, 2018.

\bibitem{Brimicombe2020}
Allan Brimicombe.
\newblock {\em {Data and information quality issues}}.
\newblock 2020.

\bibitem{Logo2021}
J~M L{\'{o}}g{\'{o}}, N~Krausz, V~Pot{\'{o}}, and A~Barsi.
\newblock {Quality Aspects of High-Definition Maps}.
\newblock {\em The International Archives of the Photogrammetry, Remote Sensing
  and Spatial Information Sciences}, XLIII(B-4):389--394, 2021.

\bibitem{Krehlik2023}
{\v{S}}t{\v{e}}p{\'{a}}n Křehl{\'{i}}k, Marek Van{\v{z}}ura, and Adam Skokan.
\newblock {Minimum required accuracy for HD maps}.
\newblock {\em Journal of Navigation}, 76(2-3):238--254, 2023.

\bibitem{Li2022}
Qi~Li, Yue Wang, Yilun Wang, and Hang Zhao.
\newblock {HDMapNet: An Online HD Map Construction and Evaluation Framework}.
\newblock {\em Proceedings - IEEE International Conference on Robotics and
  Automation}, pages 4628--4634, 2022.

\bibitem{Zhang2023b}
Gongjie Zhang, Jiahao Lin, Shuang Wu, Yilin Song, Zhipeng Luo, Yang Xue,
  Shijian Lu, and Zuoguan Wang.
\newblock {Online Map Vectorization for Autonomous Driving: A Rasterization
  Perspective}.
\newblock (NeurIPS):1--16, 2023.

\bibitem{Houston2020}
Long Chen, John Houston, Guido Zuidhof, Luca Bergamini, Yawei Ye, Long Chen,
  Ashesh Jain, Sammy Omari, Vladimir Iglovikov, and Peter Ondruska.
\newblock {One Thousand and One Hours: Self-driving Motion Prediction Dataset}.
\newblock {\em Proceedings of Machine Learning Research}, 155(CoRL
  2020):409--418, 2020.

\bibitem{Massow2016}
K.~Massow, B.~Kwella, N.~Pfeifer, F.~H{\"{a}}usler, J.~Pontow, I.~Radusch,
  J.~Hipp, F.~D{\"{o}}litzscher, and M.~Haueis.
\newblock {Deriving HD maps for highly automated driving from vehicular probe
  data}.
\newblock {\em IEEE Conference on Intelligent Transportation Systems,
  Proceedings, ITSC}, pages 1745--1752, 2016.

\bibitem{Kim2018}
Chansoo Kim, Sungjin Cho, Myoungho Sunwoo, and Kichun Jo.
\newblock {Crowd-sourced mapping of new feature layer for high-definition map}.
\newblock {\em Sensors (Switzerland)}, 18(12):1--17, 2018.

\bibitem{Liang2018}
Dun Liang, Yuanchen Guo, Shaokui Zhang, Song-Hai Zhang, Peter Hall, Min Zhang,
  and Shimin Hu.
\newblock {LineNet: a Zoomable CNN for Crowdsourced High Definition Maps
  Modeling in Urban Environments}.
\newblock pages 1--11, 2018.

\bibitem{Liebner2019}
Martin Liebner, Dominik Jain, Julian Schauseil, David Pannen, and Andreas
  Hackeloer.
\newblock {Crowdsourced HD map patches based on road model inference and
  graph-based slam}.
\newblock {\em IEEE Intelligent Vehicles Symposium, Proceedings},
  2019-June(Iv):1211--1218, 2019.

\bibitem{Pannen2019}
David Pannen, Martin Liebner, and Wolfram Burgard.
\newblock {HD map change detection with a boosted particle filter}.
\newblock {\em Proceedings - IEEE International Conference on Robotics and
  Automation}, 2019-May:2561--2567, 2019.

\bibitem{Liu2019}
Yinqi Liu, Mengxuan Song, and Yafeng Guo.
\newblock {An incremental fusing method for high-definition map updating}.
\newblock {\em Conference Proceedings - IEEE International Conference on
  Systems, Man and Cybernetics}, 2019-Octob:4251--4256, 2019.

\bibitem{Pannen2020}
David Pannen, Martin Liebner, Wolfgang Hempel, and Wolfram Burgard.
\newblock {How to Keep HD Maps for Automated Driving Up To Date}.
\newblock {\em Proceedings - IEEE International Conference on Robotics and
  Automation}, pages 2288--2294, 2020.

\bibitem{Chao2020}
Pingfu Chao, Wen Hua, and Xiaofang Zhou.
\newblock {Trajectories know where map is wrong: an iterative framework for
  map-trajectory co-optimisation}.
\newblock {\em World Wide Web}, 23(1):47--73, 2020.

\bibitem{Li2020}
Binbin Li, Dezhen Song, Aaron Kingery, Dongfang Zheng, Yiliang Xu, and Huiwen
  Guo.
\newblock {Lane marking verification for high definition map maintenance using
  crowdsourced images}.
\newblock {\em IEEE International Conference on Intelligent Robots and
  Systems}, pages 2324--2329, 2020.

\bibitem{Qi2021}
Yanli Qi, Yiqing Zhou, Zhengang Pan, Ling Liu, and Jinglin Shi.
\newblock {Crowd-Sensing Assisted Vehicular Distributed Computing for HD Map
  Update}.
\newblock {\em IEEE International Conference on Communications}, 2021.

\bibitem{Kimlidar2021}
Chansoo Kim, Sungjin Cho, Myoungho Sunwoo, Paulo Resende, Benazouz Bradai, and
  Kichun Jo.
\newblock {Updating Point Cloud Layer of High Definition (HD) Map Based on
  Crowd-Sourcing of Multiple Vehicles Installed LiDAR}.
\newblock {\em IEEE Access}, 9:8028--8046, 2021.

\bibitem{Zhanabatyrova2023}
Aziza Zhanabatyrova, Clayton~Frederick {Souza Leite}, and Yu~Xiao.
\newblock {Automatic Map Update Using Dashcam Videos}.
\newblock {\em IEEE Internet of Things Journal}, 10(13):11825--11843, 2023.

\bibitem{Xue}
Yongjie Xue, Yuru Zhang, Qiang Liu, Dawei Chen, and Kyungtae Han.
\newblock {CoMap: Proactive Provision for Crowdsourcing Map in Automotive Edge
  Computing}.
\newblock {\em IEEE International Conference on Communications (ICC}, 2023.

\bibitem{Gao2017}
Yali Gao, Xiaoyong Li, Jirui Li, and Yunquan Gao.
\newblock {A dynamic-trust-based recruitment framework for mobile crowd
  sensing}.
\newblock {\em IEEE International Conference on Communications}, pages 1--6,
  2017.

\bibitem{Gao2017a}
Yali Gao, Xiaoyong Li, Jirui Li, and Yunquan Gao.
\newblock {DTRF: A dynamic-trust-based recruitment framework for Mobile Crowd
  Sensing system}.
\newblock {\em Proceedings of the IM 2017 - 2017 IFIP/IEEE International
  Symposium on Integrated Network and Service Management}, pages 632--635,
  2017.

\bibitem{Gao2020}
Hui Gao, Yu~Xiao, Han Yan, Ye~Tian, Danshi Wang, and Wendong Wang.
\newblock {A learning-based credible participant recruitment strategy for
  mobile crowd sensing}.
\newblock {\em IEEE Internet of Things Journal}, 7(6):5302--5314, 2020.

\bibitem{Cao2019}
Xiaofeng Cao, Yan Li, Jiarong Han, Peng Yang, Feng Lyu, Deke Guo, and Xuemin
  Shen.
\newblock {Online worker selection towards high quality map collection for
  autonomous driving}.
\newblock {\em IEEE Global Communications Conference, GLOBECOM 2019 -
  Proceedings}, 2019.

\bibitem{Liu2023}
Hui Liu, Ciyun Lin, Bowen Gong, and Dayong Wu.
\newblock {Automatic Lane-Level Intersection Map Generation using Low-Channel
  Roadside LiDAR}.
\newblock {\em IEEE/CAA Journal of Automatica Sinica}, 10(5):1209--1222, 2023.

\bibitem{Yu2021}
Changqian Yu, Changxin Gao, Jingbo Wang, Gang Yu, Chunhua Shen, and Nong Sang.
\newblock {BiSeNet V2: Bilateral Network with Guided Aggregation for Real-Time
  Semantic Segmentation}.
\newblock {\em International Journal of Computer Vision}, 129(11):3051--3068,
  2021.

\bibitem{Krizhevsky2012}
Alex Krizhevsky, Ilya Sutskever, and Geoffrey~E. Hinton.
\newblock {ImageNet Classification with Deep Convolutional Neural Networks}.
\newblock {\em Advances in Neural Information Processing Systems 25 (NIPS
  2012)}, 2012.

\bibitem{Geiger2012}
Andreas Geiger, Philip Lenz, and Raquel Urtasun.
\newblock {Are we ready for autonomous driving? the KITTI vision benchmark
  suite}.
\newblock {\em Proceedings of the IEEE Computer Society Conference on Computer
  Vision and Pattern Recognition}, pages 3354--3361, 2012.

\bibitem{Cordts2016}
Marius Cordts, Mohamed Omran, Sebastian Ramos, Timo Rehfeld, Markus Enzweiler,
  Rodrigo Benenson, Uwe Franke, Stefan Roth, and Bernt Schiele.
\newblock {The Cityscapes Dataset for Semantic Urban Scene Understanding}.
\newblock {\em Proceedings of the IEEE Computer Society Conference on Computer
  Vision and Pattern Recognition}, 2016-December:3213--3223, 2016.

\bibitem{Zhou2019}
Bolei Zhou, Hang Zhao, Xavier Puig, Tete Xiao, Sanja Fidler, Adela Barriuso,
  and Antonio Torralba.
\newblock {Semantic Understanding of Scenes Through the ADE20K Dataset}.
\newblock {\em International Journal of Computer Vision}, 127(3):302--321,
  2019.

\bibitem{Long2015}
Jonathan Long, Evan Shelhamer, and Trevor Darrell.
\newblock {Fully Convolutional Networks for Semantic Segmentation}.
\newblock {\em IEEE Transactions on Pattern Analysis and Machine Intelligence},
  39(4):640--651, 2015.

\bibitem{Brostow2009}
Gabriel~J. Brostow, Julien Fauqueur, and Roberto Cipolla.
\newblock {Semantic object classes in video: A high-definition ground truth
  database}.
\newblock {\em Pattern Recognition Letters}, 30(2):88--97, 2009.

\bibitem{JiaDeng2009}
{Jia Deng}, {Wei Dong}, R.~Socher, {Li-Jia Li}, {Kai Li}, and {Li Fei-Fei}.
\newblock {ImageNet: A large-scale hierarchical image database}.
\newblock pages 248--255, 2009.

\bibitem{Badrinarayanan2017}
Vijay Badrinarayanan, Alex Kendall, and Roberto Cipolla.
\newblock {Segnet: A deep convolutional encoder-decoder architecture for image
  segmentation}.
\newblock {\em IEEE transactions on pattern analysis and machine intelligence},
  39(12):2481--2495, 2017.

\bibitem{Chen2018}
Liang~Chieh Chen, George Papandreou, Iasonas Kokkinos, Kevin Murphy, and
  Alan~L. Yuille.
\newblock {DeepLab: Semantic Image Segmentation with Deep Convolutional Nets,
  Atrous Convolution, and Fully Connected CRFs}.
\newblock {\em IEEE Transactions on Pattern Analysis and Machine Intelligence},
  40(4):834--848, 2018.

\bibitem{Yu2016}
Fisher Yu and Vladlen Koltun.
\newblock {Multi-Scale Context Aggregation by Dilated Convolutions}.
\newblock {\em 4th International Conference on Learning Representations
  (ICLR)}, 2016.

\bibitem{Zhao2017}
Hengshuang Zhao, Jianping Shi, Xiaojuan Qi, Xiaogang Wang, and Jiaya Jia.
\newblock {Pyramid scene parsing network}.
\newblock {\em Proceedings - 30th IEEE Conference on Computer Vision and
  Pattern Recognition, CVPR 2017}, 2017-Janua:6230--6239, 2017.

\bibitem{Paszke2016}
Adam Paszke, Abhishek Chaurasia, Sangpil Kim, and Eugenio Culurciello.
\newblock {ENet : A Deep Neural Network Architecture for Real-Time Semantic
  Segmentation}.
\newblock pages 1--10, 2016.

\bibitem{Li2019}
Hanchao Li, Pengfei Xiong, Haoqiang Fan, and Jian Sun.
\newblock {DFANet: Deep feature aggregation for real-time semantic
  segmentation}.
\newblock {\em Proceedings of the IEEE Computer Society Conference on Computer
  Vision and Pattern Recognition}, pages 9514--9523, 2019.

\bibitem{Zhao2018}
Hengshuang Zhao, Xiaojuan Qi, Xiaoyong Shen, Jianping Shi, and Jiaya Jia.
\newblock {ICNet for Real-Time Semantic Segmentation on High-Resolution
  Images}.
\newblock {\em 15th European Conference on Computer Vision (ECCV)}, pages
  418--434, 2018.

\bibitem{Yu2018}
Changqian Yu, Jingbo Wang, Chao Peng, Changxin Gao, Gang Yu, and Nong Sang.
\newblock {BiSeNet: Bilateral segmentation network for real-time semantic
  segmentation}.
\newblock {\em 15th European Conference on Computer Vision (ECCV)}, pages
  334--349, 2018.

\bibitem{Pan2022}
Huihui Pan, Yuanduo Hong, Weichao Sun, and Yisong Jia.
\newblock {Deep Dual-Resolution Networks for Real-Time and Accurate Semantic
  Segmentation of Traffic Scenes}.
\newblock {\em IEEE Transactions on Intelligent Transportation Systems},
  14(8):1--12, 2022.

\bibitem{Xu2022}
Jiacong Xu, Zixiang Xiong, and Shankar~P. Bhattacharyya.
\newblock {PIDNet: A Real-time Semantic Segmentation Network Inspired from PID
  Controller}.
\newblock 2022.

\bibitem{Behley2019}
Jens Behley, Martin Garbade, Andres Milioto, Jan Quenzel, Sven Behnke, Cyrill
  Stachniss, and Jurgen Gall.
\newblock {SemanticKITTI: A dataset for semantic scene understanding of LiDAR
  sequences}.
\newblock {\em Proceedings of the IEEE International Conference on Computer
  Vision}, 2019-October(iii):9296--9306, 2019.

\bibitem{Xu2020}
Chenfeng Xu, Bichen Wu, Zining Wang, Wei Zhan, Peter Vajda, Kurt Keutzer, and
  Masayoshi Tomizuka.
\newblock {SqueezeSegV3: Spatially-Adaptive Convolution for Efficient
  Point-Cloud Segmentation}.
\newblock {\em European Conference on Computer Vision (ECCV)}, pages 1--19,
  2020.

\bibitem{Milioto2019}
Andres Milioto, Ignacio Vizzo, Jens Behley, and Cyrill Stachniss.
\newblock {RangeNet ++: Fast and Accurate LiDAR Semantic Segmentation}.
\newblock {\em IEEE International Conference on Intelligent Robots and
  Systems}, (i):4213--4220, 2019.

\bibitem{Zhang2020}
Yang Zhang, Zixiang Zhou, Philip David, Xiangyu Yue, Zerong Xi, Boqing Gong,
  and Hassan Foroosh.
\newblock {Polarnet: An improved grid representation for online Lidar point
  clouds semantic segmentation}.
\newblock {\em Proceedings of the IEEE Computer Society Conference on Computer
  Vision and Pattern Recognition}, pages 9598--9607, 2020.

\bibitem{Cortinhal2020}
Tiago Cortinhal, George Tzelepis, and Eren {Erdal Aksoy}.
\newblock {SalsaNext: Fast, Uncertainty-Aware Semantic Segmentation of LiDAR
  Point Clouds}.
\newblock 12510 LNCS:207--222, 2020.

\bibitem{Hu2020}
Qingyong Hu, Bo~Yang, Linhai Xie, Stefano Rosa, Yulan Guo, Zhihua Wang, Niki
  Trigoni, and Andrew Markham.
\newblock {Randla-Net: Efficient semantic segmentation of large-scale point
  clouds}.
\newblock {\em Proceedings of the IEEE Computer Society Conference on Computer
  Vision and Pattern Recognition}, pages 11105--11114, 2020.

\bibitem{Thomas2019}
Hugues Thomas, Charles~R. Qi, Jean~Emmanuel Deschaud, Beatriz Marcotegui,
  Francois Goulette, and Leonidas Guibas.
\newblock {KPConv: Flexible and deformable convolution for point clouds}.
\newblock {\em Proceedings of the IEEE International Conference on Computer
  Vision}, 2019-October:6410--6419, 2019.

\bibitem{Choy2019}
Christopher Choy, Junyoung Gwak, and Silvio Savarese.
\newblock {4D spatio-temporal convnets: Minkowski convolutional neural
  networks}.
\newblock {\em Proceedings of the IEEE Computer Society Conference on Computer
  Vision and Pattern Recognition}, 2019-June:3070--3079, 2019.

\bibitem{Zhanglidar2020}
Feihu Zhang, Jin Fang, Benjamin Wah, and Philip Torr.
\newblock {Deep FusionNet for Point Cloud Semantic Segmentation}.
\newblock {\em European Conference on Computer Vision (ECCV)}, 12369
  LNCS(c):644--663, 2020.

\bibitem{Alonso2020}
Inigo Alonso, Luis Riazuelo, Luis Montesano, and Ana~C. Murillo.
\newblock {3D-MiniNet: Learning a 2D Representation from Point Clouds for Fast
  and Efficient 3D LIDAR Semantic Segmentation}.
\newblock {\em IEEE Robotics and Automation Letters}, 5(4):5432--5439, 2020.

\bibitem{Tang2020}
Haotian Tang, Zhijian Liu, Shengyu Zhao, Yujun Lin, Ji~Lin, Hanrui Wang, and
  Song Han.
\newblock {Searching Efficient 3D Architectures with Sparse Point-Voxel
  Convolution}.
\newblock {\em European Conference on Computer Vision (ECCV)}, pages 685--702,
  2020.

\bibitem{Xie2022}
Xing Xie, Lin Bai, and Xinming Huang.
\newblock {Real-Time LiDAR Point Cloud Semantic Segmentation for Autonomous
  Driving}.
\newblock {\em Electronics}, (11):1--13, 2022.

\bibitem{Cheng2021}
Ran Cheng, Ryan Razani, Ehsan Taghavi, Enxu Li, and Bingbing Liu.
\newblock {(AF)2 -S3Net: Attentive Feature Fusion with Adaptive Feature
  Selection for Sparse Semantic Segmentation Network}.
\newblock {\em IEEE/CVF Conference on Computer Vision and Pattern Recognition
  (CVPR)}, 2021.

\bibitem{Yan2022}
Xu~Yan, Jiantao Gao, Chaoda Zheng, Chao Zheng, Ruimao Zhang, Shuguang Cui, and
  Zhen Li.
\newblock {2DPASS: 2D Priors Assisted Semantic Segmentation on LiDAR Point
  Clouds}.
\newblock {\em European Conference on Computer Vision ECCV 2022}, ]:1--19,
  2022.

\bibitem{Chen2015}
Liang-chieh Chen, Iasonas Kokkinos, Kevin Murphy, and Alan~L Yuille.
\newblock {Semantic image segmentation with deep convolutional nets and fully
  connected CRFs}.
\newblock {\em IEEE transactions on pattern analysis and machine intelligence},
  40(4):834----848, 2015.

\bibitem{Chen2018b}
Liang~Chieh Chen, Yukun Zhu, George Papandreou, Florian Schroff, and Hartwig
  Adam.
\newblock {Encoder-decoder with atrous separable convolution for semantic image
  segmentation}.
\newblock {\em European Conference on Computer Vision (ECCV)}, pages 833--851,
  2018.

\bibitem{Zhao2018b}
Hengshuang Zhao, Yi~Zhang, Shu Liu, and Jianping Shi.
\newblock {PSANet: Point-wise Spatial Attention Network for Scene Parsing}.
\newblock {\em 15th European Conference on Computer Vision (ECCV)}, 2018.

\bibitem{Yu2020}
Changqian Yu, Jingbo Wang, Changxin Gao, Gang Yu, Chunhua Shen, and Nong Sang.
\newblock {Context prior for scene segmentation}.
\newblock {\em Proceedings of the IEEE Computer Society Conference on Computer
  Vision and Pattern Recognition}, pages 12413--12422, 2020.

\bibitem{Lin2017}
Guosheng Lin, Anton Milan, Chunhua Shen, and Ian Reid.
\newblock {RefineNet: Multi-path refinement networks for high-resolution
  semantic segmentation}.
\newblock {\em Proceedings - 30th IEEE Conference on Computer Vision and
  Pattern Recognition, CVPR 2017}, pages 5168--5177, 2017.

\bibitem{Peng2017}
Chao Peng, Xiangyu Zhang, Gang Yu, Guiming Luo, and Jian Sun.
\newblock {Large Kernel Matters - Improve Semantic Segmentation by Global
  Convolutional Network}.
\newblock {\em Proceedings - 30th IEEE Conference on Computer Vision and
  Pattern Recognition, CVPR 2017}, pages 1743--1751, 2017.

\bibitem{Mazzini2018}
Davide Mazzini.
\newblock {Guided Upsampling Network for Real-Time Semantic Segmentation}.
\newblock {\em 29th British Machine Vision Conference (BMVC)}, pages 1--12,
  2018.

\bibitem{Romera2018}
Eduardo Romera, Jose~M. Alvarez, Luis~M. Bergasa, and Roberto Arroyo.
\newblock {ERFNet: Efficient Residual Factorized ConvNet for Real-Time Semantic
  Segmentation}.
\newblock {\em IEEE Transactions on Intelligent Transportation Systems},
  19(1):263--272, 2018.

\bibitem{Chollet2017}
Francois Chollet.
\newblock {Xception: Deep Learning with Depthwise Separable Convolutions}.
\newblock {\em Proceedings - 30th IEEE Conference on Computer Vision and
  Pattern Recognition (CVPR)}, 2017.

\bibitem{Poudel2019}
Rudra P~K Poudel, Stephan Liwicki, and Roberto Cipolla.
\newblock {Fast-SCNN: Fast Semantic Segmentation Network}.
\newblock {\em 30th British Machine Vision Conference (BMVC)}, 2019.

\bibitem{Minaee2021}
Shervin Minaee, Yuri~Y. Boykov, Fatih Porikli, Antonio~J. Plaza, Nasser
  Kehtarnavaz, and Demetri Terzopoulos.
\newblock {Image Segmentation Using Deep Learning: A Survey}.
\newblock {\em IEEE Transactions on Pattern Analysis and Machine Intelligence},
  pages 1--22, 2021.

\bibitem{Tsushima2020}
F.~Tsushima, N.~Kishimoto, Y.~Okada, and W.~Che.
\newblock {Creation of high definition map for autonomous driving}.
\newblock {\em International Archives of the Photogrammetry, Remote Sensing and
  Spatial Information Sciences - ISPRS Archives}, 43(B4):415--420, 2020.

\bibitem{Jiang2021}
Peng Jiang, Philip Osteen, Maggie Wigness, and Srikanth Saripalli.
\newblock {RELLIS-3D Dataset: Data, Benchmarks and Analysis}.
\newblock pages 1110--1116, 2021.

\bibitem{Garbade2021}
Martin Garbade, Jan Quenzel, Andres Miloto, Sven Behnke, Jurgen Gall, and
  {Cyrill Stachniss}.
\newblock {Towards 3D LiDAR-based semantic scene understanding of 3D point
  cloud sequences: The SemanticKITTI Dataset}.
\newblock {\em The International Journal on Robotics Research}, 40:959--967,
  2021.

\bibitem{Wu2018}
Bichen Wu, Alvin Wan, Xiangyu Yue, and Kurt Keutzer.
\newblock {SqueezeSeg: Convolutional Neural Nets with Recurrent CRF for
  Real-Time Road-Object Segmentation from 3D LiDAR Point Cloud}.
\newblock {\em Proceedings - IEEE International Conference on Robotics and
  Automation}, pages 1887--1893, 2018.

\bibitem{Wu2019}
Bichen Wu, Xuanyu Zhou, Sicheng Zhao, Xiangyu Yue, and Kurt Keutzer.
\newblock {SqueezeSegV2: Improved model structure and unsupervised domain
  adaptation for road-object segmentation from a LiDAR point cloud}.
\newblock {\em Proceedings - IEEE International Conference on Robotics and
  Automation}, 2019-May:4376--4382, 2019.

\bibitem{Redmon2018}
Joseph Redmon and Ali Farhadi.
\newblock {YOLOv3: An Incremental Improvement}.
\newblock 2018.

\bibitem{Qi2017}
Charles~R. Qi, Hao Su, Kaichun Mo, and Leonidas~J. Guibas.
\newblock {PointNet: Deep learning on point sets for 3D classification and
  segmentation}.
\newblock {\em Proceedings - 30th IEEE Conference on Computer Vision and
  Pattern Recognition, CVPR 2017}, 2017-January:77--85, 2017.

\bibitem{Aksoy2020}
Eren~Erdal Aksoy, Saimir Baci, and Selcuk Cavdar.
\newblock {SalsaNet: Fast Road and Vehicle Segmentation in LiDAR Point Clouds
  for Autonomous Driving}.
\newblock {\em IEEE Intelligent Vehicles Symposium, Proceedings}, pages
  926--932, 2020.

\bibitem{Maturana2015}
Daniel Maturana and Sebastian Scherer.
\newblock {VoxNet: A 3D Convolutional Neural Network for real-time object
  recognition}.
\newblock {\em IEEE International Conference on Intelligent Robots and
  Systems}, 2015-December:922--928, 2015.

\bibitem{Chang2015}
Angel~X. Chang, Thomas Funkhouser, Leonidas Guibas, Pat Hanrahan, Qixing Huang,
  Zimo Li, Silvio Savarese, Manolis Savva, Shuran Song, Hao Su, Jianxiong Xiao,
  Li~Yi, and Fisher Yu.
\newblock {ShapeNet: An Information-Rich 3D Model Repository}.
\newblock 2015.

\bibitem{Wang2020}
Zongji Wang and Feng Lu.
\newblock {VoxSegNet: Volumetric CNNs for Semantic Part Segmentation of 3D
  Shapes}.
\newblock {\em IEEE Transactions on Visualization and Computer Graphics},
  26(9):2919--2930, 2020.

\bibitem{Qi2017a}
Charles~R. Qi, Li~Yi, Hao Su, and Leonidas~J. Guibas.
\newblock {PointNet++: Deep hierarchical feature learning on point sets in a
  metric space}.
\newblock {\em Advances in Neural Information Processing Systems},
  2017-Decem:5100--5109, 2017.

\bibitem{Gao2021}
Biao Gao, Yancheng Pan, Chengkun Li, Sibo Geng, and Huijing Zhao.
\newblock {Are We Hungry for 3D LiDAR Data for Semantic Segmentation? A Survey
  of Datasets and Methods}.
\newblock {\em IEEE Transactions on Intelligent Transportation Systems}, pages
  1--19, 2021.

\bibitem{Lombacher2016}
Jakob Lombacher, Markus Hahn, J{\"{u}}rgen Dickmann, and Christian
  W{\"{o}}hler.
\newblock {Potential of radar for static object classification using deep
  learning methods}.
\newblock {\em 2016 IEEE MTT-S International Conference on Microwaves for
  Intelligent Mobility, ICMIM 2016}, pages 15--18, 2016.

\bibitem{Schumann2018}
Ole Schumann, Markus Hahn, J{\"{u}}rgen Dickmann, and Christian W{\"{o}}hler.
\newblock {Semantic segmentation on radar point clouds}.
\newblock {\em 2018 21st International Conference on Information Fusion, FUSION
  2018}, pages 2179--2186, 2018.

\bibitem{Prophet2019}
Robert Prophet, Gang Li, Christian Sturm, and Martin Vossiek.
\newblock {Semantic segmentation on automotive radar maps}.
\newblock {\em IEEE Intelligent Vehicles Symposium, Proceedings},
  2019-June(Iv):756--763, 2019.

\bibitem{Lombacher2017}
Jakob Lombacher, Kilian Laudt, Markus Hahn, Jurgen Dickmann, and Christian
  Wohler.
\newblock {Semantic radar grids}.
\newblock {\em IEEE Intelligent Vehicles Symposium, Proceedings},
  (Iv):1170--1175, 2017.

\bibitem{Zhou2020}
Taohua Zhou, Mengmeng Yang, Kun Jiang, Henry Wong, and Diange Yang.
\newblock {MMW radar-based technologies in autonomous driving: A review}.
\newblock {\em Sensors (Switzerland)}, 20(24):1--21, 2020.

\bibitem{Braun2021}
M.~Braun, A.~Cennamo, M.~Schoeler, K.~Kollek, and A.~Kummert.
\newblock {Semantic Segmentation of Radar Detections using Convolutions on
  Point Clouds}.
\newblock {\em Journal of Physics: Conference Series}, 1924(1), 2021.

\bibitem{Pohlen2017}
Tobias Pohlen, Alexander Hermans, Markus Mathias, and Bastian Leibe.
\newblock {Full-resolution residual networks for semantic segmentation in
  street scenes}.
\newblock {\em Proceedings - 30th IEEE Conference on Computer Vision and
  Pattern Recognition, CVPR 2017}, 2017-January:3309--3318, 2017.

\bibitem{Scheiner2018}
Nicolas Scheiner, Nils Appenrodt, J{\"{u}}rgen DIckmann, and Bernhard Sick.
\newblock {Radar-based Feature Design and Multiclass Classification for Road
  User Recognition}.
\newblock {\em IEEE Intelligent Vehicles Symposium, Proceedings},
  2018-June:779--786, 2018.

\bibitem{Ouaknine2021}
Arthur Ouaknine, Alasdair Newson, Patrick P{\'{e}}rez, Florence Tupin, and
  Julien Rebut.
\newblock {Multi-View Radar Semantic Segmentation}.
\newblock {\em Proceedings of the IEEE International Conference on Computer
  Vision}, pages 15651--15660, 2021.

\bibitem{Bao2022}
Zhibin Bao, Sabir Hossain, Haoxiang Lang, and Xianke Lin.
\newblock {High-Definition Map Generation Technologies For Autonomous Driving}.
\newblock pages 1--21, 2022.

\bibitem{Davison2007}
Andrew~J. Davison, Ian~D. Reid, Nicholas~D. Molton, and Olivier Stasse.
\newblock {MonoSLAM: Real-time single camera SLAM}.
\newblock {\em IEEE Transactions on Pattern Analysis and Machine Intelligence},
  29(6):1052--1067, 2007.

\bibitem{Kim_mono2015}
Hanme Kim and Hyon Lim.
\newblock {SceneLib2 - MonoSLAM open-source library}, 2015.

\bibitem{Engel2013a}
Jakob Engel, Jurgen Sturm, and Daniel Cremers.
\newblock {LSD-SLAM: Large-Scale Direct Monocular SLAM}.
\newblock {\em Proceedings of the IEEE International Conference on Computer
  Vision}, pages 1449--1456, 2013.

\bibitem{Engel2015}
Jakob Engel, Jorg Stuckler, and Daniel Cremers.
\newblock {Large-scale direct SLAM for omnidirectional cameras}.
\newblock {\em IEEE International Conference on Intelligent Robots and
  Systems}, 2015-Decem:141--148, 2015.

\bibitem{Engel2018}
Jakob Engel, Vladlen Koltun, and Daniel Cremers.
\newblock {Direct Sparse Odometry}.
\newblock {\em IEEE Transactions on Pattern Analysis and Machine Intelligence},
  40(3):611--625, 2018.

\bibitem{Forster2014}
Christian Forster, Matia Pizzoli, and Davide Scaramuzza.
\newblock {SVO: Fast Semi-Direct Monocular Visual Odometry}.
\newblock {\em IEEE International Conference on Robotics and Automation
  (ICRA)}, 2014.

\bibitem{Forster2016}
Christian Forster, Zichao Zhang, Michael Gassner, Manuel Werlberger, and Davide
  Scaramuzza.
\newblock {SVO: Semidirect Visual Odometry for Monocular and Multicamera
  Systems}.
\newblock {\em IEEE Transactions on Robotics}, 33(2):249--265, 2016.

\bibitem{Mourikis2007}
Anastasios~I. Mourikis and Stergios~I. Roumeliotis.
\newblock {A multi-state constraint Kalman filter for vision-aided inertial
  navigation}.
\newblock {\em Proceedings - IEEE International Conference on Robotics and
  Automation}, (April):3565--3572, 2007.

\bibitem{Zhu2017}
Alex~Zihao Zhu, Nikolay Atanasov, and Kostas Daniilidis.
\newblock {Event-based visual inertial odometry}.
\newblock {\em Proceedings - 30th IEEE Conference on Computer Vision and
  Pattern Recognition, CVPR 2017}, 2017-January:5816--5824, 2017.

\bibitem{Zubizarreta2019}
Jon Zubizarreta, Iker Aguinaga, and Jose Maria~Martinez Montiel.
\newblock {Direct Sparse Mapping}.
\newblock {\em IEEE Transactions on Robotics}, 36(4):1363--1370, 2019.

\bibitem{Leutenegger2015}
Stefan Leutenegger, Simon Lynen, Michael Bosse, Roland Siegwart, and Paul
  Furgale.
\newblock {Keyframe-Based Visual-Inertial Odometry Using Non Linear
  Optimization}.
\newblock {\em The International Journal of Robotics Research}, 34(3):314--334,
  2015.

\bibitem{Qin2018a}
Tong Qin, Peiliang Li, and Shaojie Shen.
\newblock {VINS-Mono: A Robust and Versatile Monocular Visual-Inertial State
  Estimator}.
\newblock {\em IEEE Transactions on Robotics}, 34(4):1004--1020, 2018.

\bibitem{Qin2018}
Tong Qin and Shaojie Shen.
\newblock {Online Temporal Calibration for Monocular Visual-Inertial Systems}.
\newblock {\em IEEE International Conference on Intelligent Robots and
  Systems}, pages 3662--3669, 2018.

\bibitem{Geneva2020}
Patrick Geneva, Kevin Eckenhoff, Woosik Lee, Yulin Yang, and Guoquan Huang.
\newblock {OpenVINS: A Research Platform for Visual-Inertial Estimation}.
\newblock {\em Proceedings - IEEE International Conference on Robotics and
  Automation}, pages 4666--4672, 2020.

\bibitem{Mur-Artal2015}
Raul Mur-Artal, J.~M.M. Montiel, and Juan~D. Tardos.
\newblock {ORB-SLAM: A Versatile and Accurate Monocular SLAM System}.
\newblock {\em IEEE Transactions on Robotics}, 31(5):1147--1163, 2015.

\bibitem{Mur-Artal2017}
Raul Mur-Artal and Juan~D. Tardos.
\newblock {ORB-SLAM2: An Open-Source SLAM System for Monocular, Stereo, and
  RGB-D Cameras}.
\newblock {\em IEEE Transactions on Robotics}, 33(5):1255--1262, 2017.

\bibitem{Mur-Artal2017a}
Raul Mur-Artal and Juan~D. Tardos.
\newblock {Visual-Inertial Monocular SLAM with Map Reuse}.
\newblock {\em IEEE Robotics and Automation Letters}, 2(2):796--803, 2017.

\bibitem{Campos2021}
Carlos Campos, Richard Elvira, Juan~J.Gomez Rodriguez, Jose~M.M. Montiel, and
  Juan~D. Tardos.
\newblock {ORB-SLAM3: An Accurate Open-Source Library for Visual,
  Visual-Inertial, and Multimap SLAM}.
\newblock {\em IEEE Transactions on Robotics}, 37(6):1874--1890, 2021.

\bibitem{Cvisic2015}
Igor Cvi{\v{s}}i{\'{c}} and Ivan Petrovi{\'{c}}.
\newblock {Stereo odometry based on careful feature selection and tracking}.
\newblock {\em 2015 European Conference on Mobile Robots, ECMR 2015 -
  Proceedings}, pages 0--5, 2015.

\bibitem{Cvisic2018}
Igor Cvi{\v{s}}i{\'{c}}, Josip {\'{C}}esi{\'{c}}, Ivan Markovi{\'{c}}, and Ivan
  Petrovi{\'{c}}.
\newblock {SOFT-SLAM: Computationally efficient stereo visual simultaneous
  localization and mapping for autonomous unmanned aerial vehicles}.
\newblock {\em Journal of Field Robotics}, 35(4):578--595, 2018.

\bibitem{Cvisic2022}
Igor Cvi{\v{s}}i{\'{c}}, Ivan Markovi{\'{c}}, and Ivan Petrovi{\'{c}}.
\newblock {Enhanced calibration of camera setups for high-performance visual
  odometry}.
\newblock {\em Robotics and Autonomous Systems}, 155:104189, 2022.

\bibitem{Cvisic2023}
Igor Cvisic, Ivan Markovic, and Ivan Petrovic.
\newblock {SOFT2: Stereo Visual Odometry for Road Vehicles Based on a
  Point-to-Epipolar-Line Metric}.
\newblock {\em IEEE Transactions on Robotics}, 39(1):273--288, 2023.

\bibitem{Lu2021}
Yongqiang Lu, Hongjie Ma, Edward Smart, and Hui Yu.
\newblock {Real-Time Performance-Focused Localization Techniques for Autonomous
  Vehicle: A Review}.
\newblock {\em IEEE Transactions on Intelligent Transportation Systems}, pages
  1--19, 2021.

\bibitem{Qin2017}
Hua Qin, Yang Peng, and Wensheng Zhang.
\newblock {Vehicles on RFID: Error-cognitive vehicle localization in GPS-less
  environments}.
\newblock {\em IEEE Transactions on Vehicular Technology}, 66(11):9943--9957,
  2017.

\bibitem{Brebler2016}
Julia Bre{\ss}ler and Marcus Obst.
\newblock {GNSS positioning in non-line-of-sight context: A survey}.
\newblock {\em Advances in Science, Technology and Engineering Systems
  Journal}, 2(3):722--731, 2017.

\bibitem{Park2017}
Sul~Gee Park and Deuk~Jae Cho.
\newblock {Smart framework for GNSS-based navigation in urban environments}.
\newblock {\em International Journal of Satellite Communications and
  Networking}, 35(2):123--137, 2017.

\bibitem{Schreiber2016}
Markus Schreiber, Hendrik Konigshof, Andre~Marcel Hellmund, and Christoph
  Stiller.
\newblock {Vehicle localization with tightly coupled GNSS and visual odometry}.
\newblock {\em IEEE Intelligent Vehicles Symposium, Proceedings},
  2016-August(Iv):858--863, 2016.

\bibitem{Amini2014}
Arghavan Amini, Reza~Monir Vaghefi, Jesus~M. {De La Garza}, and R.~Michael
  Buehrer.
\newblock {Improving GPS-based vehicle positioning for Intelligent
  Transportation Systems}.
\newblock {\em IEEE Intelligent Vehicles Symposium, Proceedings},
  (Iv):1023--1029, 2014.

\bibitem{Adjrad2017}
Mounir Adjrad and Paul~D. Groves.
\newblock {Enhancing Least Squares GNSS Positioning with 3D Mapping without
  Accurate Prior Knowledge}.
\newblock {\em Navigation, Journal of the Institute of Navigation},
  64(1):75--91, 2017.

\bibitem{Kumar2014}
Rakesh Kumar and Mark~G. Petovello.
\newblock {A novel GNSS positioning technique for improved accuracy in Urban
  canyon scenarios using 3D city model}.
\newblock {\em 27th International Technical Meeting of the Satellite Division
  of the Institute of Navigation, ION GNSS 2014}, 3:2139--2148, 2014.

\bibitem{Jackson2018}
John Jackson, Brian Davis, and Demoz Gebre-Egziabher.
\newblock {A performance assessment of low-cost RTK GNSS receivers}.
\newblock {\em 2018 IEEE/ION Position, Location and Navigation Symposium, PLANS
  2018 - Proceedings}, pages 642--649, 2018.

\bibitem{Wang2016}
Shiyao Wang, Zhidong Deng, and Gang Yin.
\newblock {An accurate GPS-IMU/DR data fusion method for driverless car based
  on a set of predictive models and grid constraints}.
\newblock {\em Sensors (Switzerland)}, 16(3), 2016.

\bibitem{Azree2018}
W.~M. H.~Wan Azree, M.~A.~Abdul Rahman, and H.~Zamzuri.
\newblock {Vehicle localization using wheel speed sensor (WSS) and inertial
  measurement unit (IMU)}.
\newblock {\em Journal of the Society of Automotive Engineers Malaysia},
  2(1):43--59, 2018.

\bibitem{Belhajem2018}
Ikram Belhajem, Yann {Ben Maissa}, and Ahmed Tamtaoui.
\newblock {Improving low cost sensor based vehicle positioning with Machine
  Learning}.
\newblock {\em Control Engineering Practice}, 74(March):168--176, 2018.

\bibitem{Ndjeng2009}
Alexandre~Ndjeng Ndjeng, Dominique Gruyer, and S{\'{e}}bastien Glaser.
\newblock {New likelihood updating for the IMM approach application to outdoor
  vehicles localization}.
\newblock {\em 2009 IEEE/RSJ International Conference on Intelligent Robots and
  Systems, IROS 2009}, (November):1223--1228, 2009.

\bibitem{Gim2018}
Juhui Gim and Changsun Ahn.
\newblock {IMU-based virtual road profile sensor for vehicle localization}.
\newblock {\em Sensors (Switzerland)}, 18(10), 2018.

\bibitem{Civera2008}
Javier Civera, Andrew~J. Davison, and Jos{\'{e}}~Mar{\'{i}}a {Mart{\'{i}}nez
  Montiel}.
\newblock {Inverse Depth Parametrization for Monocular SLAM}.
\newblock {\em IEEE Transactions on Robotics}, 75(5):932--945, 2008.

\bibitem{Fischler1981}
Martin~A Fischler and Robert~C Bolles.
\newblock {Random Sample Paradigm for Model Consensus: A Apphcatlons to Image
  Fitting with Analysis and Automated Cartography}.
\newblock {\em Graphics and Image Processing}, 24(6):381--395, 1981.

\bibitem{Civera2010}
Javier Civera, Oscar~G. Grasa, Andrew~J. Davison, and J.~M.M. Montiel.
\newblock {1-Point RANSAC for extended Kalman filtering: Application to
  real-time structure from motion and visual odometry}.
\newblock {\em Journal of Field Robotics}, 27(5):609--631, 2010.

\bibitem{Strasdat2010}
Hauke Strasdat, J.~M.M. Montiel, and Andrew~J. Davison.
\newblock {Real-time monocular SLAM: Why filter?}
\newblock {\em Proceedings - IEEE International Conference on Robotics and
  Automation}, pages 2657--2664, 2010.

\bibitem{Strasdat2011}
Hauke Strasdat, J.~M.M. Montiel, and Andrew~J. Davison.
\newblock {Scale drift-aware large scale monocular SLAM}.
\newblock {\em Robotics: Science and Systems}, 6:73--80, 2011.

\bibitem{Newcombe2011}
Richard~A. Newcombe, Steven~J. Lovegrove, and Andrew~J. Davison.
\newblock {DTAM: Dense tracking and mapping in real-time}.
\newblock {\em Proceedings of the IEEE International Conference on Computer
  Vision}, pages 2320--2327, 2011.

\bibitem{Tam2013}
Gary~K.L. Tam, Zhi~Quan Cheng, Yu~Kun Lai, Frank~C. Langbein, Yonghuai Liu,
  David Marshall, Ralph~R. Martin, Xian~Fang Sun, and Paul~L. Rosin.
\newblock {Registration of 3d point clouds and meshes: A survey from rigid to
  Nonrigid}.
\newblock {\em IEEE Transactions on Visualization and Computer Graphics},
  19(7):1199--1217, 2013.

\bibitem{Zhang1994}
Zhengyou Zhang.
\newblock {Iterative point matching for registration of free-form curves and
  surfaces}.
\newblock {\em International Journal of Computer Vision}, 13(2):119--152, 1994.

\bibitem{Censi2008}
Andrea Censi.
\newblock {An ICP variant using a point-to-line metric}.
\newblock {\em Proceedings - IEEE International Conference on Robotics and
  Automation}, pages 19--25, 2008.

\bibitem{Low2004}
Kl~Low.
\newblock {Linear Least-squares Optimization for Point-to-plane ICP Surface
  Registration}.
\newblock {\em Chapel Hill, University of North Carolina}, (February):2--4,
  2004.

\bibitem{Segal2009}
Aleksandr~V Segal, Dirk Haehnel, and Sebastian Thrun.
\newblock {Generalized-ICP}.
\newblock {\em Robotics: Science and Systems Conference (RSS)}, 2:435, 2009.

\bibitem{Mendes2016}
Ellon Mendes, Pierrick Koch, and Simon Lacroix.
\newblock {ICP-based pose-graph SLAM}.
\newblock In {\em 2016 IEEE International Symposium on Safety, Security, and
  Rescue Robotics (SSRR)}, pages 195--200. IEEE, oct 2016.

\bibitem{Kuramachi2015}
Ryo Kuramachi, Akihito Ohsato, Yoko Sasaki, and Hiroshi Mizoguchi.
\newblock {G-ICP SLAM: An odometry-free 3D mapping system with robust 6DoF pose
  estimation}.
\newblock {\em 2015 IEEE International Conference on Robotics and Biomimetics,
  IEEE-ROBIO 2015}, pages 176--181, 2015.

\bibitem{Kovalenko2019}
Dmitri Kovalenko, Mikhail Korobkin, and Andrey Minin.
\newblock {Sensor aware lidar odometry}.
\newblock {\em 2019 European Conference on Mobile Robots, ECMR 2019 -
  Proceedings}, 2019.

\bibitem{Dellenbach2021}
Pierre Dellenbach, Jean-Emmanuel Deschaud, Bastien Jacquet, and Fran{\c{c}}ois
  Goulette.
\newblock {CT-ICP: Real-time Elastic LiDAR Odometry with Loop Closure}.
\newblock 2021.

\bibitem{Palieri2021}
Matteo Palieri, Benjamin Morrell, Abhishek Thakur, Kamak Ebadi, Jeremy Nash,
  Arghya Chatterjee, Christoforos Kanellakis, Luca Carlone, Cataldo
  Guaragnella, and Ali~Akbar Agha-Mohammadi.
\newblock {LOCUS: A Multi-Sensor Lidar-Centric Solution for High-Precision
  Odometry and 3D Mapping in Real-Time}.
\newblock {\em IEEE Robotics and Automation Letters}, 6(2):421--428, 2021.

\bibitem{Chen2022}
Kenny Chen, Brett~T. Lopez, Ali-akbar Agha-mohammadi, and Ankur Mehta.
\newblock {Direct LiDAR Odometry: Fast Localization with Dense Point Clouds}.
\newblock {\em IEEE Robotics and Automation Letters}, pages 1--8, 2022.

\bibitem{Zhang2014}
Ji~Zhang and Sanjiv Singh.
\newblock {Low-drift and real-time lidar odometry and mapping}.
\newblock {\em Robotics: Science and Systems Conference (RSS)}, 41(2):401--416,
  2014.

\bibitem{Ye2019}
Haoyang Ye, Yuying Chen, and Ming Liu.
\newblock {Tightly coupled 3D Lidar inertial odometry and mapping}.
\newblock {\em Proceedings - IEEE International Conference on Robotics and
  Automation}, 2019-May:3144--3150, 2019.

\bibitem{Shan2020a}
Tixiao Shan, Brendan Englot, Drew Meyers, Wei Wang, Carlo Ratti, and Daniela
  Rus.
\newblock {LVI-SAM: Tightly-coupled Lidar-Visual-Inertial Odometry via
  Smoothing and Mapping}.
\newblock {\em IEEE International Conference on Intelligent Robots and
  Systems}, pages 5135--5142, 2020.

\bibitem{Pan2021}
Yue Pan, Pengchuan Xiao, Yujie He, Zhenlei Shao, and Zesong Li.
\newblock {MULLS: Versatile LiDAR SLAM via Multi-metric Linear Least Square}.
\newblock pages 11633--11640, 2021.

\bibitem{Shan2020b}
Tixiao Shan, Brendan Englot, Drew Meyers, Wei Wang, Carlo Ratti, and Daniela
  Rus.
\newblock {LIO-SAM: Tightly-coupled Lidar Inertial Odometry via Smoothing and
  Mapping}.
\newblock {\em IEEE International Conference on Intelligent Robots and
  Systems}, pages 5135--5142, 2020.

\bibitem{Shan2018}
Tixiao Shan and Brendan Englot.
\newblock {LeGO-LOAM: Lightweight and Ground-Optimized Lidar Odometry and
  Mapping on Variable Terrain}.
\newblock {\em IEEE International Conference on Intelligent Robots and
  Systems}, pages 4758--4765, 2018.

\bibitem{Chou2021}
Chih-Chung Chou and Cheng-Fu Chou.
\newblock {Efficient and Accurate Tightly-Coupled Visual-Lidar SLAM}.
\newblock {\em IEEE Transactions on Intelligent Transportation Systems}, pages
  1--15, 2021.

\bibitem{Zheng2021}
Xin Zheng and Jianke Zhu.
\newblock {Efficient LiDAR Odometry for Autonomous Driving}.
\newblock {\em IEEE Robotics and Automation Letters}, 6(4):8458--8465, 2021.

\bibitem{Dellenbach2022}
Pierre Dellenbach, Jean~Emmanuel Deschaud, Bastien Jacquet, and Francois
  Goulette.
\newblock {CT-ICP: Real-time Elastic LiDAR Odometry with Loop Closure}.
\newblock {\em Proceedings - IEEE International Conference on Robotics and
  Automation}, pages 5580--5586, 2022.

\bibitem{Nicolai2016}
Austin Nicolai, Ryan Skeele, Christopher Eriksen, and Geoffrey~A Hollinger.
\newblock {Deep Learning for Laser Based Odometry Estimation}.
\newblock {\em Science and Systems Conf. Workshop on Limits and Potentials of
  Deep Learning in Robotics}, 2016.

\bibitem{Wang2017}
Sen Wang, Ronald Clark, Hongkai Wen, and Niki Trigoni.
\newblock {DeepVO: Towards end-to-end visual odometry with deep Recurrent
  Convolutional Neural Networks}.
\newblock {\em Proceedings - IEEE International Conference on Robotics and
  Automation}, pages 2043--2050, 2017.

\bibitem{Elbaz2017}
Gil Elbaz, Tamar Avraham, and Anath Fischer.
\newblock {3D point cloud registration for localization using a deep neural
  network auto-encoder}.
\newblock {\em Proceedings - 30th IEEE Conference on Computer Vision and
  Pattern Recognition, CVPR 2017}, 2017-January:2472--2481, 2017.

\bibitem{Yin2018}
Huan Yin, Li~Tang, Xiaqing DIng, Yue Wang, and Rong Xiong.
\newblock {LocNet: Global Localization in 3D Point Clouds for Mobile Vehicles}.
\newblock {\em IEEE Intelligent Vehicles Symposium, Proceedings},
  2018-June:728--733, 2018.

\bibitem{Cho2019}
Younggun Cho, Giseop Kim, and Ayoung Kim.
\newblock {DeepLO: Geometry-Aware Deep LiDAR Odometry}.
\newblock 2019.

\bibitem{Horn2020}
Markus Horn, Nico Engel, Vasileios Belagiannis, Michael Buchholz, and Klaus
  Dietmayer.
\newblock {DeepCLR: Correspondence-Less Architecture for Deep End-to-End Point
  Cloud Registration}.
\newblock {\em 2020 IEEE 23rd International Conference on Intelligent
  Transportation Systems, ITSC 2020}, 2020.

\bibitem{Adis2021}
Philipp Adis, Nicolas Horst, and Mathias Wien.
\newblock {D3dlo: Deep 3d Lidar Odometry}.
\newblock pages 3128--3132, 2021.

\bibitem{Mohamed2019}
Sherif~A.S. Mohamed, Mohammad~Hashem Haghbayan, Tomi Westerlund, Jukka
  Heikkonen, Hannu Tenhunen, and Juha Plosila.
\newblock {A Survey on Odometry for Autonomous Navigation Systems}.
\newblock {\em IEEE Access}, 7:97466--97486, 2019.

\bibitem{Schuster2016}
F.~Schuster, M.~Worner, C.~G. Keller, M.~Haueis, and C.~Curio.
\newblock {Robust localization based on radar signal clustering}.
\newblock {\em IEEE Intelligent Vehicles Symposium, Proceedings},
  2016-August(Iv):839--844, 2016.

\bibitem{Yoneda2018}
Keisuke Yoneda, Naoya Hashimoto, Ryo Yanase, Mohammad Aldibaja, and Naoki
  Suganuma.
\newblock {Vehicle Localization using 76GHz Omnidirectional Millimeter-Wave
  Radar for Winter Automated Driving}.
\newblock {\em IEEE Intelligent Vehicles Symposium, Proceedings},
  2018-June(Iv):971--977, 2018.

\bibitem{Holder2019}
Martin Holder, Sven Hellwig, and Hermann Winner.
\newblock {Real-time pose graph SLAM based on radar}.
\newblock {\em IEEE Intelligent Vehicles Symposium, Proceedings},
  2019-June(Iv):1145--1151, 2019.

\bibitem{Adolfsson2021}
Daniel Adolfsson, Martin Magnusson, Anas Alhashimi, Achim~J. Lilienthal, and
  Henrik Andreasson.
\newblock {CFEAR Radarodometry-Conservative Filtering for Efficient and
  Accurate Radar Odometry}.
\newblock {\em IEEE International Conference on Intelligent Robots and
  Systems}, pages 5462--5469, 2021.

\bibitem{Burnett2021}
Keenan Burnett, David~J. Yoon, Angela~P. Schoellig, and Timothy~D. Barfoot.
\newblock {Radar Odometry Combining Probabilistic Estimation and Unsupervised
  Feature Learning}.
\newblock {\em Robotics: Science and Systems}, 2021.

\bibitem{Quddus2007}
Mohammed~A. Quddus, Washington~Y. Ochieng, and Robert~B. Noland.
\newblock {Current map-matching algorithms for transport applications:
  State-of-the art and future research directions}.
\newblock {\em Transportation Research Part C: Emerging Technologies},
  15(5):312--328, 2007.

\bibitem{Kubicka2018}
Matej Kubicka, Hugues Mounier, and Silviu-lulian Niculescu.
\newblock {Comparative Study and Application-Oriented Classification of
  Vehicular}.
\newblock {\em IEEE Intelligent Transportation Systems Magazine}, (Summer
  2018):150--166, 2018.

\bibitem{White2000}
Christopher~E. White, David Bernstein, and Alain~L. Kornhauser.
\newblock {Some map matching algorithms for personal navigation assistants}.
\newblock {\em Transportation Research Part C: Emerging Technologies},
  8(1-6):91--108, 2000.

\bibitem{Bernstein1996}
David Bernstein and Alain Kornhauser.
\newblock {An Introduction to Map Matching for Personal Navigation Assistants}.
\newblock 24064(August):587--602, 1996.

\bibitem{Chen2011}
Daniel Chen, Anne Driemel, Leonidas~J. Guibas, Andy Nguyen, and Carola Wenk.
\newblock {Approximate map matching with respect to the Fr{\'{e}}chet
  distance}.
\newblock {\em 2011 Proceedings of the 13th Workshop on Algorithm Engineering
  and Experiments, ALENEX 2011}, pages 75--83, 2011.

\bibitem{Schindler2013}
Andreas Schindler.
\newblock {Vehicle Self-localization with High-Precision Digital Maps}.
\newblock 2013.

\bibitem{Wolcott2014}
Ryan~W. Wolcott and Ryan~M. Eustice.
\newblock {Visual localization within LIDAR maps for automated urban driving}.
\newblock {\em IEEE International Conference on Intelligent Robots and
  Systems}, (Iros):176--183, 2014.

\bibitem{Quddus2015}
Mohammed Quddus and Simon Washington.
\newblock {Shortest path and vehicle trajectory aided map-matching for low
  frequency GPS data}.
\newblock {\em Transportation Research Part C: Emerging Technologies},
  55:328--339, 2015.

\bibitem{Caselitz2016}
Tim Caselitz, Bastian Steder, Michael Ruhnke, and Wolfram Burgard.
\newblock {Monocular camera localization in 3D LiDAR maps}.
\newblock {\em IEEE International Conference on Intelligent Robots and
  Systems}, 2016-November:1926--1931, 2016.

\bibitem{Werber2016}
Klaudius Werber, Jens Klappstein, Jurgen Dickmann, and Christian Waldschmidt.
\newblock {Point group associations for radar-based vehicle self-localization}.
\newblock {\em FUSION 2016 - 19th International Conference on Information
  Fusion, Proceedings}, pages 1638--1646, 2016.

\bibitem{Suhr2017}
Jae~Kyu Suhr, Jeungin Jang, Daehong Min, and Ho~Gi Jung.
\newblock {Sensor Fusion-Based Low-Cost Vehicle Localization System for Complex
  Urban Environments}.
\newblock {\em IEEE Transactions on Intelligent Transportation Systems},
  18(5):1078--1086, 2017.

\bibitem{Li2017}
Franck Li, Philippe Bonnifait, Javier Iba{\~{n}}ez-Guzm{\'{a}}n, and Javier
  Ibanez-Guzman.
\newblock {Using High Definition Maps to Estimate GNSS Positioning
  Uncertainty}.
\newblock pages 1--6, 2017.

\bibitem{Li2018}
Franck Li, Philippe Bonnifait, and Javier Ibanez-Guzman.
\newblock {Estimating localization uncertainty using multi-hypothesis
  map-matching on high-definition road maps}.
\newblock {\em IEEE Conference on Intelligent Transportation Systems,
  Proceedings, ITSC}, 2018-March:1--6, 2018.

\bibitem{Ghallabi2018}
Farouk Ghallabi, Fawzi Nashashibi, Ghayath El-Haj-Shhade, and Marie~Anne
  Mittet.
\newblock {LIDAR-Based Lane Marking Detection for Vehicle Positioning in an HD
  Map}.
\newblock {\em IEEE Conference on Intelligent Transportation Systems,
  Proceedings, ITSC}, 2018-November:2209--2214, 2018.

\bibitem{Stenborg2018}
Erik Stenborg, Carl Toft, and Lars Hammarstrand.
\newblock {Long-Term Visual Localization Using Semantically Segmented Images}.
\newblock {\em Proceedings - IEEE International Conference on Robotics and
  Automation}, pages 6484--6490, 2018.

\bibitem{Xiao2018a}
Zhongyang Xiao, Kun Jiang, Shichao Xie, Tuopu Wen, Chunlei Yu, and DIange Yang.
\newblock {Monocular Vehicle Self-localization method based on Compact Semantic
  Map}.
\newblock {\em IEEE Conference on Intelligent Transportation Systems,
  Proceedings, ITSC}, 2018-Novem:3083--3090, 2018.

\bibitem{Ma2019}
Wei~Chiu Ma, Raquel Urtasun, Ignacio Tartavull, Ioan~Andrei Barsan, Shenlong
  Wang, Min Bai, Gellert Mattyus, Namdar Homayounfar, Shrinidhi~Kowshika
  Lakshmikanth, and Andrei Pokrovsky.
\newblock {Exploiting Sparse Semantic HD Maps for Self-Driving Vehicle
  Localization}.
\newblock {\em IEEE International Conference on Intelligent Robots and
  Systems}, pages 5304--5311, 2019.

\bibitem{MinKang2020}
Jeong {Min Kang}, Tae~Sung Yoon, Euntai Kim, and Jin~Bae Park.
\newblock {Lane-level map-matching method for vehicle localization using GPS
  and camera on a high-definition map}.
\newblock {\em Sensors (Switzerland)}, 20(8):1--22, 2020.

\bibitem{Xiao2020}
Zhongyang Xiao, Diange Yang, Tuopu Wen, Kun Jiang, and Ruidong Yan.
\newblock {Monocular localization with vector HD map (MLVHM): A low-cost method
  for commercial IVs}.
\newblock {\em Sensors (Switzerland)}, 20(7):1--24, 2020.

\bibitem{Wang2021}
Huayou Wang, Changliang Xue, Yanxing Zhou, Feng Wen, and Hongbo Zhang.
\newblock {Visual Semantic Localization based on HD Map for Autonomous Vehicles
  in Urban Scenarios}.
\newblock {\em Proceedings - IEEE International Conference on Robotics and
  Automation}, 2021-May(Icra):11255--11261, 2021.

\bibitem{Jiang2021a}
Kun Jiang, Diange Yang, Benny Wijaya, Bowei Zhang, Mengmeng Yang, Kai Zhang,
  and Xuewei Tang.
\newblock {Adding ears to intelligent connected vehicles by combining
  microphone arrays and high definition map}.
\newblock {\em IET Intelligent Transport Systems}, pages 14--27, 2021.

\bibitem{Zhang2022aa}
Zhihuang Zhang, Jintao Zhao, Changyao Huang, and Liang Li.
\newblock {Learning Visual Semantic Map-Matching for Loosely Multi-sensor
  Fusion Localization of Autonomous Vehicles}.
\newblock {\em IEEE Transactions on Intelligent Vehicles}, 8(1):358--367, 2022.

\bibitem{Zhang2022}
Zhihuang Zhang, Meng Xu, Wenqiang Zhou, Tao Peng, Liang Li, and Stefan Poslad.
\newblock {BEV-Locator: An End-to-end Visual Semantic Localization Network
  Using Multi-View Images}.
\newblock 14(8):1--12, 2022.

\bibitem{Kim2023}
Soyeong Kim, Sangkwon Kim, Student Member, Jiwon Seok, Chorong Ryu, Daesung
  Hwang, and Kichun Jo.
\newblock {Road Shape Classification-Based Matching Between Lane Detection and
  HD Map for Robust Localization of Autonomous Cars}.
\newblock {\em IEEE Transactions on Intelligent Vehicles}, 8(5):3431--3443,
  2023.

\bibitem{Bender2014}
Philipp Bender, Julius Ziegler, and Christoph Stiller.
\newblock {Lanelets: Efficient map representation for autonomous driving}.
\newblock {\em IEEE Intelligent Vehicles Symposium, Proceedings},
  (Iv):420--425, 2014.

\bibitem{Feng2020}
Jie Feng, Yong Li, Kai Zhao, Zhao Xu, Tong Xia, Jinglin Zhang, and Depeng Jin.
\newblock {DeepMM: Deep Learning Based Map Matching with Data Augmentation}.
\newblock {\em IEEE Transactions on Mobile Computing}, 1233(c):1--13, 2020.

\bibitem{KichunJo2014}
Kichun Jo and Myoungho Sunwoo.
\newblock {Generation of a Precise Roadway Map for Autonomous Cars}.
\newblock {\em IEEE Transactons on Intelligent Transportation Systems},
  15(3):925--937, 2014.

\bibitem{EbrahimiSoorchaei2022}
Babak {Ebrahimi Soorchaei}, Mahdi Razzaghpour, Rodolfo Valiente, Arash Raftari,
  and Yaser~Pourmohammadi Fallah.
\newblock {High-Definition Map Representation Techniques for Automated
  Vehicles}.
\newblock {\em Electronics (Switzerland)}, 11(20):1--16, 2022.

\bibitem{Greenspan2003}
M.~Greenspan and M.~Yurick.
\newblock {Approximate k-d tree search for efficient ICP}.
\newblock {\em Proceedings of International Conference on 3-D Digital Imaging
  and Modeling, 3DIM}, 2003-January:442--448, 2003.

\bibitem{Xiao2018}
Zhongyang Xiao, Zhaobin Mo, Kun Jiang, and Diange Yang.
\newblock {Multimedia Fusion at Semantic Level in Vehicle Cooperative
  Perception}.
\newblock {\em 2018 IEEE International Conference on Multimedia \& Expo
  Workshops (ICMEW)}, pages 1--6.

\bibitem{Rusu2010}
Radu~Bogdan Rusu.
\newblock {Semantic 3D Object Maps for Everyday Manipulation in Human Living
  Environments}.
\newblock {\em KI - Kunstliche Intelligenz}, 24(4):345--348, 2010.

\bibitem{Poggenhans2018}
Fabian Poggenhans, Jan~Hendrik Pauls, Johannes Janosovits, Stefan Orf,
  Maximilian Naumann, Florian Kuhnt, and Matthias Mayr.
\newblock {Lanelet2: A high-definition map framework for the future of
  automated driving}.
\newblock {\em IEEE Conference on Intelligent Transportation Systems,
  Proceedings, ITSC}, 2018-November:1672--1679, 2018.

\bibitem{Kang2020}
Yunfan Kang and Amr Magdy.
\newblock {HiDaM: A Unified Data Model for High-definition (HD) Map Data}.
\newblock {\em Proceedings - 2020 IEEE 36th International Conference on Data
  Engineering Workshops, ICDEW 2020}, pages 26--32, 2020.

\bibitem{Jiang2023}
Bo~Jiang, Shaoyu Chen, Qing Xu, Bencheng Liao, Jiajie Chen, Helong Zhou, Qian
  Zhang, Wenyu Liu, Chang Huang, and Xinggang Wang.
\newblock {VAD: Vectorized Scene Representation for Efficient Autonomous
  Driving}.
\newblock {\em International Conference on Computer Vision (ICCV)}, 2023.

\bibitem{Guo2023}
Hongyan Guo, Qingyu Meng, Xiaoming Zhao, Jun Liu, Dongpu Cao, and Hong Chen.
\newblock {Map-enhanced generative adversarial trajectory prediction method for
  automated vehicles}.
\newblock {\em Information Sciences}, 622:1033--1049, 2023.

\bibitem{Jeong2022}
Jinseop Jeong, Jun~Yong Yoon, Hwanhong Lee, Hatem Darweesh, and Woosuk Sung.
\newblock {Tutorial on High-Definition Map Generation for Automated Driving in
  Urban Environments}.
\newblock {\em Sensors}, 22(18):1--25, 2022.

\bibitem{longuet-higgins1987computer}
H.~C. Longuet-Higgins.
\newblock {\em A computer algorithm for reconstructing a scene from two
  projections}, pages 61--62.
\newblock MA Fischler and O. Firschein, eds, 1987.

\bibitem{schoenberger2016sfm}
Johannes~Lutz Sch\"{o}nberger and Jan-Michael Frahm.
\newblock Structure-from-motion revisited.
\newblock In {\em Conference on Computer Vision and Pattern Recognition
  (CVPR)}, 2016.

\bibitem{schoenberger2016mvs}
Johannes~Lutz Sch\"{o}nberger, Enliang Zheng, Marc Pollefeys, and Jan-Michael
  Frahm.
\newblock Pixelwise view selection for unstructured multi-view stereo.
\newblock In {\em European Conference on Computer Vision (ECCV)}, 2016.

\bibitem{Dabeer2017}
Onkar Dabeer, Wei Ding, Radhika Gowaiker, Slawomir~K Grzechnik, Mythreya~J
  Lakshman, Sean Lee, Gerhard Reitmayr, Arunandan Sharma, Kiran Somasundaram,
  Ravi~Teja Sukhavasi, and Xinzhou Wu.
\newblock {An End-to-End System for Crowdsourced 3D Maps for Autonomous
  Vehicles : The Mapping Component}.
\newblock {\em IEEE/RSJ International Conference on Intelligent Robots and
  Systems (IROS)}, 2017.

\bibitem{Golovnin2020}
O.~K. Golovnin and D.~V. Rybnikov.
\newblock {Video Processing Method for High-Definition Maps Generation}.
\newblock {\em 2020 International Multi-Conference on Industrial Engineering
  and Modern Technologies, FarEastCon 2020}, pages 0--4, 2020.

\bibitem{Paz2020}
David Paz, Hengyuan Zhang, Qinru Li, Hao Xiang, and Henrik~I. Christensen.
\newblock {Probabilistic semantic mapping for urban autonomous driving
  applications}.
\newblock {\em IEEE International Conference on Intelligent Robots and
  Systems}, pages 2059--2064, 2020.

\bibitem{Chawla2020}
H~Chawla, M~Jukola, T~Brouns, E~Arani, B~Zonooz, Self-supervised~Learning
  Approach, and Las Vegas.
\newblock {Crowdsourced 3D Mapping: A Combined Multi-View Geometry and
  Self-Supervised Learning Approach}.
\newblock 2020.

\bibitem{Das2020}
Anweshan Das, Joris Ijsselmuiden, and Gijs Dubbelman.
\newblock {Pose-graph based crowdsourced mapping framework}.
\newblock {\em 2020 IEEE 3rd Connected and Automated Vehicles Symposium, CAVS
  2020 - Proceedings}, 2020.

\bibitem{Wen2022}
Tuopu Wen, Kun Jiang, Jinyu Miao, Benny Wijaya, Peijin Jia, Mengmeng Yang, and
  Diange Yang.
\newblock {Roadside HD Map Object Reconstruction Using Monocular Camera}.
\newblock {\em IEEE Robotics and Automation Letters}, 7(3):7722--7729, 2022.

\bibitem{Zhang2023a}
Songyi Zhang, Runsheng Wang, Zhiqiang Jian, Wei Zhan, Nanning Zheng, and
  Masayoshi Tomizuka.
\newblock {Clothoid-Based Reference Path Reconstruction for HD Map Generation}.
\newblock {\em IEEE Transactions on Intelligent Transportation Systems},
  25(1):587--601, 2023.

\bibitem{Zhang2023}
Zhixin Zhang, Yiyuan Zhang, Xiaohan Ding, Fusheng Jin, and Xiangyu Yue.
\newblock {Online Vectorized HD Map Construction using Geometry}.
\newblock 2023.

\bibitem{Mooney2014}
John~G. Mooney and Eric~N. Johnson.
\newblock {A Comparison of Automatic Nap-of-the-earth Guidance Strategies for
  Helicopters}.
\newblock {\em Journal of Field Robotics}, 33(1):1--17, 2014.

\bibitem{Thrun2006}
Sebastian Thrun and Michael Montemerlo.
\newblock {The graph SLAM algorithm with applications to large-scale mapping of
  urban structures}.
\newblock {\em International Journal of Robotics Research}, 25(5-6):403--429,
  2006.

\bibitem{Grisetti2010}
Giorgio Grisetti, Rainer Kummerle, Cyrill Stachniss, and Wolfram Burgard.
\newblock {A tutorial on graph-based SLAM}.
\newblock {\em IEEE Intelligent Transportation Systems Magazine}, 2(4):31--43,
  2010.

\bibitem{Soheilian2010}
Bahman Soheilian, Nicolas Paparoditis, and Didier Boldo.
\newblock {3D road marking reconstruction from street-level calibrated stereo
  pairs}.
\newblock {\em ISPRS Journal of Photogrammetry and Remote Sensing},
  65(4):347--359, 2010.

\bibitem{Ilci2020}
Veli Ilci and Charles Toth.
\newblock {High definition 3D map creation using GNSS/IMU/LiDAR sensor
  integration to support autonomous vehicle navigation}.
\newblock {\em Sensors (Switzerland)}, 20(3), 2020.

\bibitem{Gao2020a}
Jiyang Gao, Chen Sun, Hang Zhao, Yi~Shen, Dragomir Anguelov, Congcong Li, and
  Cordelia Schmid.
\newblock {VectorNet: Encoding HD maps and agent dynamics from vectorized
  representation}.
\newblock {\em Proceedings of the IEEE Computer Society Conference on Computer
  Vision and Pattern Recognition}, pages 11522--11530, 2020.

\bibitem{Mi2021}
Lu~Mi, Hang Zhao, Charlie Nash, Xiaohan Jin, Jiyang Gao, Chen Sun, Cordelia
  Schmid, Nir Shavit, Yuning Chai, and Dragomir Anguelov.
\newblock {HDMapGen: A Hierarchical Graph Generative Model of High Definition
  Maps}.
\newblock {\em Proceedings of the IEEE Computer Society Conference on Computer
  Vision and Pattern Recognition}, pages 4225--4234, 2021.

\bibitem{Fenwick2002}
John~W. Fenwick, Paul~M. Newman, and John~J. Leonard.
\newblock {Cooperative concurrent mapping and localization}.
\newblock {\em Proceedings - IEEE International Conference on Robotics and
  Automation}, 2(May):1810--1817, 2002.

\bibitem{Gil2010}
Arturo Gil, {\'{O}}scar Reinoso, M{\'{o}}nica Ballesta, and Miguel Juli{\'{a}}.
\newblock {Multi-robot visual SLAM using a Rao-Blackwellized particle filter}.
\newblock {\em Robotics and Autonomous Systems}, 58(1):68--80, 2010.

\bibitem{Riazuelo2014}
L.~Riazuelo, Javier Civera, and J.~M.M. Montiel.
\newblock {C2TAM: A Cloud framework for cooperative tracking and mapping}.
\newblock {\em Robotics and Autonomous Systems}, 62(4):401--413, 2014.

\bibitem{Arumugam2010}
Rajesh Arumugam, Vikas~Reddy Enti, Liu Bingbing, Wu~Xiaojun, Krishnamoorthy
  Baskaran, Foong~Foo Kong, A.~Senthil Kumar, Kang~Dee Meng, and Goh~Wai Kit.
\newblock {DAvinCi: A cloud computing framework for service robots}.
\newblock {\em Proceedings - IEEE International Conference on Robotics and
  Automation}, pages 3084--3089, 2010.

\bibitem{Montemerlo-2003-8695}
Michael Montemerlo.
\newblock {\em FastSLAM: A Factored Solution to the Simultaneous Localization
  and Mapping Problem with Unknown Data Association}.
\newblock PhD thesis, Carnegie Mellon University, Pittsburgh, PA, July 2003.

\bibitem{Mohanarajah2015}
Gajamohan Mohanarajah, Dominique Hunziker, Raffaello D'Andrea, and Markus
  Waibel.
\newblock {Rapyuta: A Cloud Robotics Platform}.
\newblock {\em IEEE Transactions on Automation Science and Engineering},
  12(2):481--493, 2015.

\bibitem{Kehoe2015}
Ben Kehoe, Sachin Patil, Pieter Abbeel, and Ken Goldberg.
\newblock {A Survey of Research on Cloud Robotics and Automation}.
\newblock {\em IEEE Transactions on Automation Science and Engineering},
  12(2):398--409, 2015.

\bibitem{Cadena2016}
Cesar Cadena, Luca Carlone, Henry Carrillo, Yasir Latif, Davide Scaramuzza,
  Jose Neira, Ian Reid, and John~J. Leonard.
\newblock {Past, present, and future of simultaneous localization and mapping:
  Toward the robust-perception age}.
\newblock {\em IEEE Transactions on Robotics}, 32(6):1309--1332, 2016.

\bibitem{Philion2020}
Jonah Philion and Sanja Fidler.
\newblock {Lift, Splat, Shoot: Encoding Images from Arbitrary Camera Rigs by
  Implicitly Unprojecting to 3D}.
\newblock {\em The European Conference on Computer Vision (ECCV)}, pages
  194--210, 2020.

\bibitem{Dong2022}
Hao Dong, Xianjing Zhang, Jintao Xu, Rui Ai, Weihao Gu, Huimin Lu, Juho
  Kannala, and Xieyuanli Chen.
\newblock {SuperFusion: Multilevel LiDAR-Camera Fusion for Long-Range HD Map
  Generation}.
\newblock 2022.

\bibitem{carion2020end}
Nicolas Carion, Francisco Massa, Gabriel Synnaeve, Nicolas Usunier, Alexander
  Kirillov, and Sergey Zagoruyko.
\newblock End-to-end object detection with transformers.
\newblock In {\em European Conference on Computer Vision}, pages 213--229.
  Springer, 2020.

\bibitem{Liao2022}
Bencheng Liao, Shaoyu Chen, Xinggang Wang, Tianheng Cheng, Qian Zhang, Wenyu
  Liu, and Chang Huang.
\newblock {MapTR: Structured Modeling and Learning for Online Vectorized HD Map
  Construction}.
\newblock {\em International Conference on Learning Representations (ICLR)},
  pages 1--18, 2022.

\bibitem{Liao2023}
Bencheng Liao, Shaoyu Chen, Yunchi Zhang, Bo~Jiang, Qian Zhang, Wenyu Liu,
  Chang Huang, and Xinggang Wang.
\newblock {MapTRv2: An End-to-End Framework for Online Vectorized HD Map
  Construction}.
\newblock {\em International Conference on Learning Representations (ICLR)},
  pages 1--17, 2023.

\bibitem{Liu2023a}
Yicheng Liu, Tianyuan Yuan, Yue Wang, Yilun Wang, and Hang Zhao.
\newblock {VectorMapNet: End-to-end Vectorized HD Map Learning}.
\newblock {\em Proceedings of Machine Learning Research}, 202:21839--21866,
  2023.

\bibitem{Qiao2023a}
Limeng Qiao, Wenjie Ding, Xi~Qiu, and Chi Zhang.
\newblock {End-to-End Vectorized HD-map Construction with Piecewise
  B{\'{e}}zier Curve}.
\newblock {\em Proceedings of the IEEE Computer Society Conference on Computer
  Vision and Pattern Recognition}, 2023-June:13218--13228, 2023.

\bibitem{Shin2023}
Juyeb Shin, Francois Rameau, Hyeonjun Jeong, and Dongsuk Kum.
\newblock {InstaGraM: Instance-level Graph Modeling for Vectorized HD Map
  Learning}.
\newblock 2023.

\bibitem{Ding2023}
Wenjie Ding, Limeng Qiao, Xi~Qiu, and Chi Zhang.
\newblock {PivotNet: Vectorized Pivot Learning for End-to-end HD Map
  Construction}.
\newblock {\em IEEE / CVF Computer Vision and Pattern Recognition Conference
  (CVPR)}, pages 3672--3682, 2023.

\bibitem{Yu2023}
Jingyi Yu, Zizhao Zhang, Shengfu Xia, and Jizhang Sang.
\newblock {ScalableMap: Scalable Map Learning for Online Long-Range Vectorized
  HD Map Construction}.
\newblock {\em The Conference on Robot Learning (CoRL)}, (CoRL), 2023.

\bibitem{Yuan2023}
Tianyuan Yuan, Yicheng Liu, Yue Wang, Yilun Wang, and Hang Zhao.
\newblock {StreamMapNet: Streaming Mapping Network for Vectorized Online HD Map
  Construction}.
\newblock {\em IEEE / CVF Computer Vision and Pattern Recognition Conference
  (CVPR)}, 2023.

\bibitem{Qiao2023}
Limeng Qiao, Yongchao Zheng, Peng Zhang, Wenjie Ding, Xi~Qiu, Xing Wei, and Chi
  Zhang.
\newblock {MachMap: End-to-End Vectorized Solution for Compact HD-Map
  Construction}.
\newblock (2):1--5, 2023.

\bibitem{TsinghuaMARS2023OnlineHDMap}
{Tsinghua MARS Lab}.
\newblock Online hd map construction for cvpr 2023 challenge, 2023.
\newblock Accessed: 2/1/2024.

\bibitem{li2024}
Toyota Li.
\newblock {MapNeXt: Revisiting Training and Scaling Practices for Online
  Vectorized HD Map Construction}.
\newblock 2024.

\bibitem{Hu2024}
Haotian Hu, Fanyi Wang, Yaonong Wang, Laifeng Hu, Jingwei Xu, and Zhiwang
  Zhang.
\newblock {ADMap: Anti-disturbance framework for reconstructing online
  vectorized HD map}.
\newblock pages 1--9, 2024.

\bibitem{Tao2010}
Tong Tao, Yalou Huang, Jing Yuan, Fengchi Sun, and Xiaolin Wu.
\newblock {Cooperative simultaneous localization and mapping for multi-robot:
  Approach \& experimental validation}.
\newblock {\em Proceedings of the World Congress on Intelligent Control and
  Automation (WCICA)}, pages 2888--2893, 2010.

\bibitem{Stoven-Dubois2019}
Alexis Stoven-Dubois, Kuntima~Kiala Miguel, Aziz Dziri, Bertrand Leroy, and
  Roland Chapuis.
\newblock {A Collaborative Framework for High-Definition Mapping}.
\newblock {\em IEEE Intelligent Transportation Systems Conference, ITSC 2019},
  pages 1845--1850, 2019.

\bibitem{Stoven-Dubois2020}
Alexis Stoven-Dubois, Aziz Dziri, Bertrand Leroy, and Roland Chapuis.
\newblock {Graph Optimization Methods for Large-Scale Crowdsourced Mapping}.
\newblock pages 1--8, 2020.

\bibitem{Song2023}
Haina Song, Binjie Hu, Qinyan Huang, Yi~Zhang, and Jiwei Song.
\newblock {A Lightweight High Definition Mapping Method Based on Multi-Source
  Data Fusion Perception}.
\newblock {\em Applied Sciences (Switzerland)}, 13(5), 2023.

\bibitem{Quddus2006}
Mohammed~A. Quddus, Washington~Y. Ochieng, and Robert~B. Noland.
\newblock {Integrity of map-matching algorithms}.
\newblock {\em Transportation Research Part C: Emerging Technologies},
  14(4):283--302, 2006.

\bibitem{Kubicka2015}
Matej Kubicka, Arben Cela, Philippe Moulin, Hugues Mounier, and S.~I.
  Niculescu.
\newblock {Dataset for testing and training of map-matching algorithms}.
\newblock {\em IEEE Intelligent Vehicles Symposium, Proceedings},
  2015-August(Iv):1088--1093, 2015.

\bibitem{Zhang2021}
Pan Zhang, Mingming Zhang, and Jingnan Liu.
\newblock {Real-time hd map change detection for crowdsourcing update based on
  mid-to-high-end sensors}.
\newblock {\em Sensors (Switzerland)}, 21(7):1--12, 2021.

\bibitem{Jo2018}
Kichun Jo, Chansoo Kim, and Myoungho Sunwoo.
\newblock {Simultaneous localization and map change update for the high
  definition map-based autonomous driving car}.
\newblock {\em Sensors (Switzerland)}, 18(9), 2018.

\bibitem{Lambert2021}
John Lambert and James Hays.
\newblock {Trust, but Verify: Cross-Modality Fusion for HD Map Change
  Detection}.
\newblock (NeurIPS):1--14, 2021.

\bibitem{Cong2023}
Yangzi Cong, Chi Chen, Bisheng Yang, Fuxun Liang, Ruiqi Ma, and Fei Zhang.
\newblock {CAOM: Change-aware online 3D mapping with heterogeneous multi-beam
  and push-broom LiDAR point clouds}.
\newblock {\em ISPRS Journal of Photogrammetry and Remote Sensing},
  195(November 2022):204--219, 2023.

\bibitem{Wijaya2023}
Benny Wijaya, Mengmeng Yang, Tuopu Wen, Kun Jiang, Yunlong Wang, Zheng Fu,
  Xuewei Tang, Dennis~Octovan Sigomo, Jinyu Miao, and Diange Yang.
\newblock {Multi-Session High-Definition Map-Monitoring System for Map Update}.
\newblock {\em ISPRS International Journal of Geo-Information}, 13(1):6, 2023.

\bibitem{TomTomInternationalBV2019}
{TomTom International BV}.
\newblock {HD map with RoadDNA}.
\newblock 2019.

\bibitem{Cho2023}
Minwoo Cho, Kitae Kim, Soohyun Cho, Seung-mo Cho, and Woojin Chung.
\newblock {Frequent and Automatic Update of Lane-Level HD Maps with a Large
  Amount of Crowdsourced Data Acquired from Buses and Taxis in Seoul}.
\newblock 2023.

\bibitem{Wijaya2022a}
Benny Wijaya, Kun Jiang, Mengmeng Yang, Tuopu Wen, Xuewei Tang, Diange Yang,
  Yanhai Ma, and Ricky Albert.
\newblock {CrowdRep: A Blockchain-based Reputation System for Crowdsourced HD
  Map Update}.
\newblock {\em IEEE Conference on Intelligent Transportation Systems,
  Proceedings, ITSC}, 2022-Octob:3050--3057, 2022.

\bibitem{Xiong2023}
Xuan Xiong, Yicheng Liu, Tianyuan Yuan, Yue Wang, Yilun Wang, and Hang Zhao.
\newblock {Neural Map Prior for Autonomous Driving}.
\newblock {\em Proceedings of the IEEE Computer Society Conference on Computer
  Vision and Pattern Recognition}, 2023-June:17535--17544, 2023.

\bibitem{Sayed2023}
Mohamed Sayed, Stepan Perminov, and Dzmitry Tsetserukou.
\newblock {PolyMerge: A Novel Technique aimed at Dynamic HD Map Updates
  Leveraging Polylines}.
\newblock {\em 21st International Conference on Advanced Robotics (ICAR)},
  2023.

\bibitem{Chen2023}
Dawei Chen, Yifei Zhu, Dan Wang, and Senior Member.
\newblock {Love of Variety Based Latency Analysis for High Definition Map
  Updating : Age of Information and Distributional Robust Perspectives}.
\newblock {\em IEEE Transactions on Intelligent Vehicles}, 8(2):1751--1764,
  2023.

\bibitem{M.2010}
Everingham M., Van-Gool L., Williams C~K I., Winn J., and Zisserman A.
\newblock {The Pascal Visual Object Classes (VOC) Challenge}.
\newblock {\em International Journal of Computer Vision}, 88(2):303--338, 2010.

\bibitem{Jeong2018}
Jinyong Jeong, Younggun Cho, Young~Sik Shin, Hyunchul Roh, and Ayoung Kim.
\newblock {Complex Urban LiDAR Data Set}.
\newblock {\em 2018 IEEE International Conference on Robotics and Automation,
  Brisbane, 21-26 May 2018.}

\bibitem{Chang_2019_CVPR}
Ming-Fang Chang, John Lambert, Patsorn Sangkloy, Jagjeet Singh, Slawomir Bak,
  Andrew Hartnett, De~Wang, Peter Carr, Simon Lucey, Deva Ramanan, and James
  Hays.
\newblock Argoverse: 3d tracking and forecasting with rich maps.
\newblock In {\em Proceedings of the IEEE/CVF Conference on Computer Vision and
  Pattern Recognition (CVPR)}, June 2019.

\bibitem{Argoverse2}
Benjamin Wilson, William Qi, Tanmay Agarwal, John Lambert, Jagjeet Singh,
  Siddhesh Khandelwal, Bowen Pan, Ratnesh Kumar, Andrew Hartnett, Jhony
  Kaesemodel~Pontes, Deva Ramanan, Peter Carr, and James Hays.
\newblock Argoverse 2: Next generation datasets for self-driving perception and
  forecasting.
\newblock In J.~Vanschoren and S.~Yeung, editors, {\em Proceedings of the
  Neural Information Processing Systems Track on Datasets and Benchmarks},
  volume~1. Curran, 2021.

\bibitem{Caesar2020}
Holger Caesar, Varun Bankiti, Alex~H. Lang, Sourabh Vora, Venice~Erin Liong,
  Qiang Xu, Anush Krishnan, Yu~Pan, Giancarlo Baldan, and Oscar Beijbom.
\newblock {Nuscenes: A multimodal dataset for autonomous driving}.
\newblock {\em Proceedings of the IEEE Computer Society Conference on Computer
  Vision and Pattern Recognition}, (March):11618--11628, 2020.

\bibitem{DA4AD_2020_ECCV}
Yao Zhou, Guowei Wan, Shenhua Hou, Li~Yu, Gang Wang, Xiaofei Rui, and Shiyu
  Song.
\newblock Da4ad: End-to-end deep attention-based visual localization for
  autonomous driving.
\newblock In {\em Proceedings of the European Conference on Computer Vision
  (ECCV)}, 2020.

\bibitem{Xue2022aa}
Yongjie Xue, Yuru Zhang, and Qiang Liu.
\newblock {CoMap : Proactive Provision for Crowdsourcing Map in Automotive Edge
  Computing}.

\bibitem{WangT2016}
Gang Wang, Bolun Wang, Tianyi Wang, Ana Nika, Haitao Zheng, and Ben~Y. Zhao.
\newblock {Attacks and Defenses in Crowdsourced Mapping Services}.
\newblock {\em MobiSys 2016 Companion - Companion Publication of the 14th
  Annual International Conference on Mobile Systems, Applications, and
  Services}, page 146, 2016.

\bibitem{Wang2018}
Gang Wang, Bolun Wang, Tianyi Wang, Ana Nika, Haitao Zheng, Ben~Y Zhao, and
  Senior Member.
\newblock {Ghost Riders : Sybil Attacks on Crowdsourced Mobile Mapping
  Services}.
\newblock 26(3):1123--1136, 2018.

\end{thebibliography}

\begin{IEEEbiography}
    [{\includegraphics[width=1in,height=1.25in,clip,keepaspectratio]{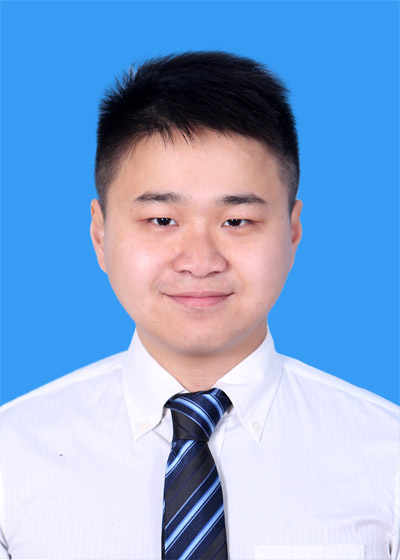}}]{Benny Wijaya}
received the B.S. degree in mechanical engineering from University of Newcastle, Newcastle, Australia in 2015. Then he received the Master degree in mechanical engineering from Tsinghua University, Beijing, China in 2020. He is currently working toward the Ph.D. degree at the School of Vehicle and Mobility in Tsinghua University, Beijing, China.
His research interests include, sensor fusion, confidence modelling, and high definition mapping and update for autonomous driving.
\end{IEEEbiography}

\vskip 0pt plus -1fil
\begin{IEEEbiography}
    [{\includegraphics[width=1in,height=1.25in,clip,keepaspectratio]{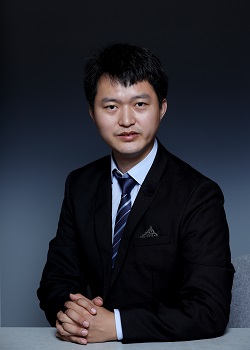}}]{Kun Jiang}
received the B.S. degree in mechanical and automation engineering from Shanghai Jiao Tong University, Shanghai, China in 2011. Then he received the Master degree in mechatronics system and the Ph.D. degree in information and systems technologies from University of Technology of Compi\`egne (UTC), Compi\`egne, France, in 2013 and 2016, respectively. He is currently an assistant research professor at Tsinghua University, Beijing, China. His research interests include autonomous vehicles, high precision digital map, and sensor fusion.
\end{IEEEbiography}
\vskip 0pt plus -1fil

\begin{IEEEbiography}
    [{\includegraphics[width=1in,height=1.25in,clip,keepaspectratio]{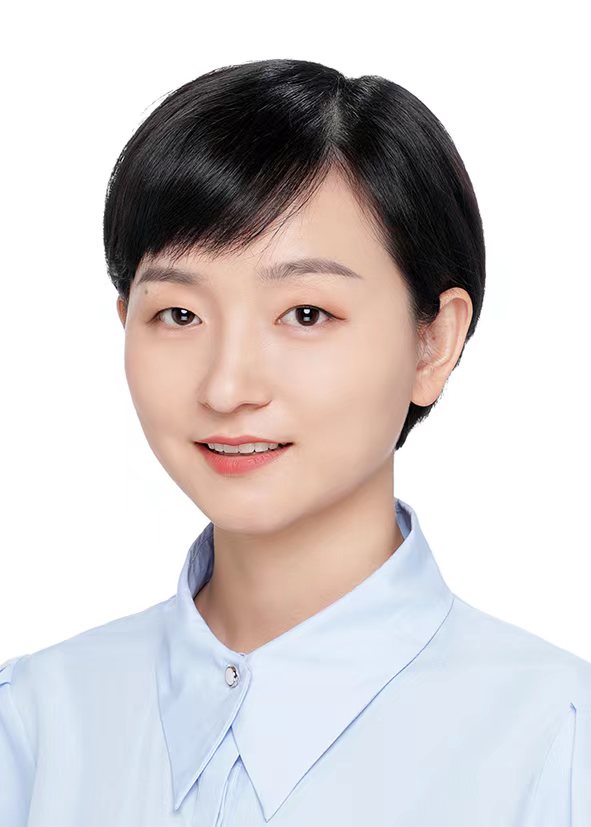}}]{Mengmeng Yang}
received the PhD degree in Photogrammetry and Remote Sensing from Wuhan University, Wuhan, China in 2018. She is currently an assistant research professor at Tsinghua University, Beijing, China. Her research interests include autonomous vehicles, high precision digital map, and sensor fusion. 
\end{IEEEbiography}

\vskip 0pt plus -1fil
\begin{IEEEbiography}
    [{\includegraphics[width=1in,height=1.25in,clip,keepaspectratio]{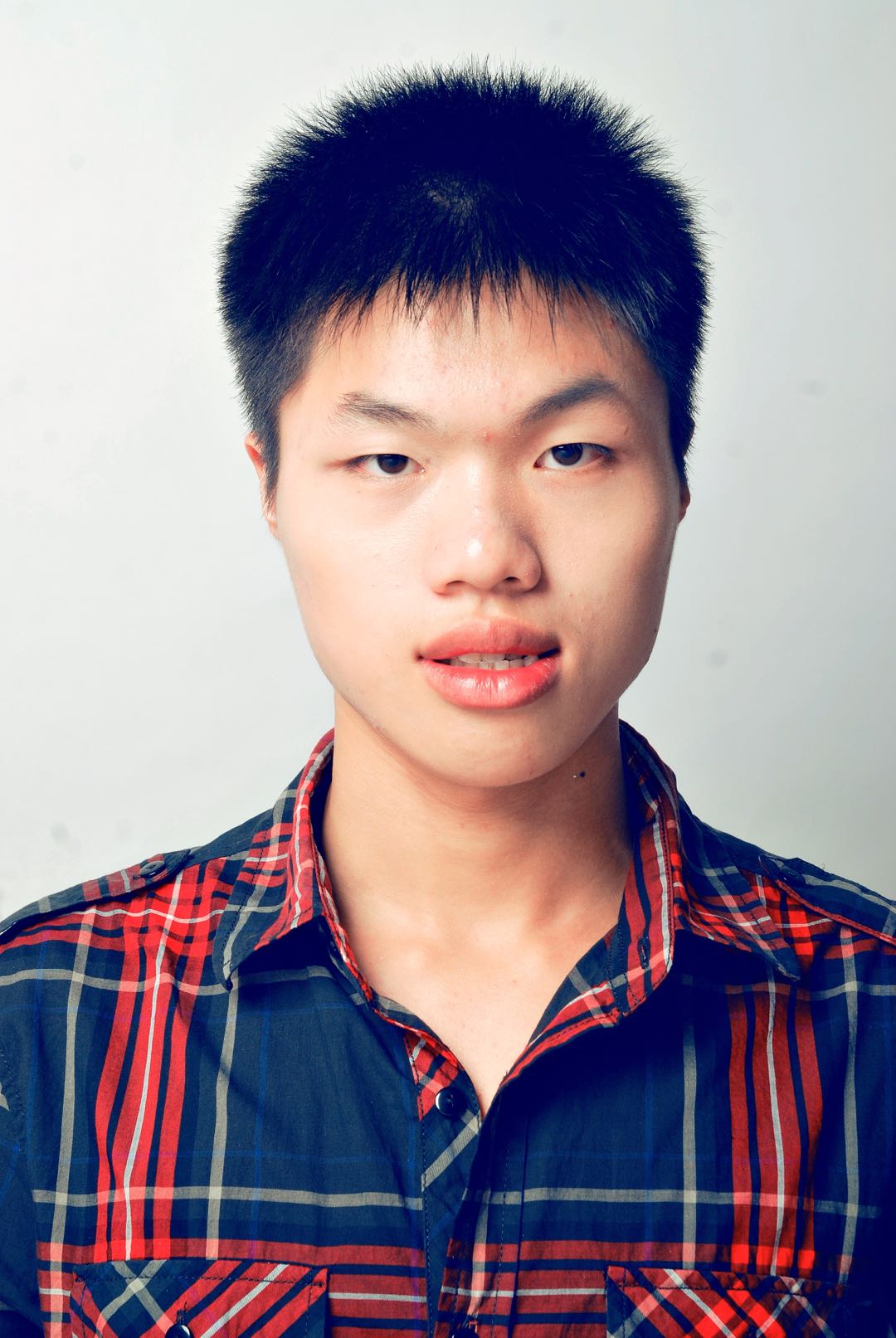}}]{Tuopu Wen}
received the B.S. degree from Electronic Engineering, Tsinghua University, Beijing, China in 2018. He is currently working toward the Ph.D. degree at the School of Vehicle and Mobility of Tsinghua University, Beijing, China.
His research interests include computer vision, high definition map, and high precision localization for autonomous driving.
\end{IEEEbiography}
\vskip 0pt plus -1fil

\begin{IEEEbiography}
    [{\includegraphics[width=1in,height=1.25in,clip,keepaspectratio]{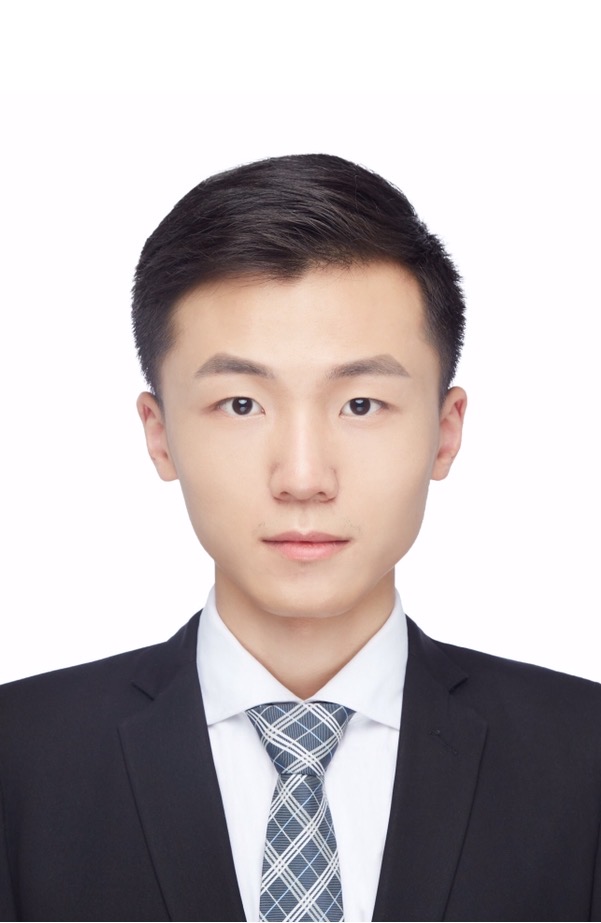}}]{Yunlong Wang}
received the B.S. degree from Electronic Engineering, Tsinghua University, Beijing, China in 2019. He is currently working toward the Ph.D. degree at the School of Vehicle and Mobility of Tsinghua University, Bejing, China. His research interests include LiDAR point cloud semantic segmentation, motion prediction and 3D object detection for autonomous driving.
\end{IEEEbiography}

\vskip 0pt plus -1fil

\begin{IEEEbiography}
    [{\includegraphics[width=1in,height=1.25in,clip,keepaspectratio]{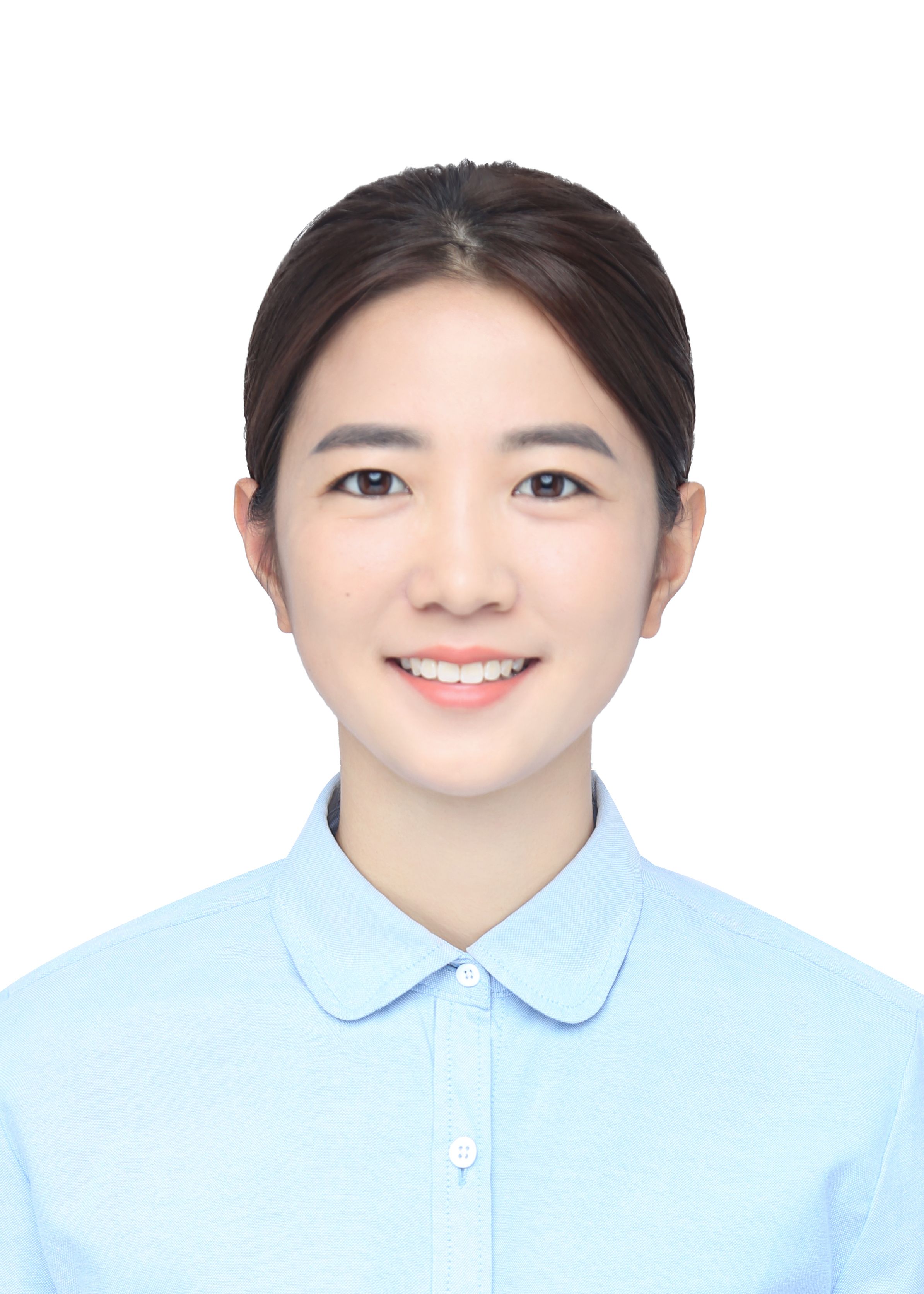}}]{Zheng Fu}
received the M.S. degree in Pattern Recognition and Intelligent Systems from Nanjing University of Posts and Telecommunications, Jiangsu, China, in 2019. She is pursuing a Ph.D. degree at the School of Vehicle and Mobility, Tsinghua University, Beijing, China. Her current research interests include autonomous driving algorithms and human-vehicle interaction in autonomous driving.
\end{IEEEbiography}
\vskip 0pt plus -1fil

\begin{IEEEbiography}
    [{\includegraphics[width=1in,height=1.25in,clip,keepaspectratio]{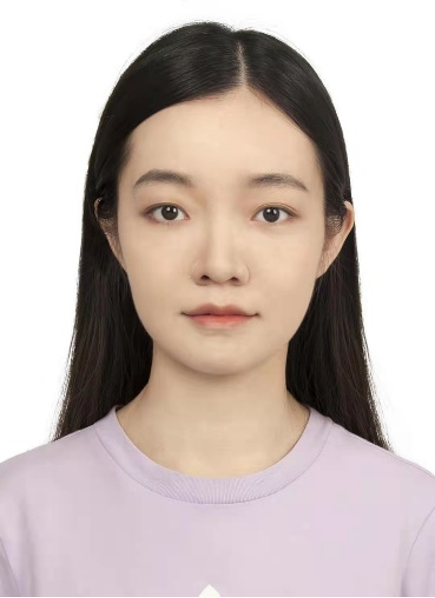}}]{Xuewei Tang}
received the B.S. degree from the Department of Automation, Tsinghua University, Beijing, China, in 2020. She is currently working toward the Ph.D. degree at the School of Vehicle and Mobility of Tsinghua University, Beijing, China. Her research interests include computer vision and environment modeling.
\end{IEEEbiography}

\vskip 0pt plus -1fil

\begin{IEEEbiography}
    [{\includegraphics[width=1in,height=1.25in,clip,keepaspectratio]{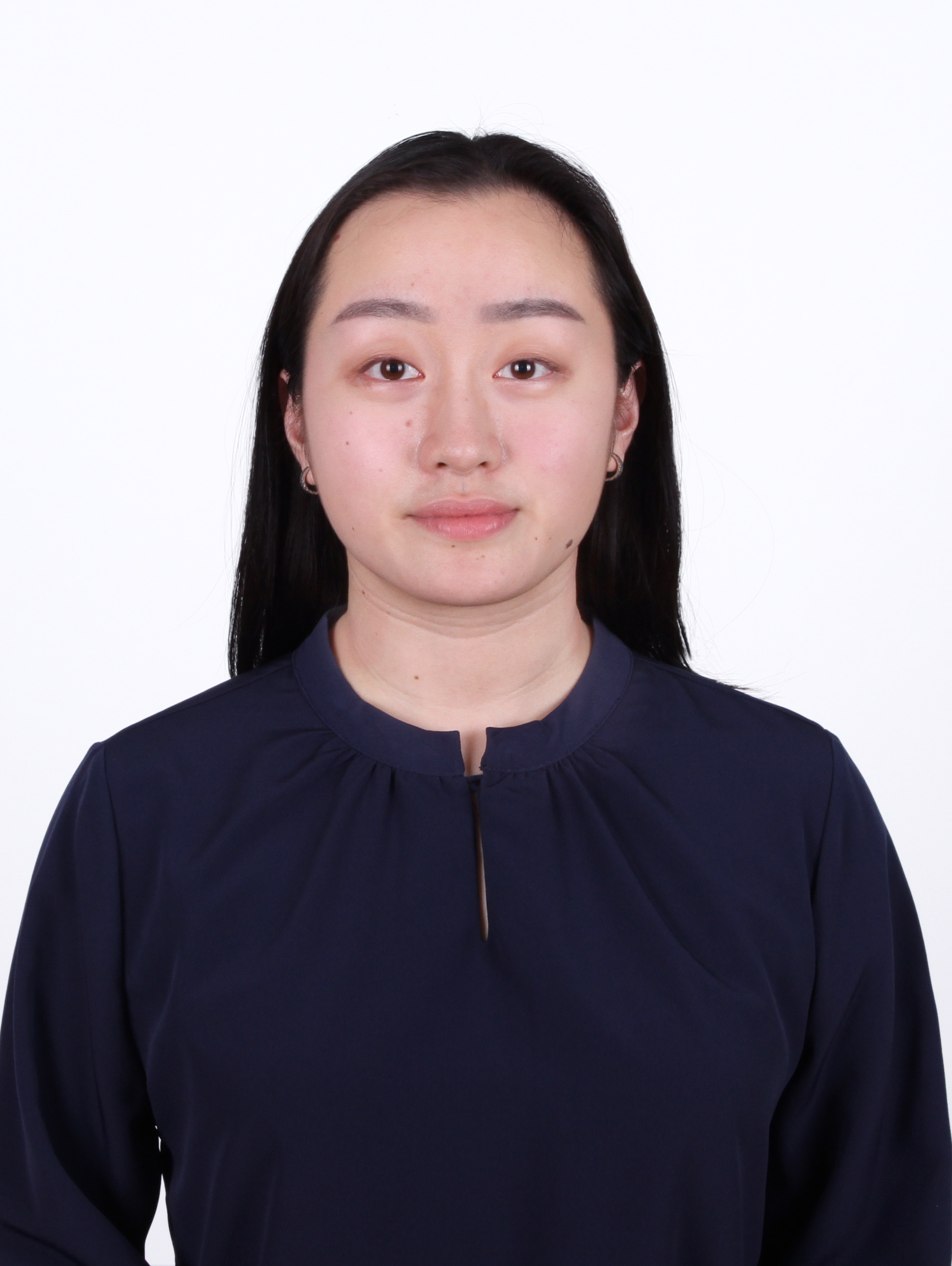}}]{Gracelynn Soesanto}
received the B.S.E. degree in Chemical and Biomolecular Engineering from the University of Pennsylvania, Philadelphia, PA, USA in 2021. She is currently a Master’s degree student in the Industrial Engineering department at Tsinghua University.  
\end{IEEEbiography}
\vskip 0pt plus -1fil

\begin{IEEEbiography}
    [{\includegraphics[width=1in,height=1.25in,clip,keepaspectratio]{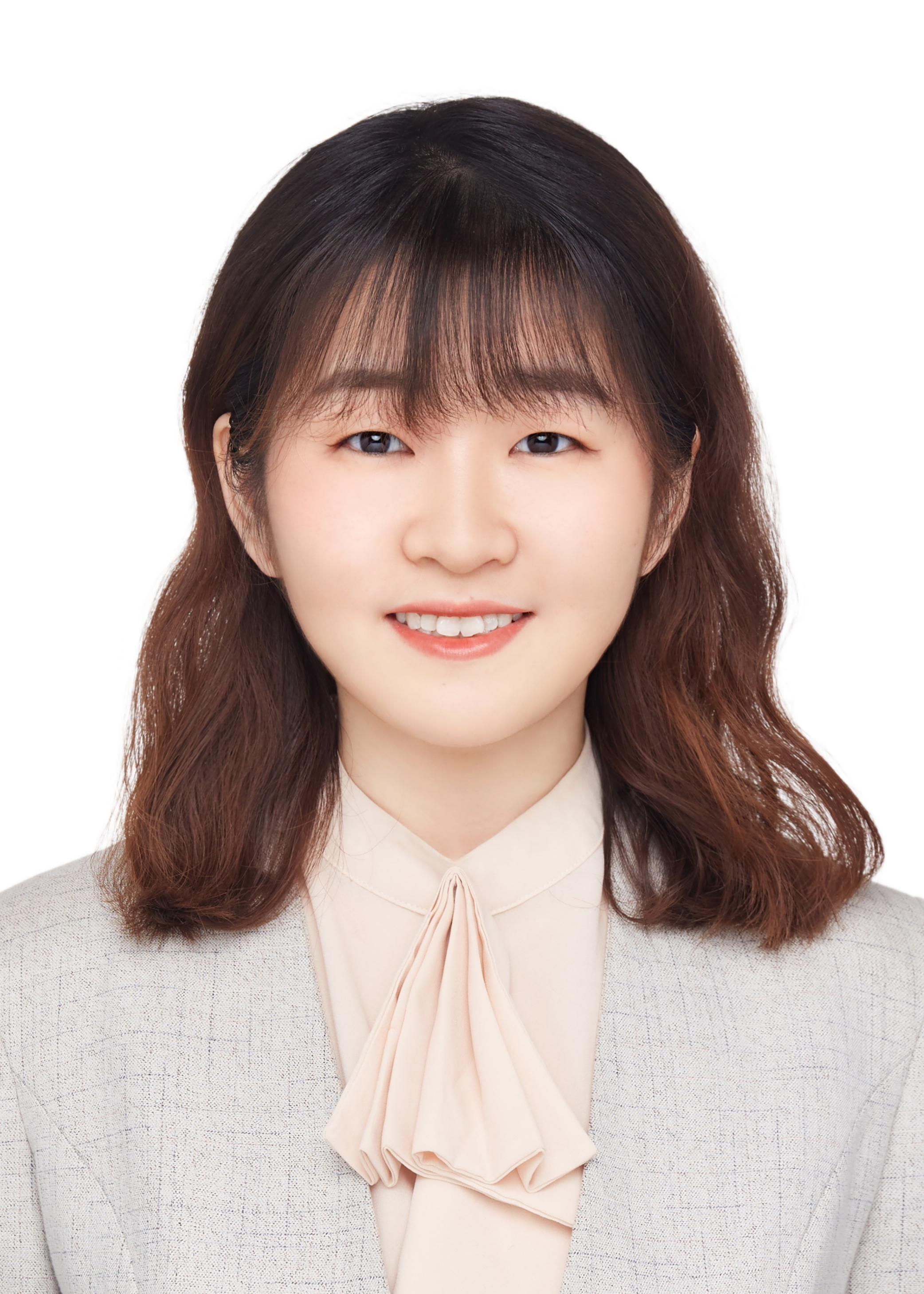}}]{Taohua Zhou}
received the B.S. degree from Department of Automation, Tsinghua University, Beijing, China in 2018. She is currently working toward the Doctor degree at the School of Vehicle and Mobility, Tsinghua University, Beijing, China. Her research interests  include computer vision, information fusion, and environmental perception of autonomous driving.
\end{IEEEbiography}
\vskip 0pt plus -1fil

\begin{IEEEbiography}
    [{\includegraphics[width=1in,height=1.25in,clip,keepaspectratio]{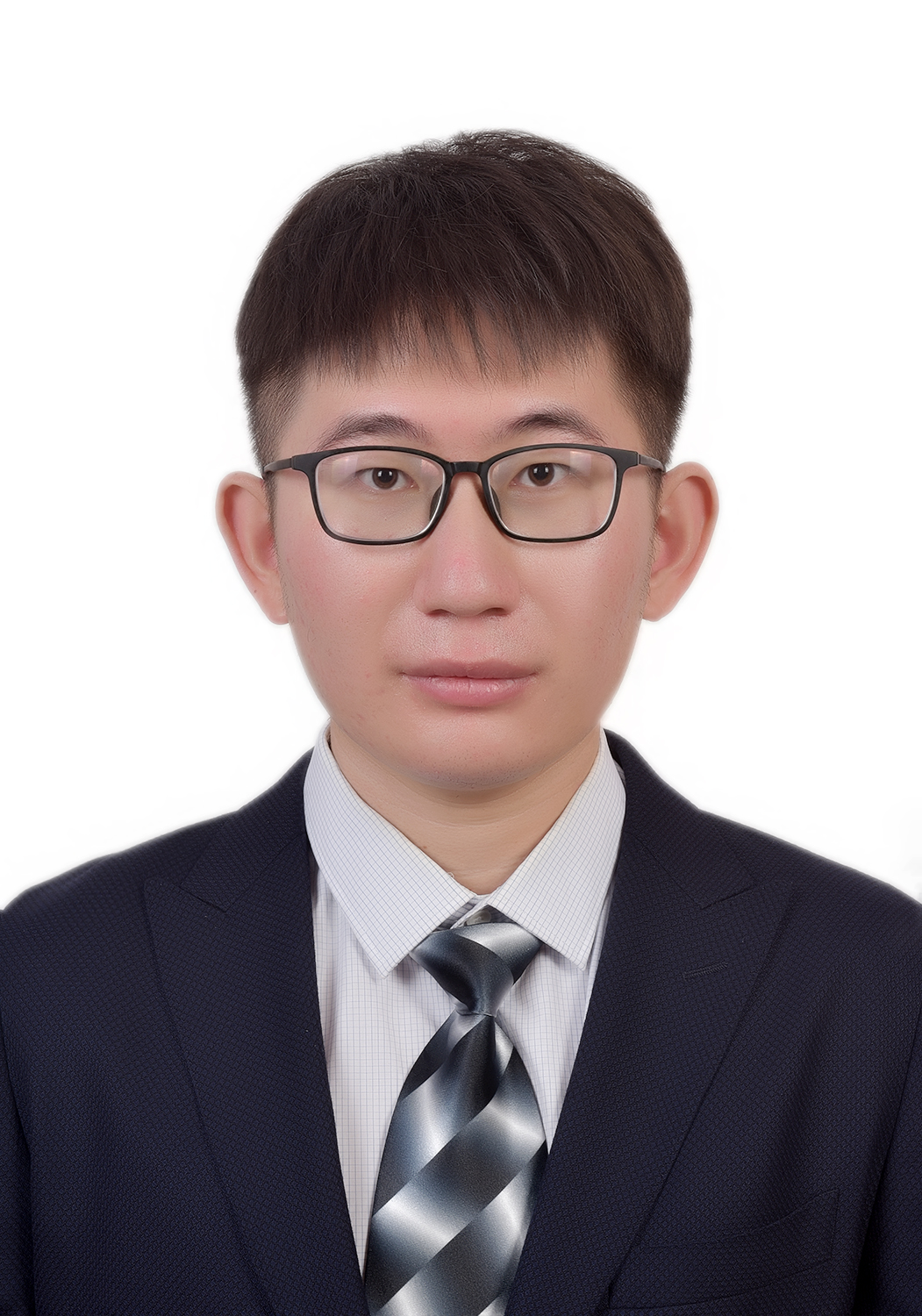}}]{Jinyu Miao}
received his B.S. degree in Automation and his M.S. degree in Control Science and Engineering from the School of Automation Science and Electrical Engineering, Beihang University, Beijing, China, in 2019 and 2022, respectively. He is now currently working toward his Ph.D. degree at the School of Vehicle and Mobility, Tsinghua University, Beijing, China. His research interests include visual SLAM, high definition map, and high-precision localization for autonomous driving.
\end{IEEEbiography}
\vskip 0pt plus -1fil

\begin{IEEEbiography}
    [{\includegraphics[width=1in,height=1.25in,clip,keepaspectratio]{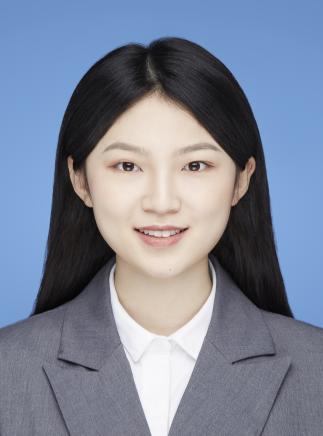}}]{Peijin Jia}
received the B.S. degree from the Department of Automation, Beihang University, Beijing, China, in 2022. She is currently working toward the Master degree at the School of Vehicle and Mobility of Tsinghua University, Beijing, China. Her research interests include computer vision and online HD Mapping.
\end{IEEEbiography}

\vskip 0pt plus -1fil

\begin{IEEEbiography}
    [{\includegraphics[width=1in,height=1.25in,clip,keepaspectratio]{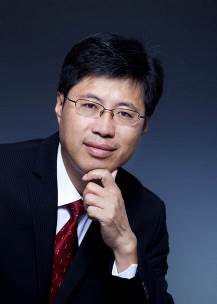}}]{Diange Yang}
received his Ph.D. in Automotive Engineering from Tsinghua University in 2001. He is now a Professor at the school of vehicle and mobility at Tsinghua University. He currently serves as the director of the Tsinghua University Development \& Planning Division. His research interests include autonomous driving, environment perception, and HD map.
He has more than 180 technical publications and 100 patents. He received the National Technology Invention Award and Science and Technology Progress Award in 2010, 2013, 2018, and the Special Prize for Progress in Science and Technology of China Automobile Industry in 2019. 

\end{IEEEbiography}

\begin{appendices}

\begin{sidewaystable*} 
\centering
\resizebox{1.0\linewidth}{!}{
\begin{tabular}{cccccccccc}\hline
Reference &Year & Sensors & Map type & Core Solution & Confidence model & Map element & Change detection & Data collection & Environment \\\hline
\cite{Jo2018} & 2018 & \makecell{GNSS, monocular\\ camera} & Vector HD map & \makecell{Rao-Blackwellized\\ particle filter} & \makecell{DS evidence\\ theory} & \makecell{Road marking,lane \\lines, traffic light,\\ traffic sign} & Single-session & Centralized &\makecell{Online \\vehicle data}\\

\cite{Liu2019} & 2019 & \makecell{-} & \makecell{Lightweight \\ Vector HD map} & \makecell{Incremental fusing} & \makecell{Distance \\analysis} & \makecell{Lane lines} & - &Crowdsourced &\makecell{Simulation}\\

\cite{Pannen2019} & 2019 & GNSS, camera & Vector HD map & Particle filter & \makecell{Particle,belief \\ weight}& Traffic bin & Multi-session &- &Simulation\\

\cite{Liebner2019} & 2019 & \makecell{GNSS, monocular \\ camera} & Vector HD map & \makecell{Graph SLAM} &\makecell{Consistency,\\Spatial fit}  & \makecell{Lane lines} & - &Crowdsourced &\makecell{Offline \\vehicle data}\\

\cite{Heo2020} & 2020 & \makecell{GNSS, monocular\\ camera} & Vector HD map & \makecell{Deep metric \\learning} & Similarity score & \makecell{Road marking,\\ lane lines} & Single-session &- &\makecell{Offline \\vehicle data}\\

\cite{Pannen2020} & 2020 & GNSS, camera & Vector HD map & Probabilistic model & \makecell{Consistency,\\Spatial fit} &\makecell{Lane lines} & Multi-session &- &\makecell{Offline \\vehicle data}\\

\cite{Li2020} & 2020 & \makecell{GNSS, monocular\\ camera} & Vector HD map & \makecell{Uncertainty \\analysis} & Bayesian model & \makecell{Lane lines} & Multi-session &- &\makecell{Offline \\vehicle data}\\

\cite{Kimlidar2021} & 2021 & \makecell{GNSS, LIDAR} & Dense HD map & \makecell{Graph SLAM} &  \makecell{Likelihood \\analysis} & \makecell{Road boundary} & Multi-session &Crowdsourced &\makecell{Offline \\vehicle data}\\

\cite{Zhang2021} & 2021 & \makecell{GNSS, IMU, CAN,\\ monocular camera} & Vector HD map & \makecell{Clustering} & Matching degree & \makecell{Road marking, \\ lane lines} & Single-session &- &\makecell{Online \\vehicle data}\\

\cite{Lambert2021} & 2021 & \makecell{LIDAR, monocular \\ camera} & Vector HD map & \makecell{Deep learning} & -  & \makecell{Lane lines} & Multi-session & - &\makecell{Offline \\vehicle data}\\

\cite{Kim2021} & 2021 & \makecell{GNSS, monocular \\ camera} & Vector HD map & \makecell{Clustering} &  \makecell{Degree of change} & \makecell{Lane lines} & - &Crowdsourced &\makecell{Offline \\vehicle data}\\

\cite{Wijaya2022a} & 2022 & \makecell{-} & Vector HD map & \makecell{Blockchain} &  \makecell{-} & \makecell{Lane lines,\\ poles} & - &Crowdsourced &\makecell{Simulation}\\

\cite{Cho2023} & 2023 & \makecell{GNSS, monocular \\ camera} & Vector HD map & \makecell{Convex hull\\map matching} &  \makecell{Bayesian model} & \makecell{Lane lines} & Multi-session &Crowdsourced &\makecell{Offline \\vehicle data}\\

\cite{Zhanabatyrova2023} & 2023 & \makecell{GNSS, monocular \\ camera} & Dense HD map & \makecell{Multi view \\ geometry} &  \makecell{Object detection \\ confidencel} & \makecell{Traffic signs} & Multi-session &Crowdsourced &\makecell{Offline and Online \\vehicle data}\\
\cite{Sayed2023} & 2023 & \makecell{GNSS, monocular \\ camera} & Vector HD map & \makecell{VectorMapNet} &  \makecell{Object detection \\ confidencel} & \makecell{Road markings} & - &Crowdsourced &\makecell{Online \\nuScenes dataset}\\

\cite{Xiong2023} & 2023 & \makecell{GNSS/IMU, \\ surround-view camera} & Vector HD map & \makecell{Fusion learning} &  \makecell{Attention module} & \makecell{Road markings} & Single-session &Crowdsourced &\makecell{Online \\ nuScenes dataset}\\

\hline
\end{tabular}
}
\caption{Summary of existing papers discussed in map update module.}
\end{sidewaystable*} 

\end{appendices}
\end{document}